%% file: template.tex
\RequirePackage{fix-cm}
\documentclass[twocolumn]{svjour3}          
\smartqed  
\usepackage{graphicx}
\usepackage{amsmath}
\usepackage{amssymb}
\usepackage{booktabs}
\usepackage{colortbl}
\usepackage{xcolor}
\usepackage{float}
\usepackage{multirow}

\usepackage{gensymb} 

\usepackage{makecell} 
\usepackage[pagebackref=true,breaklinks=true,letterpaper=true,colorlinks,bookmarks=false]{hyperref}
\begin{document}
\sloppy

\title{Learning Transparent Object Matting}

\author{Guanying Chen* \and
        Kai Han* \and 
        Kwan-Yee K. Wong
}

\institute{Guanying Chen* \at
              The University of Hong Kong, Hong Kong, China\\
              \email{gychen@cs.hku.hk}           
           \and
           Kai Han* \at
              University of Oxford, Oxford, United Kingdom \\
              \email{khan@robots.ox.ac.uk}           \\
              (* indicates equal contribution) 
           \and
           Kwan-Yee K. Wong \at
              The University of Hong Kong, Hong Kong, China
}

\date{Received: date / Accepted: date}

\maketitle

\begin{abstract}
This paper addresses the problem of image matting for transparent objects. Existing approaches often require tedious capturing procedures and long processing time, which limit their practical use. In this paper, we formulate transparent object matting as a refractive flow estimation problem, and propose a deep learning framework, called {\em TOM-Net}, for learning the refractive flow. Our framework comprises two parts, namely a multi-scale encoder-decoder network for producing a coarse prediction, and a residual network for refinement. At test time, TOM-Net takes a single image as input, and outputs a matte (consisting of an object mask, an attenuation mask and a refractive flow field) in a fast feed-forward pass. As no off-the-shelf dataset is available for transparent object matting, we create a large-scale synthetic dataset consisting of $178K$ images of transparent objects rendered in front of images sampled from the Microsoft COCO dataset. We also capture a real dataset consisting of $876$ samples using $14$ transparent objects and $60$ background images. 
Besides, we show that our method can be easily extended to handle the cases where a trimap or a background image is available. 
Promising experimental results have been achieved on both synthetic and real data, which clearly demonstrate the effectiveness of our approach.
\end{abstract}

\section{Introduction}
\begin{figure}[h!] \centering
    \includegraphics[width=0.5\textwidth]{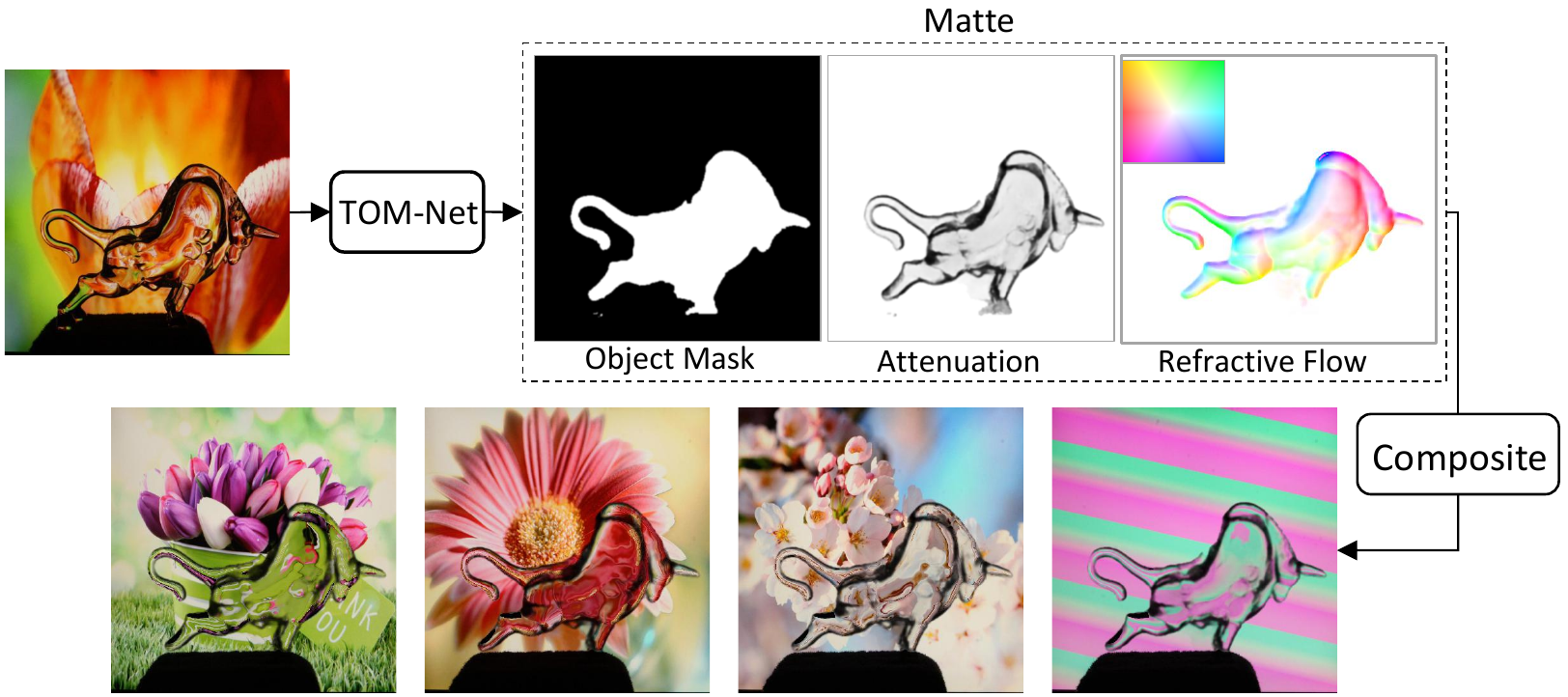}
    \caption{Given an image of a transparent object as input, our model can estimate the environment matte (consisting of an object mask, an attenuation mask and a refractive flow field) in a feed-forward pass. The transparent object can then be composited onto new background images with the extracted matte.}
    \label{fig:intro}
\end{figure}

Image matting refers to the process of extracting the foreground matte of an image by locating the region of the foreground object and estimating the opacity of each pixel inside the foreground region. The foreground object can then be composited onto a new background image using the \emph{matting equation} \cite{smith1996blue} 
\begin{equation}
    \label{eq:alphamatte}
    C = F + (1-\alpha)B,  \quad \alpha \in [0, 1],
\end{equation}
where $C$ denotes the composited color, $F$ the foreground color, $B$ the background color, and $\alpha$ the opacity.

Image matting has been widely used in image editing and film production. However, most of the existing methods are tailored for opaque objects, and cannot handle transparent objects whose appearance depends on how light is refracted from the background.

To model the effect of refraction, Zongker~\textit{et al.} \cite{zongker1999environment} introduced \emph{environment matting} as 
\begin{equation}
    \label{eq:em_general}
    C = F + (1-\alpha)B + \Phi, \quad \alpha \in [0, 1],
\end{equation}
where $\Phi$ is the contribution of environment light caused by refraction or reflection at the foreground object. Besides estimating the foreground shape, environment matting also describes how objects interact with the background. 

Many efforts \cite{chuang2000environment,wexler2002image,peers2003wavelet,zhu2004frequency,duan2011flexible,duan2015compressive} have been devoted to improving the seminal work of  \cite{zongker1999environment}. The resulting methods often require either a huge number of input images to achieve a higher accuracy, or specially designed patterns to reduce the number of required images. They are in general all very computational expensive.

In this paper, we focus on environment matting for transparent objects. It is highly ill-posed, if not impossible, to estimate an accurate environment matte for transparent objects from a single image with an arbitrary background. Given the huge solution space, there exist multiple objects and backgrounds which can produce the same refractive effect. In order to make the problem more tractable, we simplify our problem to estimating an environment matte that can produce visually realistic refractive effect from a single image, instead of estimating a highly accurate refractive flow. We define the environment matte in our model as a triplet consisting of an object mask, an attenuation mask and a refractive flow field. Realistic refractive effect can then be obtained by compositing the transparent object onto new background images (see Fig.~\ref{fig:intro}). We then show that the performance of the proposed method can be improved when a trimap or a background image is available.

Inspired by the great successes of convolutional neural networks (CNNs) in high-level computer vision tasks, we propose a convolutional neural network, called TOM-Net, for simultaneous learning of an object mask, an attenuation mask and a refractive flow field from a single image with an arbitrary background. The key contributions of this paper can be summarized as follows:

\begin{itemize}
  \item We introduce a simple and efficient model for transparent object matting as simultaneous estimation of an object mask, an attenuation mask and a refractive flow field.
  \item We propose a convolutional neural network, TOM-Net, to learn an environment matte of a transparent object from a single image. To the best of our knowledge, TOM-Net is the first CNN that is capable of learning transparent object matting.
  \item We create a large-scale synthetic dataset and a real dataset as a benchmark for learning transparent object matting. Our TOM-Net has produced promising results on both the synthetic and real datasets.
  \item We propose two convolutional neural networks, denoted as TOM-Net$^{\text{+Trimap}}$ and TOM-Net$^{\text{+Bg}}$, for handling the cases where a trimap or a background image is available, respectively.
\end{itemize}  

A preliminary version of this work appeared in \cite{chen2018tomnet}. This paper extends \cite{chen2018tomnet} in several aspects. First, we provide a more comprehensive comparison between our method and previous methods. 
Second, we present a more detailed ablation study, more experimental results, as well as analysis for failure cases. 
Third, we showcase an interesting application of image editing of transparent objects by manipulating the extracted environment matte. 
Fourth, we investigate how the performance of our method can be improved when a trimap or a background image is available.
Last, we discuss in detail the limitations and potential extensions of the current model. In particular, we introduce a potential formulation for handling colored objects with specular highlights.

The remainder of this paper is organized as follows. Section \ref{sec:related_work} briefly reviews existing methods for environment matting and recent CNN based methods for image matting. Section \ref{sec:formulation} introduces a simplified environment matting formulation for transparent object matting from a single image. Section \ref{sec:method} describes the proposed two-stage framework and learning details. Section \ref{sec:dataset} presents our synthetic and real dataset. Experimental results on both synthetic and real dataset are shown in Section \ref{sec:experiments}. Limitations and potential extensions are discussed in Section \ref{sec:discussion}, followed by conclusions in Section \ref{sec:conclusion}.

Our code, trained model and datasets can be found at \url{https://guanyingc.github.io/TOM-Net}.

\begin{table*} \centering
    \caption{Comparison of different environment matting methods. $k$ indicates the image size and mapping type stands for how a foreground point is composited by the point(s) in the background image.}
    \label{tab:related_work}
    \resizebox{\textwidth}{!}{
    \Large
    \begin{tabular}{ccccccc}
        \toprule
        \textbf{Methods} & \makecell{\textbf{Asymptotic} \\ \textbf{\# images}}  & \makecell{\textbf{\# images} \\($k=1024)$} & \makecell{\textbf{Typical runtime}\\($k=1024$)} & \textbf{Mapping type} & \textbf{Materials} & \textbf{Remarks} \\
        \midrule
        Ours                                         & $O(1)$  & $1$              & \makecell{$0.5$ secs when $k=512$ \\ (run on a GPU)} & single-pixel & colorless, specularly refractive & \makecell{aims for visually realistic effect} \\
        \midrule                                                                    
RTCEM \cite{chuang2000environment}                    & $O(1)$  & $1$              & $2$ mins & single-pixel & colorless, specularly refractive & \makecell{requires a coded background \\ and off-line processing}\\
        \midrule                                                                    
Yeung et. al \cite{yeung2011matting}                 & $O(1)$  & $1$              & $30$ secs & single-pixel & colored refractive  & \makecell{requires human interaction, \\ aims for visually realistic effect}\\
        \midrule                                                                    
Zongker \textit{et al}. \cite{zongker1999environment}& $O(\log k)$ & $20$         & $20$ mins when $k=512$& single-region & \makecell{colored refractive, translucent, \\ highly specular} & assumes rectangular support region \\
        \midrule                                                                    
Chuang \textit{et al}. \cite{chuang2000environment}  & $O(k)$ & $1800$             & not available & multi-region & \makecell{Zongker \textit{et al}. \cite{zongker1999environment} + (color dispersion, \\ multiple mapping, glossy reflection)} & \makecell{requires solving a complex \\ optimization problem}\\
        \midrule                                                                    
Wavelet \cite{wang2004image}                         & $O(k)$ & $ 2400$            & $12$ hours & multi-region&  same as Chuang \textit{et al}. \cite{chuang2000environment} & runtime includes data acquisition \\
        \midrule                                                                    
Frequency \cite{zhu2004frequency}                    & $O(k)$ & $ 4096$            & $5-10$ mins & multi-pixel & \makecell{Zongker \textit{et al}. \cite{zongker1999environment} + (color dispersion, \\ glossy reflection)} & slow data acquisition   \\
        \midrule                                                                    
Duan \textit{et al}. \cite{duan2011fast}             & $O(s\log (k^2/s))$ & $ 340$ & $2.8$ mins & multi-region&  same as Chuang \textit{et al}. \cite{chuang2000environment} & $s$ denotes the sparsity of a signal \\ 
        \midrule                                                                    
Qian \textit{et al}. \cite{qian2015frequency}        & $O(s\log (2k/s))$ & $ 400$  & $3.3$ mins & multi-pixel & same as Frequency \cite{zhu2004frequency} & $s$ denotes the sparsity of a signal\\
        \bottomrule
    \end{tabular}
    }

\end{table*}

\section{Related Work}
\label{sec:related_work}
In this section, we briefly review representative works on environment matting and recent works on CNN based image matting. 

\paragraph{\bf Environment matting}
\label{par:Environment Matting}
Zongker~\textit{et al.} \cite{zongker1999environment} introduced the concept of environment matting, and assumed each foreground pixel being originated from a single rectangular region of the background. They obtained the environment matte by identifying the corresponding background region for each foreground pixel using three monitors and multiple images. Chuang \textit{et al.} \cite{chuang2000environment} extended \cite{zongker1999environment} in two ways. First, they replaced the single rectangular supporting area for a foreground pixel with multiple 2D oriented Gaussian strips. This makes it possible for their method to model the effects of color dispersion, multiple mapping and glossy reflection. Second, they simplified the environment matting equation by assuming the object being colorless and perfectly transparent. This allows them to achieve real time capture environment matting (RTCEM). The environment matte was then extracted with one image taken in front of a pre-designed pattern. However, RTCEM requires background images to segment the transparent objects, and depends on a time-consuming off-line processing. Wexler \textit{et al.} \cite{wexler2002image} introduced a probabilistic model based method which assumes each background point has a probability to make contribution towards the color of a certain foreground point. Their approach does not require pre-designed patterns during data acquisition, but it still needs multiple images and can only model thin transparent objects. Peers and Dutr{\'e} \cite{peers2003wavelet} used a large number of wavelet basis backgrounds to obtain the environment matte, and their method can also model the effect of diffuse reflection. Based on the fact that a signal can be decomposed uniquely in the frequency domain, Zhu and Yang \cite{zhu2004frequency} proposed a frequency-based approach to extract an accurate environment matte. They used Fourier analysis to solve the decomposition problem. Both \cite{peers2003wavelet} and \cite{zhu2004frequency} require a large number of images to extract the matte (e.g., \cite{peers2003wavelet} needs $2,400$ images and \cite{zhu2004frequency} needs $4,096$ images for an image of size $1024\times 1024$), making them not very practical. Recently, compressive sensing theory has been applied to environment matting to reduce the number of images required. Duan \textit{et al.} \cite{duan2011fast} applied this theory in  the spatial domain and Qian \textit{et al.} \cite{qian2015frequency} applied it in the frequency domain. However, the number of images needed is still in the order of hundreds. In contrast, our work can estimate an environment matte from a single image in a fast feed-forward computation without the need for pre-designed patterns or additional background images.

Yeung \textit{et al.} \cite{yeung2011matting} proposed an interactive way to estimate an environment matte given an image containing a transparent object. Their method requires users to manually mark the foreground and background in the image, and models the refractive effect using a thin-plate-spline transformation. Their method does not produce an accurate environment matte, but instead a visually pleasing refractive effect. Our method shares the same spirit, but does not involve any human interaction. 

Tab. \ref{tab:related_work} shows a comparison of different environment matting methods. Compared with other methods, our method requires only a single image and can extract a matte in $0.5$ second without the need for any predefined backgrounds.

\paragraph{\bf CNN based image matting}
Although the potential of CNN on transparent object matting has not yet been explored, some existing work have adopted CNNs for solving the general image matting problem. Shen \textit{et al.} \cite{shen2016deep} introduced a CNN for image matting of color portrait images. Cho \textit{et al.} \cite{cho2016natural} proposed a network to predict a better alpha matte by taking the matting results of the traditional method and normalized color images as input. Xu \textit{et al.} \cite{xu2017deep} introduced a deep learning framework that can estimate an alpha matte from an image and its trimap. However, none of these methods can be applied directly to the task of transparent object matting as object opacity alone is not sufficient to model the refractive effect. 

\begin{figure*}[htb] \centering
    \includegraphics[width=\textwidth]{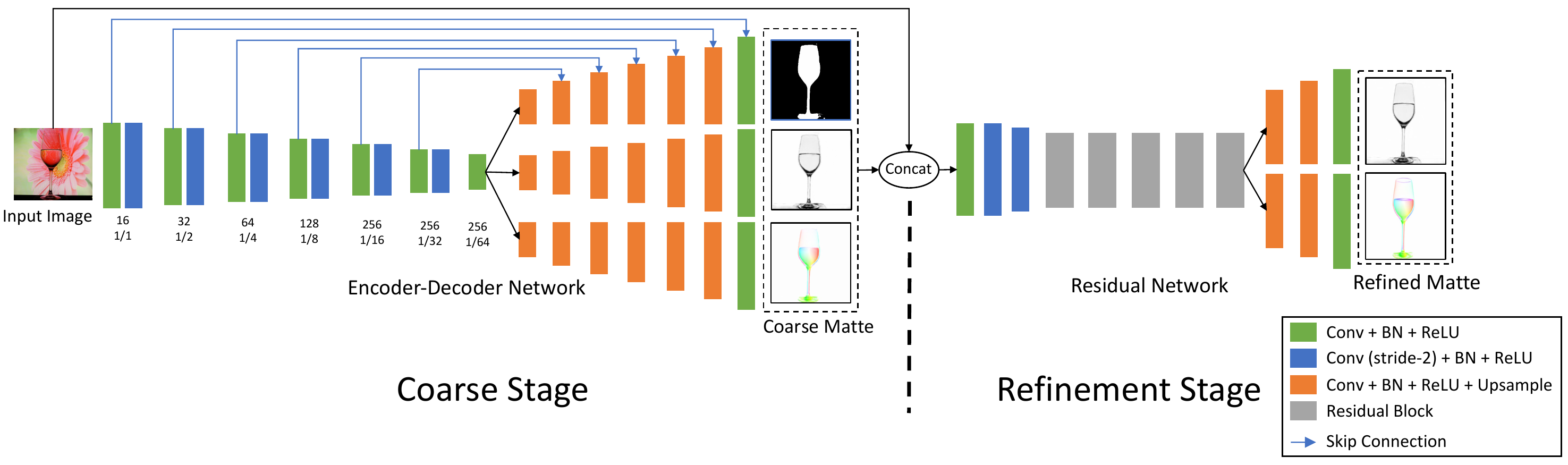}
    \caption{TOM-Net architecture. The left subnetwork is the CoarseNet and the right subnetwork is the RefineNet. (Cross-link and multi-scale outputs are not shown for simplicity.)} \label{fig:networkStructure}
\end{figure*}

\section{Matting Formulation}
\label{sec:formulation}
As a transparent object may have multiple optical properties (e.g., color attenuation, translucency and reflection), estimating an accurate environment matte for a generic transparent object from a single image is very challenging.  
Following the work of \cite{chuang2000environment}, we cast environment matting to a refractive flow estimation problem by assuming that each foreground pixel only originates from one point in the background due to refraction. Compared to the seminal work of \cite{zongker1999environment}, which models each foreground pixel as a linear combination of a patch in the background, our formulation is more tractable and can be easily encoded using a CNN.

In \cite{zongker1999environment}, the per-pixel environment matting is obtained through leveraging color information from multiple background images. Given a set of pre-designed background patterns, matting is formulated as
\begin{equation}
    \label{eq:em_origin}
    C = F + (1-\alpha)B + \sum_{i=1}^{k} R_i \mathcal{M}(\mathbf{T}_i, \mathbf{A}_i),
\end{equation}
where $F$, $B$ and $\alpha$ denote the ambient illumination, background color and opacity, respectively. The last term in (\ref{eq:em_origin}) accounts for the environment light accumulated from $k$ pre-designed background images ($k=3$ in \cite{zongker1999environment}). $R_i$ is a factor describing the contribution of light emanating from the $i$-$th$ background image $\mathbf{T}_i$. $\mathcal{M}(\mathbf{T}_i, \mathbf{A}_i)$ denotes the average color of a rectangular region $\mathbf{A}_i$ on the background image $\mathbf{T}_i$. 

To obtain an environment matte, the transparent object is placed in front of the monitor(s), and multiple pictures of the object are captured with the monitor(s) displaying different background patterns\footnote{For an image of size $512\times 512$, $18$ pictures and around $20$ minutes processing time are needed.}. Generally, a surface point receives light from multiple directions, especially for a diffuse surface. When it comes to a perfectly transparent object, however, a surface point will only receive light from one direction as determined by the law of refraction. Consider a single background image as the only light source (i.e., no ambient illumination), the problem can be modeled as
\begin{equation}
    \label{eq:em_simplify1}
    C = (1-\alpha)B + R \mathcal{M}(\mathbf{T}, P),
\end{equation}
where $\mathcal{M}(\mathbf{T}, P)$ is a bilinear sampling operation at location $P$ on the background image $\mathbf{T}$. Further, by assuming a colorless transparent object, $R$ becomes a  light attenuation index $\rho$ (a scalar value). The formulation in (\ref{eq:em_simplify1}) can be simplified to
\begin{equation}
    \label{eq:em_simplify2}
    C = (1-\alpha)B + \rho \mathcal{M}(\mathbf{T}, P),
\end{equation}
where $\rho \in [0, 1]$ denotes the attenuation index.

Here, we use refractive flow to model the refractive effect of a transparent object. The refractive flow of a foreground pixel is defined as the offset between the foreground pixel and its refraction correspondence on the background image. 

We further introduce a binary foreground mask to define the object region in the image. The matting equation can now be rewritten as
\begin{equation}
    \label{eq:em_simplify3}
    C = (1 - m) B + m\rho \mathcal{M}(\mathbf{T}, P),
\end{equation}
where $m \in\{0, 1\}$ denotes background ($m = 0$) or foreground ($m = 1$). The matte can then be estimated by solving $m$, $\rho$ and $P$ for each pixel in the input image containing the transparent 
object\footnote{For an image with $n$ pixel, we have $7$ unknowns ($3$ for $B$, $2$ for $P$, $1$ for $m$, and $1$ for $\rho$) for each pixel, resulting in a total of $7n$ unknowns.}.

\section{Learning Transparent Object Matting}
\label{sec:method}
In this section, we present a two-stage deep learning framework, called TOM-Net, for learning transparent object matting (see Fig.~\ref{fig:networkStructure}). The first stage, denoted as CoarseNet, is a multi-scale encoder-decoder network that takes a single image as input, and predicts an object mask, an attenuation mask and a refractive flow field simultaneously. CoarseNet is capable of predicting a robust object mask. However, the estimated attenuation mask and refractive flow field lack local structural details. 
To overcome this problem, we introduce the second stage of TOM-Net, denoted as RefineNet, to achieve a sharper attenuation mask and a more detailed refractive flow field. RefineNet is a residual network \cite{he2016deep} that takes both the input image and the output of CoarseNet as input. After training, our TOM-Net can predict an environment matte from a single image in a fast feed-forward pass.

\subsection{Encoder-Decoder for Coarse Prediction}
\label{sub:Encoder-decoder Net for Coarse Prediction}
The first stage of our TOM-Net (i.e., CoarseNet) is based on mirror-link CNN introduced in \cite{shi2016learning}. Mirror-link CNN was proposed to learn non-lambertian object intrinsic decomposition. Its output consists of an albedo map, a shading map and a specular map. It shares a similar output structure with our transparent object matting task (i.e., three output branches sharing the same spatial dimensionality). Therefore, it is reasonable for us to adapt mirror-link CNN for our CoarseNet. 

The mirror-link CNN adapted for our CoarseNet consists of one shared encoder and three distinct decoders. The encoder contains six down-sampling convolutional blocks, leading to a down-sampling factor of $64$ in the bottleneck layer. Features in the encoder layers are connected to the decoder layers having the same spatial dimensions through skip connections \cite{ronneberger2015u}. Cross-links \cite{shi2016learning} are introduced to make different decoders share the same input in each layer, so that decoders can better utilize the correlation between different predictions.

Learning with multi-scale loss has been proven to be helpful in dense prediction tasks (e.g., \cite{eigen2014depth,fischer2015flownet}). Since we formulate the problem of transparent object matting as refractive flow estimation, which is a dense prediction task, we augment our mirror-link CNN with multi-scale loss similar to \cite{fischer2015flownet}. 
We use four different scales in our model, where the first scale starts from the decoder features with a down-sampling factor of $8$ and the largest scale has the same spatial dimensions as the input.

In contrast to the recent two-stage framework for image matting \cite{xu2017deep}, our TOM-Net has a shared encoder and three parallel decoders to accommodate different outputs. Besides, we augment our CoarseNet with multi-scale loss and cross-link. Moreover, TOM-Net is trained from scratch while the encoder in \cite{xu2017deep} is initialized with the pre-trained VGG16.

\subsection{Loss Function for Coarse Stage}
\label{sub:Loss Function for CoarseNet}
CoarseNet takes a single image as input and predicts the environment matte as a triplet consisting of an object mask, an attenuation mask and a refractive flow field. The learning of CoarseNet is supervised by the ground-truth matte using an object mask segmentation loss $\mathcal{L}_{ms}$, an attenuation regression loss $\mathcal{L}_{ar}$, and a refractive flow regression loss $\mathcal{L}_{fr}$. Besides, the predicted matte is expected to render an image as close to the input image as possible when applied to the ground-truth background. Hence, in addition to the supervision of the matte, we also take image reconstruction loss $\mathcal{L}_{ir}$ into account. Note that the ground-truth background is only used to calculate the reconstruction error during training but not needed during testing. CoarseNet can therefore be trained by minimizing 
\begin{align}
    \mathcal{L}^c = \alpha^c_{ms} \mathcal{L}_{ms} + \alpha^c_{ar} \mathcal{L}_{ar} + \alpha^c_{fr} \mathcal{L}_{fr} + \alpha^c_{ir} \mathcal{L}_{ir},
\end{align}
where 
$\alpha^c_{ms}, \alpha^c_{ar}, \alpha^c_{fr}, \alpha^c_{ir}$ are weights for the corresponding loss terms. 

\paragraph{\bf Object mask segmentation loss}
\label{par:Object Mask Classification Loss}
Object mask segmentation is simply a spatial binary classification problem. The output of the object mask decoder has a dimension of $2\times H\times W$, where $H$ and $W$ denote the height and width of the input. We normalize the output with {\em softmax} and compute the loss using the binary cross-entropy function
\begin{equation}
    \mathcal{L}_{ms} = -\frac{1}{HW} \sum_{ij} (\tilde{M}_{ij}\log(P_{ij}) + (1-\tilde{M}_{ij}) \log(1-P_{ij})),
\end{equation}
where $\tilde{M}_{ij} \in \{0,1\}$ and $P_{ij}\in [0,1]$ represent ground truth and normalized foreground probability of the pixel at $(i, j)$, respectively.

\paragraph{\bf Attenuation regression loss} 
The predicted attenuation mask has a dimension of  $1\times H\times W$. The value of this mask is in the range of $[0, 1]$, where $0$ indicates no light can pass and $1$ indicates the light will not be attenuated. 
We adopt a mean square error (MSE) loss
\begin{equation}
    \mathcal{L}_{ar} = \frac{1}{HW} \sum_{ij} (A_{ij}-\tilde{A}_{ij})^2,
\end{equation}
where $A_{ij}$ is the predicted attenuation index and $\tilde{A}_{ij}$ the ground truth at $(i, j)$.
\paragraph{\bf Refractive flow regression loss}
\label{par:Disparity Smoothness Loss}
The predicted refractive flow field has a dimension of $2\times H\times W$, where we have one channel for the horizontal displacement and another for the vertical displacement. We normalize the refractive flow with $tanh$ activation and multiply it by the width of the input, such that the output is constrained in the range of $[-W, W]$.
We adopt an average end-point error (EPE) loss
\begin{equation}
    \mathcal{L}_{fr} = \frac{1}{HW} \sum_{ij} \sqrt{(F^x_{ij}-\tilde{F}^x_{ij})^2 + (F^y_{ij}-\tilde{F}^y_{ij})^2},
\end{equation}
where $(F^x, F^y)$ and $(\tilde{F}^x, \tilde{F}^y)$ denote the predicted flow and the ground truth, respectively.

\paragraph{\bf Image reconstruction loss}
\label{par:Image Reconstruction Loss}
We use MSE loss to measure the dissimilarity between the reconstructed image and the input image. 
Denoting the reconstructed image by $I$ and the ground-truth image (i.e., the input image) by $\tilde{I}$, the reconstruction loss is given by
\begin{equation}
    \mathcal{L}_{ir} = \frac{1}{HW} \sum_{ij} \Vert I_{ij}-\tilde{I}_{ij}\Vert_2^2.
\end{equation}

\paragraph{\bf Implementation details}
\label{par:Implementation Details}
In all experiments, we empirically set $\alpha^c_{ms}=0.1, \alpha^c_{ar}=1, \alpha^c_{fr}=0.01,$ and $\alpha^c_{ir}=1$. The loss weights for different scales are $\frac{1}{2^{(4 -\text{s})}}$, where $s \in\{1,2,3,4\}$ denotes the scale. 
CoarseNet contains $8M$ parameters and it takes about $2.5$ days to train with Adam optimizer \cite{kingma2014adam} on a single NVIDIA Titan X Pascal GPU.  We first train the CoarseNet from scratch until convergence and then train the RefineNet. 

\subsection{Residual Learning for Matte Refinement}
\label{par:Residual Learning for Matte Refinement}
As the attenuation mask and the refractive flow field predicted by the CoarseNet lack structural details, a refinement stage is needed to produce a detailed matte. Observing that residual learning is particularly suitable for tasks whose input and output are largely similar \cite{kim2016accurate,Nah_2017_CVPR}, we propose a residual network, denoted as RefineNet, to refine the matte predicted by the CoarseNet. 
Similar strategy has also been successfully applied to progressively refine the estimated optical flow in \cite{ilg2017flownet}.

We concatenate the input image and the output of the CoarseNet to form the input of the RefineNet. As the object mask predicted by the CoarseNet is already plausible, the RefineNet only outputs an attenuation mask and a refractive flow field. The parameters of the CoarseNet are fixed when training the refinement stage. 

\paragraph{\bf Loss for the refinement stage}
\label{par:Loss for Refinement}
The overall loss for the refinement stage is
\begin{align}
    \mathcal{L}^r = \alpha^r_{ar} \mathcal{L}_{ar} + \alpha^r_{fr} \mathcal{L}_{fr} ,
\end{align}
where $\mathcal{L}_{ar}$ is the refinement attenuation regression loss, $\mathcal{L}_{fr}$ the refinement flow regression loss,  and $\alpha^r_{ar}$, $\alpha^r_{fr}$ their weights. The definitions of these two losses are identical to those defined in the first stage. 
We found that adding the image reconstruction loss in the refinement stage did reduce the image reconstruction error during training, but was not helpful in preserving sharp edges of the refractive flow field (e.g., mouth of a glass). This could be explained by the fact that a lower image reconstruction loss does not guarantee a better refractive flow field. As the matte estimated by the CoarseNet has already achieved a small reconstruction error, simultaneously optimizing the flow regression loss and image reconstruction loss in the refinement stage may compromise the flow estimation.
Since our goal in the refinement stage is to estimate a more detailed matte, we remove the image reconstruction loss to make our network focus on reducing the flow regression loss.


\paragraph{\bf Implementation details}
\label{par:Implementation Details}
We set $\alpha^r_{ar}=1$, $\alpha^r_{fr}=1$ for the refinement. RefineNet contains $1M$ parameters and it takes about $2$ days to train with Adam optimizer on a single NVIDIA Titan X Pascal GPU. RefineNet is randomly initialized during training.
\subsection{Improvement with Trimap and Background Image}
\label{ssec:trimap}
As the problem of transparent object matting from a single image is highly ill-posed, we investigate how to reinforce our framework by utilizing additional information. 
In particular, we consider the cases where a trimap or a background image is available.
Our framework can be easily extended to make use of these additional information by taking the concatenation of the input image and the background image (or trimap) as input, while keeping the overall network architecture unchanged.

\paragraph{\bf TOM-Net$^{\text{+Trimap}}$}
Trimap can provide a rough location of the transparent object to help the model better locate the transparent object. The trimap used in this paper is a single channel image with $3$ different values, where values $0$, $1$, and $2$ indicate background, unknown, and foreground regions, respectively.  
During training, we randomly generate trimaps based on the ground-truth object mask. We first perform random erosion and cropping on the object mask to form the known (rough) foreground region. The unknown region is then generated by subtracting the foreground region from a tight bounding box of the object mask, leaving the rest of the regions as the background region.
The variant model, denoted as TOM-Net$^{\text{+Trimap}}$, takes both the input image and trimap as input, giving rise to an input channel number of $4$ in the first convolutional layer. 
\paragraph{\bf TOM-Net$^{\text{+Bg}}$}
Given the background image, the model can easily identify the accurate location of the transparent object based on the difference of the input and background images. Moreover, having access to the background image allows the model to better estimate the refractive flow field.
The variant model, denoted as TOM-Net$^{\text{+Bg}}$, takes both the input and background images as input, giving rise to an input channel number of $6$ in the first convolutional layer.

TOM-Net$^{\text{+Trimap}}$ and TOM-Net$^{\text{+Bg}}$ are trained with the same procedure as TOM-Net.
Our experimental results show that with the additional information, our framework can achieve better results on both synthetic and real dataset.
\begin{figure} \centering
    \input{syn_data_samples}
    \caption{Examples of synthetic data. Up to down: examples of glass, glass with water, lens and complex shape, respectively. First three columns: background image, rendered image, refractive flow visualization (sparse). Last three columns: ground-truth refractive flow field, object mask, attenuation mask. (Best viewed in PDF with zoom.)} 
    \label{fig:syn_data_samples}
\end{figure}
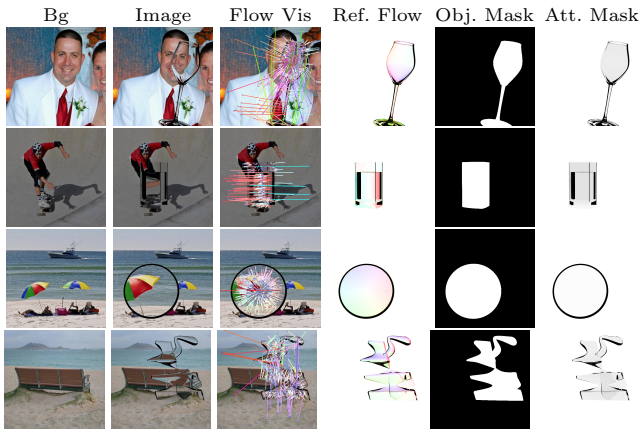

\section{Dataset for Learning and Evaluation}
\label{sec:dataset}
As no off-the-shelf dataset for transparent object matting is available, and it is very tedious and difficult to produce a large real dataset with ground-truth object masks, attenuation masks and refractive flow fields, we created a large-scale synthetic dataset by using \emph{POV-Ray} \cite{povray} to render images of synthetic transparent objects. Besides, we also captured a real dataset for evaluation. We will show that our TOM-Net trained on the synthetic dataset can generalize well to real world objects, demonstrating its good transferability.

\begin{table} \centering
    \caption{Statistics of our synthetic datasets.}
    \label{tab:synth}
    \resizebox{0.48\textwidth}{!}{
    \Large
    \begin{tabular}{c|*{4}{c}|c}
        \toprule
        Type & Glass & Glass \& Water & Lens & Complex & Total \\
        \midrule
        Synthetic Train & $52K$ & $26K$ & $20K$ & $80K$ & $178K$\\
        Synthetic Test   & $250$ & $250$ & $200$ & $200$ & $900$\\
        \bottomrule
    \end{tabular}
    }
\end{table}

\subsection{Synthetic Dataset} \label{sub:Synthetic Dataset}
We used a large number of background images and 3D models to render our training samples. We randomly changed the pose of the models, as well as the viewpoint and focal length of the camera in the rendering process to avoid overfitting to a fixed setting.

\paragraph{\bf Background images}
\label{par:Backgrounds Images}
We employed two types of background images, namely scene images and synthetic patterns. For scene images, we randomly sampled images from the Microsoft COCO \cite{lin2014microsoft} dataset\footnote{Other large-scale datasets like ImageNet \cite{deng2009imagenet} can also be used.}. The background images for the synthetic training set are sampled from COCO Train2014 and Test2015, while that for the synthetic test dataset are from COCO Val2014, giving rise to $100K$ scene images in total. 
For synthetic patterns, we rendered $40K$ patterns of size $512\times 512$ using \emph{POV-Ray} built-in textures. 

\paragraph{\bf Transparent objects}
\label{par:Transparent Object}
We divided common transparent objects into four categories, namely glass, glass with water, lens, and complex shape (see Fig.~\ref{fig:syn_data_samples} for examples). We constructed parametric 3D models for the first three categories, and generated a large number of models using random parameters. For complex shapes, we constructed parametric 3D models for basic shapes like sweeping-spheres and squashed surface of revolution (SOR) parts, and composed a larger number of models using these basic shapes. We generated $178K$ 3D models in total, with each model assigned a random refractive index $\lambda \in [1.3, 1.5]$. The distribution of these models in four categories is shown in Tab.~\ref{tab:synth}. 

\paragraph{\bf Ground-truth matte generation}
\label{par:Ground Truth Generation}
We obtained the ground-truth object mask of a model by rendering it in front of a black background image and setting its color to white. Similarly, we obtained the ground-truth attenuation mask of a model by simply rendering it in front of a white background image. Finally, we obtained the ground-truth refractive flow field (see Fig.~\ref{fig:syn_data_samples}) of a model by rendering it in front of a sequence of Gray-coded patterns. Technical details for the data rendering can be found at \url{https://github.com/guanyingc/TOM-Net_Rendering}

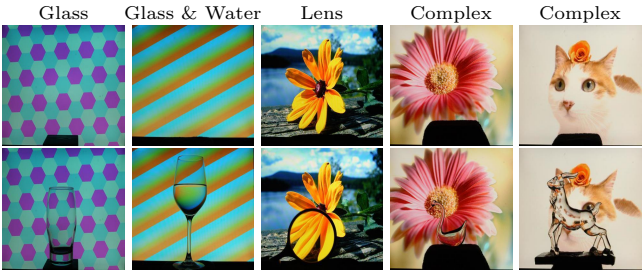
\begin{figure}[t] \centering
    \input{real_data_samples}
    \caption{Sample images in real dataset. The first row shows the background images and the second row shows the images of transparent objects.} \label{fig:real_sample}
\end{figure}

\begin{table}[t] \centering
    \caption{Statistics of our real dataset. The first and second rows show the number of objects and the number of backgrounds used during data acquisition, respectively. The last row shows the number of captured samples. Note that the category of glass with water are created by filling five of the glasses with different amount of water, and some backgrounds are shared between different shape categories.}
    \label{tab:real_data}
    \resizebox{0.45\textwidth}{!}{
    \Large
    \begin{tabular}{c|*{4}{c}}
        \toprule
        & Glass & Glass \& Water & Lens & Complex \\
        \midrule                                    %
        \# Objects     & $7$   & ($5$ glasses used)  & $1$   & $6$      \\
        \# Backgrounds & $60$  & $38$  & $4$   & $18$     \\
        \midrule                                    %
        \# Samples     & $470$ & $103$ & $61$  & $242$    \\
        \bottomrule
    \end{tabular}
    }
\end{table}

\paragraph{\bf Data augmentation}
\label{par:}
To improve the diversity of the training data and narrow the gap between real and synthetic data, extensive data augmentation was carried out on-the-fly. 
For an image of size $512\times 512$ with color intensity normalized to $[0, 1]$, we randomly performed color (brightness, contrast and saturation) augmentation (in a range of $[-0.2, 0.2]$), image scaling (in a range of $[0.875, 1.05]$), noise perturbation (in a range of $[-0.05, 0.05]$), and horizontal/vertical flipping. Besides, we also blurred the object boundary to make the synthetic data visually more natural. A patch with a size of $448\times 448$ was then randomly cropped from an augmented image and used as input to train CoarseNet. To speed up the training and save memory, a smaller patch with a size of $384\times 384$ was used to train RefineNet after the training of CoarseNet.

\subsection{Real Dataset}
\label{sub:Real Dataset}
To validate the transferability of TOM-Net, we introduce a real dataset, which was captured using $14$ objects\footnote{The objects consist of $7$ glasses, $1$ lens and $6$ complex objects. Glasses with water are implicitly included.} and $60$ background images, resulting in a dataset of $876$ images. Note that the background images for real data have not been used in the synthetic training or test dataset. The data distribution is summarized in Tab.~\ref{tab:real_data}. During the data capturing process, the objects were placed under different poses, with the distances between the camera, object and background uncontrolled. Fig.~\ref{fig:real_sample} shows some sample images from the real dataset.  Note that we do not have the ground-truth matte for the real dataset. We instead captured images of the backgrounds without the transparent objects to facilitate evaluation.

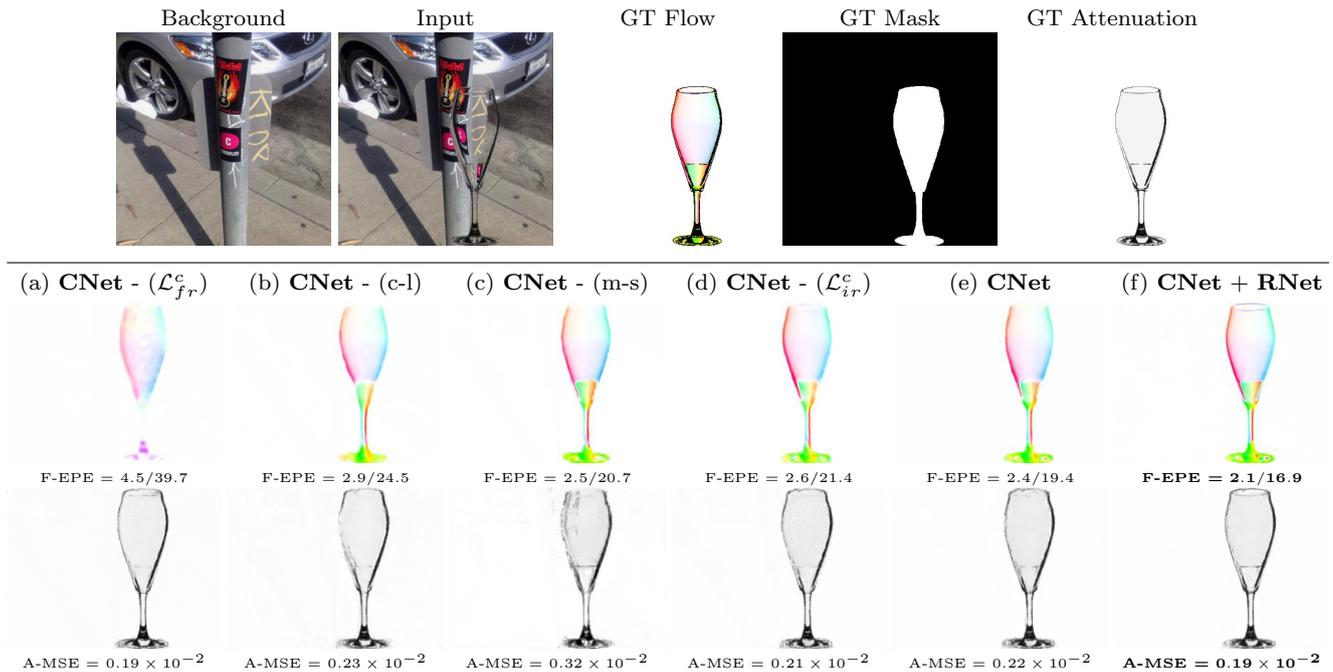
\begin{figure*} \centering
    \input{syn_ablation_study}
    \caption{Qualitative comparison of different model variants in ablation study. The first row shows a sample of \textit{glass with water} from the synthetic test dataset. The second and third rows show the estimated refractive flow fields and attenuation masks by different variants, respectively. (\textbf{CNet} and \textbf{RNet} are short for CoarseNet and RefineNet.)} \label{fig:syn_ablation_study}
\end{figure*}

\begin{table}
    \centering
    \caption{Ablation study. F, A, I, and M are short for flow, attenuation, image reconstruction, and object mask, respectively. (The first value for EPE is measured on the whole image and the second measured within the object region. A-MSE and I-MSE are computed on the whole image.)
    }
    \label{tab:self_compare}
    \resizebox{0.25\textwidth}{!}{
        \begin{tabular}{ccc}
            \midrule
            MSE ($\cdot10^{-2}$)  & \cellcolor{red!25} $\downarrow$ better &  \cellcolor{blue!25} $\uparrow$ better \\
            \midrule
        \end{tabular}
    }
    \\
    \resizebox{0.48\textwidth}{!}{
        \begin{tabular}{*{6}{>{\Large}c}}
        \toprule
        ID & Model Variants & \cellcolor{red!25}F-EPE  & \cellcolor{red!25}A-MSE & \cellcolor{red!25}I-MSE &  \cellcolor{blue!25}M-IoU \\ 
        \midrule
        0 & Background                         & 6.5 / 41.0 & 1.58 & 0.87 & 0.15 \\
        1 & CoarseNet - ($\mathcal{L}^c_{fr}$) & 3.9 / 26.5 & 0.24 & 0.23 & 0.98 \\
        2 & CoarseNet - (cross-link)           & 2.5 / 17.2 & 0.30 & 0.21 & 0.97  \\
        3 & CoarseNet - (multi-scale)          & 2.4 / 16.6 & 0.69 & 0.25 & 0.94 \\
        4 & CoarseNet - ($\mathcal{L}^c_{ir}$) & 2.3 / 15.7 & 0.25 & 0.22 & 0.98 \\
        5 & CoarseNet                          & \textbf{2.2 / 15.4} & 0.28 & 0.18 & 0.97 \\
        \midrule
        6 & CoarseNet + RefineNet              & \textbf{2.0 / 13.7} & 0.25 & 0.19 & 0.97 \\
        7 & CoarseNet + (RefineNet+$\mathcal{L}^r_{ir}$) & 2.0 / 13.9 & 0.24 & 0.18 & 0.97 \\
        \bottomrule
    \end{tabular}
    }
\end{table}

\section{Experiments and Results}
\label{sec:experiments}
In this section, we present experimental results and analysis.
We performed ablation study for TOM-Net, and evaluated our approach on both synthetic and real data.  For synthetic data, we evaluated end-point error (EPE) for refractive flow fields, intersection over union (IoU) for object masks, mean square error (MSE) for attenuation masks and image reconstruction results, respectively.  For real data, due to the absence of ground-truth matte, evaluation on the absolute error with respect to the ground truth is not possible. 
Instead, we reconstructed the input images using the estimated mattes and background images, and then evaluated the PSNR and SSIM metrics \cite{wang2004image} between each pair of input image (i.e., photograph) and reconstructed image (i.e., composite). In addition, a user study was conducted to validate the realism of TOM-Net composites.

We showcased an application of image editing of transparent object by manipulating the extracted matte, and analyzed typical failure cases. We also investigated how the performance of our method can be improved when a trimap or a background image is available.

\subsection{Ablation Study for Network Structure}
\label{sub:Network Analysis}
We quantitatively analyzed different components of TOM-Net using synthetic dataset\footnote{Complex shape is excluded in experiments here to speed up training.}. We first verified the effectiveness of \emph{refractive flow regression loss} ($\mathcal{L}^{c}_{fr}$), \emph{cross-link}, \emph{multi-scale loss} and \emph{image reconstruction loss} ($\mathcal{L}^{c}_{ir}$) in the coarse stage by removing each of them from \emph{CoarseNet} during training. We then validated the effectiveness of \emph{RefineNet} in recovering details of the refractive flow field.  RefineNet was evaluated by adding it to a trained CoarseNet and was trained while fixing the parameters of CoarseNet. 
For comparison, we also included a naive baseline, denoted as \emph{Background}, by considering a zero matte case (i.e., whole image as object mask, no attenuation, and no refractive flow) where the reconstructed image is the same as the background image. The quantitative results are summarized in Tab.~\ref{tab:self_compare} and the qualitative comparisons are shown in Fig. \ref{fig:syn_ablation_study}.
Overall, the baseline \emph{Background} was outperformed by all TOM-Net variants with a large margin for all the evaluation metrics, which clearly shows that TOM-Net can successfully learn the matte.

\paragraph{\bf Effectiveness of refractive flow regression loss} Comparing experiments with IDs 1 \& 5 in Tab. \ref{tab:self_compare}, it can be clearly seen that the CoarseNet trained with the refractive flow regression loss significantly outperformed that without it in refractive flow estimation. This result indicates that image reconstruction loss alone is not enough to supervise the learning of refractive flow. Fig. \ref{fig:syn_ablation_study} (a \& e) qualitatively show that the refractive flow regression loss improved the performance of refractive flow estimation.

\begin{figure} \centering
    \input{refine_vis}
    \caption{Visualization of the effectiveness of the refinement stage on real data. After refinement, the refractive flow and attenuation mask have more clear structural details (e.g., glass mouth).} \label{fig:refine}
\end{figure}
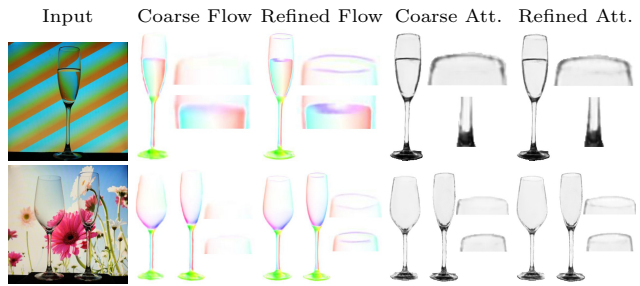

\begin{table*} \centering
    \caption{Quantitative results on the synthetic test dataset. (The first value for EPE is measured on the whole image and the second measured within the object region. A-MSE and I-MSE are computed on the whole image.)}
    \input{syn_quant}

    \label{tab:quant_synth}
\end{table*}

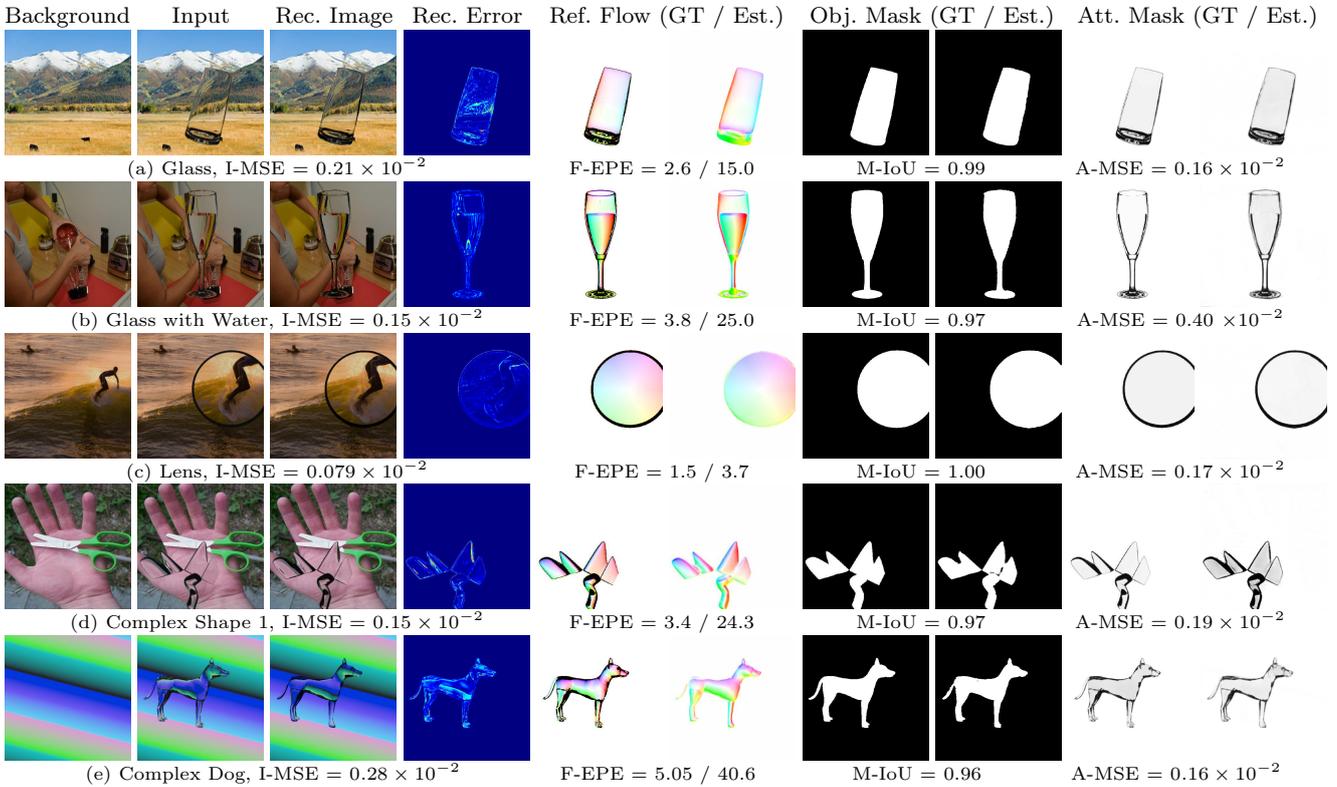
\begin{figure*}[tb] \centering
    \input{syn_qual}
    \caption{Qualitative results on synthetic data. The first to the fourth columns show background, input image, reconstructed image, and reconstruction error map, respectively. Quantitative results are shown below each example. Dark region in GT flow indicates no valid flow. (Best viewed in PDF with zoom.)} \label{fig:qual_synth}
\end{figure*}

\paragraph{\bf Effectiveness of cross-link} Comparing experiments with IDs 2 \& 5 in Tab. \ref{tab:self_compare}, we can see that augmenting the decoders of CoarseNet with the cross-link helped improve the performance in all metrics, suggested that utilizing correlation is helpful for the matte estimation.
Fig. \ref{fig:syn_ablation_study} (b \& e) qualitatively show the results without and with the cross-link during training.

\paragraph{\bf Effectiveness of multi-scale loss} Comparing experiments with IDs 3 \& 5 in Tab. \ref{tab:self_compare}, we can see that multi-scale loss boosted performance of CoarsNet in all of the evaluation metrics, particularly the attenuation mask MSE (see Fig. \ref{fig:syn_ablation_study} (c \& e) for qualitative comparison).

\paragraph{\bf Effectiveness of image reconstruction loss} Comparing experiments with IDs 4 \& 5 in Tab. \ref{tab:self_compare}, we can see that adding image reconstruction loss in the coarse stage slightly improved the performance of refractive flow estimation and was very effective for reducing the image reconstruction error (see Fig. \ref{fig:syn_ablation_study} (d \& e) for qualitative comparison).

\paragraph{\bf Effectiveness of RefineNet}
Comparing experiments with IDs 5 \& 6 in Tab. \ref{tab:self_compare}, we can clearly see that RefineNet can significantly improve the refractive flow estimation. Fig. \ref{fig:syn_ablation_study} (e \& f) and Fig. \ref{fig:refine} show that RefineNet can infer sharp details on both the synthetic and real data based on the outputs of CoarseNet, demonstrating the effectiveness of the RefineNet. We also found that image reconstruction loss is not helpful for refractive flow estimation in the refinement stage (experiments with IDs 6 \& 7 in Tab. \ref{tab:self_compare}). This is reasonable since the matte produced by CoarseNet already gives a small image reconstruction error, and further reducing the image reconstruction error does not guarantee a better refractive flow field. 

\subsection{Results on Synthetic Data}
\label{sub:Results on Synthetic data}
Quantitative results for synthetic test dataset are presented in Tab. \ref{tab:quant_synth}. We compared TOM-Net against \emph{Background} and CoarseNet. Here, to accelerate training convergence, we first trained CoarseNet from scratch using our synthetic dataset excluding the complex shape subset. The trained CoarseNet was then fine-tuned using the entire training set including complex shapes, followed by training of RefineNet on the entire training set with random initialization. Similar to previous experiments, TOM-Net outperformed \emph{Background} by a large margin, and slightly outperformed CoarseNet in both EPE and MSE, which implies more local details can be learned by RefinedNet. 

\begin{table} \centering
    \caption{Quantitative results on real data. (Value the higher the better.)}
    \resizebox{0.48\textwidth}{!}{ 
    \huge
    \begin{tabular}{c|*{2}{c}|*{2}{c}|*{2}{c}|*{2}{c}|*{2}{c}}
        \toprule
        \multirow{2}{*}{} & \multicolumn{2}{c}{Glass} 
                          & \multicolumn{2}{c}{G \& W} 
                          & \multicolumn{2}{c}{Lens} 
                          & \multicolumn{2}{c}{Complex} 
                          & \multicolumn{2}{c}{Avg} \\
        & PSNR & SSIM  & PSNR & SSIM & PSNR & SSIM & PSNR & SSIM & PSNR & SSIM \\
        \midrule
        Background    & 22.05 & 0.894 & 20.75 & 0.886 & 18.60 & 0.860 & 16.85 & 0.816 & 19.56 & 0.864 \\ 
        CoarseNet     & 25.09 & 0.921 & 23.53 & 0.911 & 21.13 & 0.895 & 17.89 & 0.835 & 21.91 & 0.891  \\ 
        TOM-Net        & 25.06 & 0.920 & 23.53 & 0.911 & 20.89 & 0.893 & 17.88 & 0.835 & 21.84 & 0.890 \\ 
        \bottomrule
    \end{tabular}
    }
    \label{tab:real_quant}
\end{table}

The average IoU for object mask estimation is $0.96$, indicates that TOM-Net can robustly segment the transparent object given only a single image as input.
Although TOM-Net is not expected to learn highly accurate refractive flow, the average EPE errors ($2.7/18.6$)\footnote{The first value is measured on the whole image and the second measured within the object region.} are very small compared with the size of the input image ($448\times 448$). In this sense, our predicted flow is capable of producing visually plausible refractive effect. The errors of complex shape category are larger than that of others, because complex shapes contain more sharp regions that will induce more errors. 
Fig. \ref{fig:qual_synth} shows the qualitative results on synthetic dataset. The objects in the first four rows come from the test set where each row shows a specific object category. Although the background images and objects in the test set never appear in the training set, TOM-Net can still predict robust matte. 
The last row shows a sample of complex dog shape, which was rendered using a 3D dog model. The pleasing result on the complex dog shape demonstrates that our model can generalize well from simple shapes to complex shapes.

\subsection{Results on Real Data}
\label{sub:Results}
\begin{figure*} \centering
    \input{real_qual}
    \caption{Qualitative results on real data. The PSNR and SSIM between input photographs and reconstructed images are shown below each example. The last column shows the composites on novel backgrounds given the estimated matte. (Best viewed in PDF with zoom.)} 
    \label{fig:real_qualitative}
\end{figure*}
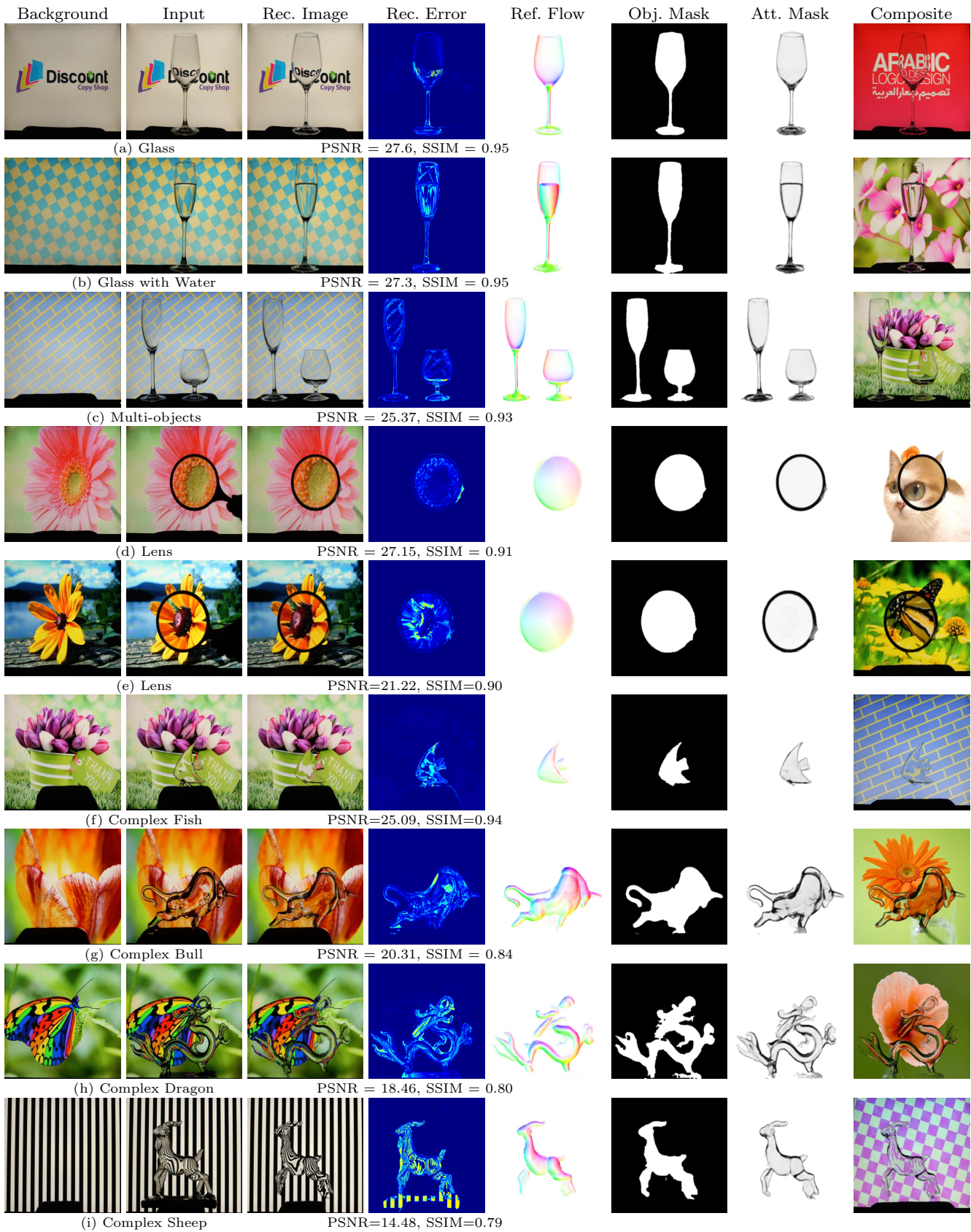

We evaluated TOM-Net on our captured real dataset, which consists of $876$ images of real objects. The results are shown in Tab. \ref{tab:real_quant}. The average PSNR and SSIM are above $21.0$ and $0.89$ respectively. The values are a bit lower for complex shapes, due to the opaque base of complex objects as well as the sharp regions of the objects that might induce large errors. After training, TOM-Net generalized well to common real transparent objects (see Fig. \ref{fig:real_qualitative}). It is worth to note that during training, each sample contains only one object, while TOM-Net can predict reliable matte for images containing multiple objects (see Fig. \ref{fig:real_qualitative} (c)), which indicates the transferability and robustness of TOM-Net.

\begin{table} \centering
    \caption{User study results. P, C, and N are short for votes for photograph, composite, and not distinguishable.}
    \huge
    \resizebox{0.48\textwidth}{!}{
        \begin{tabular}{c|*{3}{c}|*{3}{c}|*{3}{c}|*{3}{c}|*{3}{c}}
        \toprule
        \multirow{2}{*}{} & \multicolumn{3}{c}{Glass} 
                               & \multicolumn{3}{c}{G \& W} 
                               & \multicolumn{3}{c}{Lens} 
                               & \multicolumn{3}{c}{Complex}  
                               & \multicolumn{3}{c}{All}  \\
                               & P & C & N 
                               & P & C & N
                               & P & C & N 
                               & P & C & N 
                               & P & C & N \\
        \midrule
        Photographs        & 522 & 275 & 31 & 163 & 97 & 16 & 74 & 48 & 16 & 91 & 35 & 12 & 850 & 455 & 75 \\
        Composites    & 531 & 266 & 31 & 145 & 113 & 18 & 73 & 52 & 13 & 78 & 51 & 9 & 827 & 482 & 71 \\
        \bottomrule
    \end{tabular}
    }
    \label{tab:user_study}
\end{table}

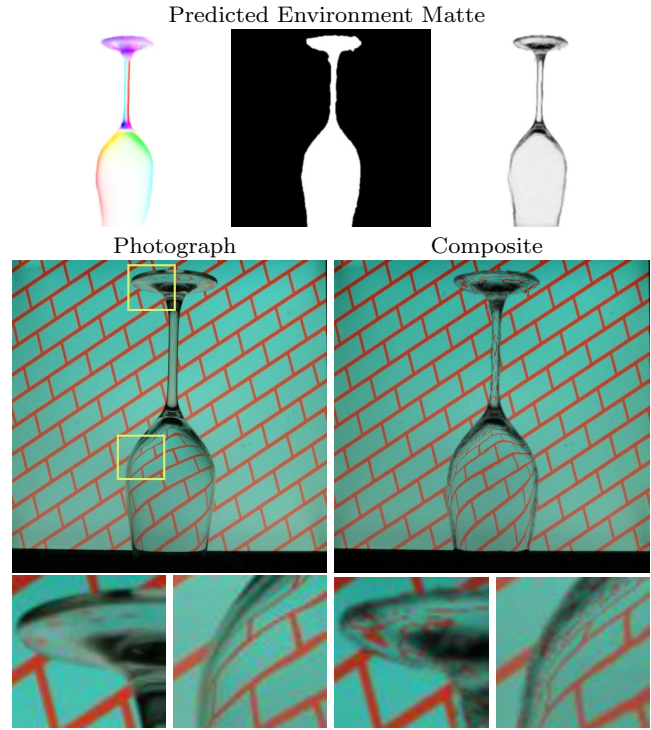
\begin{figure} \centering
    \input{user_study_reason}
    \caption{The first row shows the predicted matte, which is estimated by taking the photograph as input to our method. The second row compares the photograph and composite and the third row shows the zoom-in comparisons. When looking at the photograph and composite simultaneously, users can easily spot some imperfections of the composites (mostly in the boundary region).}
    \label{fig:sup_user_study}
\end{figure}

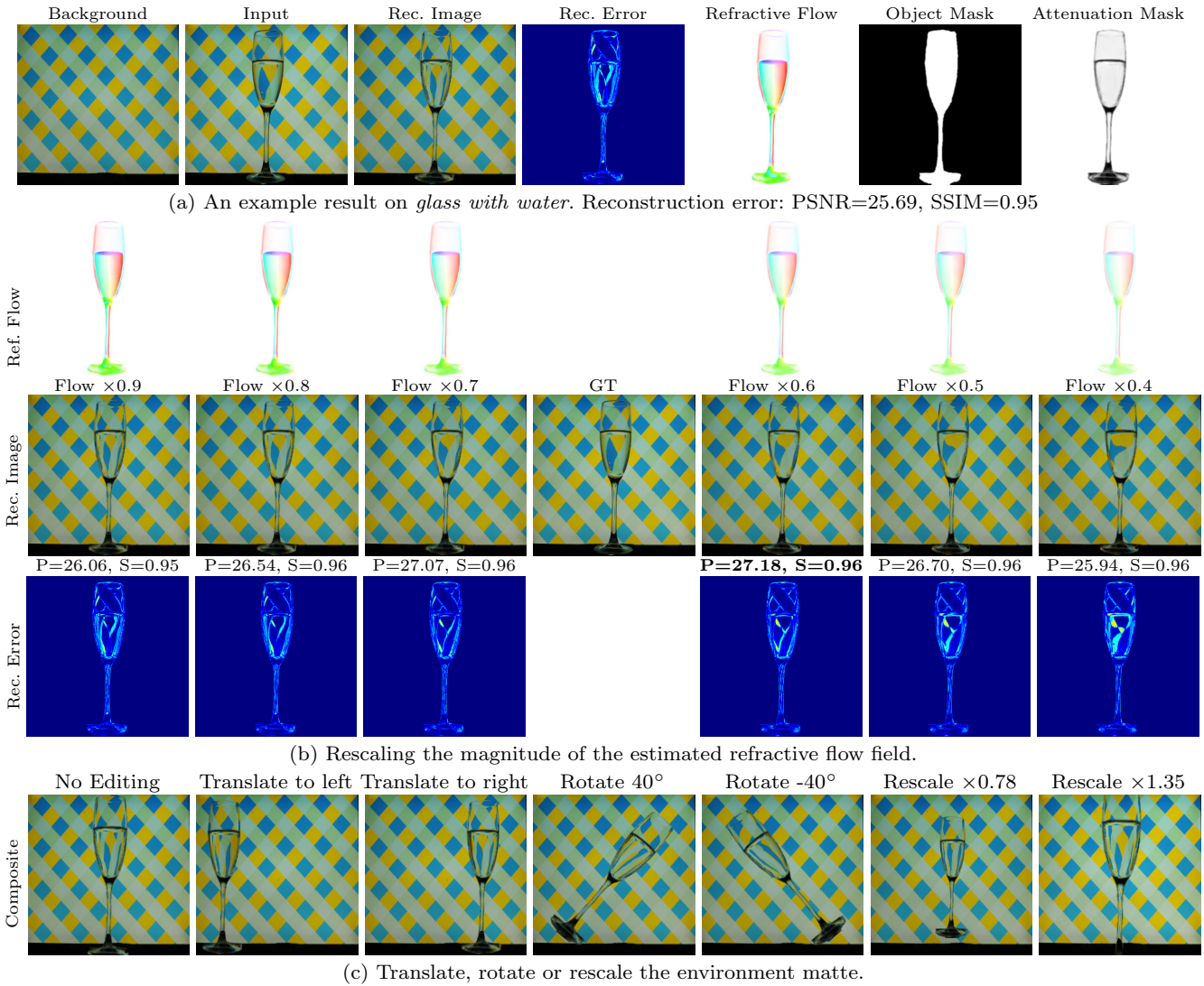
\begin{figure*}[t] \centering
    \makebox[0.135\textwidth]{\scriptsize Background} 
    \makebox[0.135\textwidth]{\scriptsize Input} 
    \makebox[0.135\textwidth]{\scriptsize Rec. Image} 
    \makebox[0.135\textwidth]{\scriptsize Rec. Error} 
    \makebox[0.135\textwidth]{\scriptsize Refractive Flow} 
    \makebox[0.135\textwidth]{\scriptsize Object Mask} 
    \makebox[0.135\textwidth]{\scriptsize Attenuation Mask} 
    \\
    \input{sup_editFlow_example}
    \makebox[\textwidth]{(a) An example result on \emph{glass with water}. Reconstruction error: PSNR=25.69, SSIM=0.95}
    \\
    \input{sup_scaleFlow}
    \vspace{-0.2em}\makebox[\textwidth]{(b) Rescaling the magnitude of the estimated refractive flow field.}
    \\
    \vspace{0.4em}
    \input{sup_editFlow} 
    \makebox[\textwidth]{(c) Translate, rotate or rescale the environment matte.}
    \caption{Various novel composites of a \emph{glass with water} shape obtained by manipulating the predicted environment matte.} \label{fig:sup_edit}
\end{figure*}

\paragraph{\bf User study}
\label{par:User Study}
A user study was carried out to validate the realism of TOM-Net composites. $69$ subjects participated in our user study. At the beginning, we showed each participant photographs of the transparent objects that will be seen during the user study. The objects consisted of $3$ different glasses, $1$ glass with water, $1$ lens, and $1$ complex shape. $40$ samples, including $20$ photographs\footnote{glass $\times$12, glass \& water $\times$4, lens $\times$2, and complex shape $\times$2.} and the corresponding $20$ TOM-Net composites, were then randomly presented to each subject. When showing each sample, we also showed the corresponding background image to the subject for reference. We provided $3$ options for each sample: (P) {\em photograph}, (C) {\em composite}, (N) {\em not distinguishable}.
Tab. \ref{tab:user_study} shows the statistics of the user study. The $69$ participants produced $1,380$ votes for the $20$ real photographs, and $1,380$ votes for the $20$ composites, respectively. The P:C:N ratios are $850:455:75$ and $827:482:71$ for photographs and composites respectively. The per-category ratios also follow a similar trend, indicating close chance of photographs and composites to be considered real, which further demonstrates TOM-Net can produce realistic matte. 

Although we stress that TOM-Net can produce visually realistic composites, the results are still less than perfect. When looking at the real image and our composite side-by-side, users can spot some imperfections of the composite (mostly in the boundary region, see Fig. \ref{fig:sup_user_study}). Therefore, we did not include such a user study by showing the real image and our composite side-by-side. Otherwise, the result will be biased. In the future, we will strengthen our approach to produce more realistic composites, so that the real image and our composite are indistinguishable even when showing them side-by-side.

\begin{table*} \centering
    \caption{Quantitative comparison between TOM-Net, TOM-Net$^{\text{+Trimap}}$ and TOM-Net$^{\text{+Bg}}$ on the synthetic test dataset.}
    \input{syn_quant_trimap_bg}
    \label{tab:stereo_quant_synth}
\end{table*}

\subsection{Transparent Object Editing by Manipulating Environment Matte}
\label{sec:edit_flow}
Given a single image as input, our TOM-Net can estimate the environment matte as a triplet (consisting of an object mask, an attenuation mask and a refractive flow field) in a fast feed-forward pass (see Fig. \ref{fig:sup_edit} (a) for an example).
Note that the goal of the proposed TOM-Net is to extract an environment matte that can produce realistic refractive effect from a single image, instead of estimating highly accurate environment matte. 
The reconstructed image in Fig. \ref{fig:sup_edit} (a) looks realistic but does not have the same refractive effect as the original input, as the refractive effect of the estimated matte seems stronger. 
By decreasing the magnitude of the estimated refractive flow field\footnote{We simply multiply the refractive flow field by a scaling factor ($<1$).}, we can produce a similar refractive effect as the input image (see Fig. \ref{fig:sup_edit} (b)). When the scaling factor becomes $0.6$, the reconstructed image achieves the lowest reconstruction error, with an improvement of $1.49$ and $0.01$ in PSNR and SSIM, respectively.
Apart from rescaling the magnitude of the refractive flow field to adjust the refractive effect of the object, more interesting composites can be obtained by translating, rotating and rescaling the environment matte (see Fig. \ref{fig:sup_edit} (c)). 

\begin{figure} \centering
    \makebox[0.113\textwidth]{\scriptsize Input} 
    \makebox[0.113\textwidth]{\scriptsize Refractive Flow} 
    \makebox[0.113\textwidth]{\scriptsize Object Mask} 
    \makebox[0.113\textwidth]{\scriptsize Attenuation Mask} \\
    \input{sup_failure}
    \caption{Two failure cases in real data. In (a), our model fails to estimate the upper-part of the matte as there is no visual clue to find the object. In (b), the bottom part of the estimated matte is incomplete as the background image is heavily cluttered and the bottom part of the object is very dark.}
    \label{fig:sup_fail}
\end{figure}
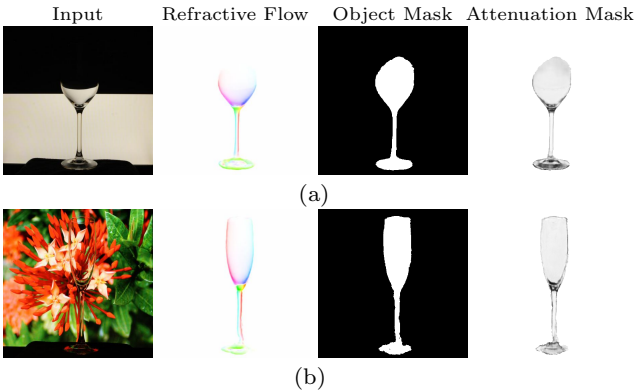

\subsection{Failure Cases}
\label{sec:Failure Cases}
Our model can robustly estimate environment matte for different transparent objects in front of different backgrounds, however, when there is no visual clue for the objects or the image is too cluttered to separate the object from the background, our model may fail. Fig. \ref{fig:sup_fail} shows two failure cases of our model on real data. In Fig. \ref{fig:sup_fail} (a), our model fails to extract the upper-part of the environment matte for the transparent glass due to the lack of visual clue. 
In Fig. \ref{fig:sup_fail} (b), although our model is still able to estimate a reasonable matte, the bottom part of the estimated matte is incomplete due to the very cluttered background. 

\begin{table} \centering
    \caption{Quantitative comparison between TOM-Net, TOM-Net$^{\text{+Trimap}}$ and TOM-Net$^{\text{+Bg}}$ on real data.}
    \resizebox{0.48\textwidth}{!}{ 
    \huge
    \begin{tabular}{c|*{2}{c}|*{2}{c}|*{2}{c}|*{2}{c}|*{2}{c}}
        \toprule
        \multirow{2}{*}{} & \multicolumn{2}{c}{Glass} 
                          & \multicolumn{2}{c}{G \& W} 
                          & \multicolumn{2}{c}{Lens} 
                          & \multicolumn{2}{c}{Complex} 
                          & \multicolumn{2}{c}{Avg} \\
        & PSNR & SSIM  & PSNR & SSIM & PSNR & SSIM & PSNR & SSIM & PSNR & SSIM \\
        \midrule
        Background     & 22.05 & 0.894 & 20.75 & 0.886 & 18.60 & 0.860 & 16.85 & 0.816 & 19.56 & 0.864 \\ 
        TOM-Net        & 25.06 & 0.920 & 23.53 & 0.911 & 20.89 & 0.893 & 17.88 & 0.835 & 21.84 & 0.890 \\ 
        TOM-Net$^{\text{+Trimap}}$ & 25.48 & 0.924 & 23.77 & 0.914 & 23.98 & 0.913 & 20.88 & 0.868 & 23.53 & 0.905\\ 
TOM-Net$^{\text{+Bg}}$     & 26.10 & 0.931 & 24.58 & 0.922 & 25.52 & 0.924 & 22.23 & 0.884 &\textbf{24.61} & \textbf{0.915} \\ 
        \bottomrule
    \end{tabular}
    }
    \label{tab:real_quant_stereo}
\end{table}

\begin{figure*} \centering
    \input{qual_mono_trimap_stereo}
    \caption{Qualitative comparison between TOM-Net, TOM-Net$^{\text{+Trimap}}$ and TOM-Net$^{\text{+Bg}}$ on real data.
    For each testing object, the input to the model is shown on the first two columns, and the results of TOM-Net (up), TOM-Net$^{\text{+Trimap}}$ (middle) and TOM-Net$^{\text{+Bg}}$ (bottom) are shown on the rest of the columns. The PSNR and SSIM between input photographs and reconstructed images are shown right after the error maps.}
    \label{fig:qual_stereo_mono}
\end{figure*}
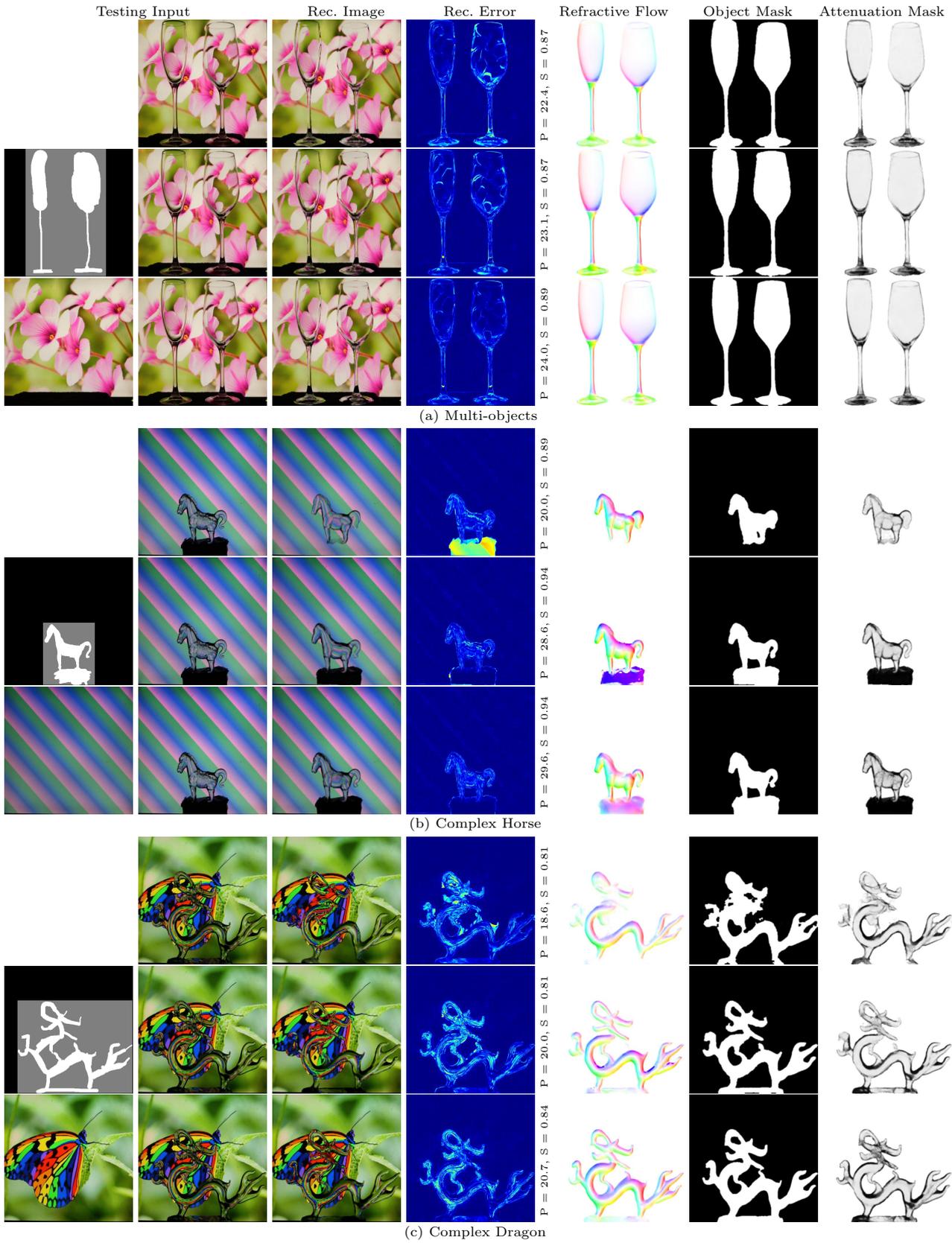

\begin{figure*}[t] \centering
    \input{limitation}
    \caption{Qualitative results of TOM-Net on colored transparent object (first row) and objects under natural illumination (last four rows).}
    \label{fig:limitation}
\end{figure*}
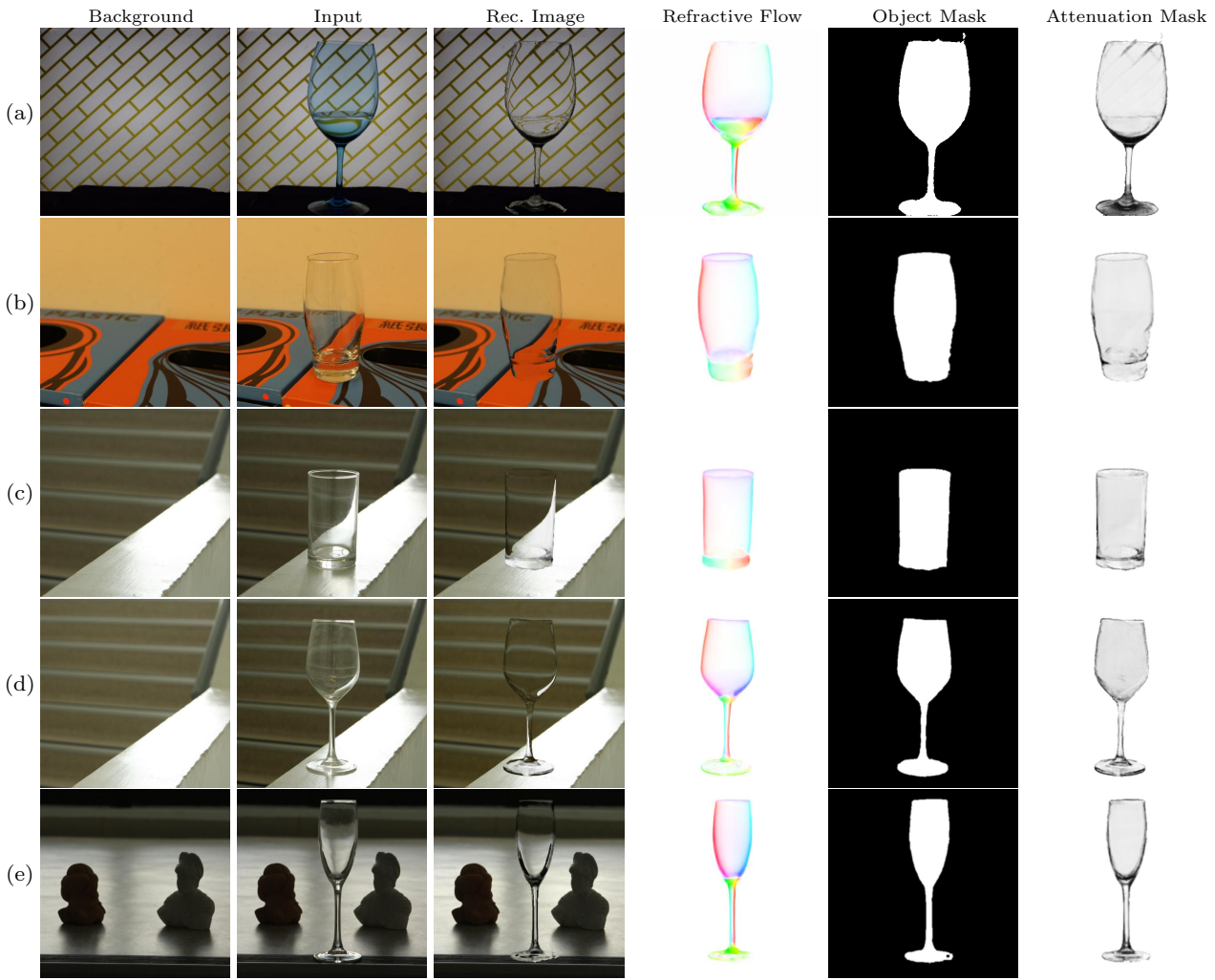

\subsection{Improvement with Trimap and Background Image}
At test time, the input trimaps for TOM-Net$^{\text{+Trimap}}$ were generated in the same way adopted in the training (as described in Subsection \ref{ssec:trimap}), except that the foreground regions were obtained by performing erosion operation on the ground-truth object mask with a fixed (rather than a random) kernel size of $10$ pixels for evaluation.
Tab. \ref{tab:stereo_quant_synth} shows the quantitative comparisons between TOM-Net, TOM-Net$^{\text{+Trimap}}$ and TOM-Net$^{\text{+Bg}}$ on the synthetic test dataset. 
As expected, with the access to the additional information, both TOM-Net$^{\text{+Trimap}}$ and TOM-Net$^{\text{+Bg}}$ performed better than TOM-Net. Due to the fact that a background image contains more useful information than a trimap, TOM-Net$^{\text{+Bg}}$ achieved the best results.

Tab. \ref{tab:real_quant_stereo} presents the quantitative comparison on real data. Compared with TOM-Net, TOM-Net$^{\text{+Trimap}}$ and TOM-Net$^{\text{+Bg}}$ achieved an improvement of $1.69$ and $2.77$ in average PSNR and an improvement of $0.015$ and $0.024$ in average SSIM, respectively. 
Fig. \ref{fig:qual_stereo_mono} shows the qualitative comparison on real data, where the foreground region of the trimap was marked by the user. It can be seen that with the additional information, TOM-Net$^{\text{+Trimap}}$ and TOM-Net$^{\text{+Bg}}$ can identify the transparent object from the cluttered background more accurately than TOM-Net and model the opaque base of the transparent object (Fig. \ref{fig:qual_stereo_mono} (b)). As a result, the environment matte predicted by TOM-Net$^{\text{+Trimap}}$ and TOM-Net$^{\text{+Bg}}$ can produce more realistic composites and achieve lower reconstruction errors, clearly demonstrating the effectiveness of our framework in handling cases where a trimap or a background image is available.

\section{Discussion}
\label{sec:discussion}
\subsection{Limitations}
Although our method can produce plausible results for transparent object matting, there do exist limitations that require further study.
First, our model assumes objects to be colorless so that the attenuation property of an object can be depicted as a scalar value $\rho$ in our formulation. However, this is not applicable to colored transparent objects, as shown in see Fig.~\ref{fig:limitation}~(a). 
Although our method can estimate a reasonably good object mask and refractive flow field for the glass with water, the estimated attenuation mask cannot model the colored effect of the object.

Second, our model assumes a single planar background (following most of the previous works) as the only light source and simplifies the interaction between object and background image to a point-to-point (single) mapping. However, more complicated effects exist in the real world, such as specular highlights, translucent, multi-mapping (i.e., refraction and reflection happen simultaneously at a surface point), and color dispersion (i.e., different color components may have different supporting background regions). 
Fig.~\ref{fig:limitation}~(b)-(e) show four example results of TOM-Net on transparent objects under different types of natural illuminations. Regardless of the fact that TOM-Net can estimate a plausible object mask and refractive flow field, the composites do not look very realistic. 
This is because our current formulation does not consider the more sophisticated refractive properties of a transparent object under natural illumination like complex interaction with environment lighting, specular highlight, Fresnel effect, and acoustic shadow.

\subsection{Colored Objects and Specular Highlights}
Here we sketch the potential solutions to colored transparent objects as well as the cases when specular highlights appear on transparent objects. In Section \ref{sec:formulation}, we simplified matting equation as (\ref{eq:em_simplify3}).
To handle colored objects, the scalar attenuation index $\rho$ should be expanded to a color attenuation 3-vector $R$, in which each value corresponds to an attenuation index for a specific color channel. The matting equation then becomes 
\begin{equation}
    \label{eq:color_2}
    C = (1 - m) B + mR \circ \mathcal{M}(\mathbf{T}, P),
\end{equation}
where $\circ$ represents element-wise multiplication.

Consider a white near point light source, we can simplify the specular highlight effect with a specular highlight component $S$, then the generalized matting equation can be written as
\begin{equation}
    \label{eq:color_2}
    C = (1 - m) B + mR \circ \mathcal{M}(\mathbf{T}, P) + S,
\end{equation}
where $S$ is a 3-vector containing three identical values.
The problem of transparent object matting now becomes simultaneously estimating an object mask, a color attenuation mask, a refractive flow field and a specular highlight mask from a single image, while more efforts are needed to implement them for practical use and we leave this as our future work.

\subsection{Difficulty in Comparison with Previous Works}
Currently, it is not trivial to have a fair comparison with existing methods. On one hand, applying our method on the data used in the previous methods is difficult. Most of the previous methods require multiple images of the transparent object captured in front of pre-designed patterns, which are not publicly available and lack enough textures for our method to estimate the refractive effect of the transparent object. 
The single image based methods RTCEM \cite{chuang2000environment} and \cite{yeung2011matting} have additional requirements. In particular, RTCEM \cite{chuang2000environment} requires the object to be captured in front of a coded-pattern (also not publicly available), and the background image is needed to segment the foreground object. 
\cite{yeung2011matting} requires human interaction to segment the foreground object and model the object's refractive effect with thin-plate-spline transformation. The data used in \cite{yeung2011matting} does not follow our assumption that the light comes from a single background image, thus it cannot be directly processed by our method.
On the other hand, there are no public implementations for the previous methods, and even if there were, those methods cannot be applied to our dataset which is created for single image transparent object matting.

Different from the previous methods, our method aims to estimate the foreground mask, attenuation mask and refractive flow field from a single natural image. Since our code and datasets have been made publicly available, it will ease the comparison for the following work. We believe our work can serve as a baseline and provide meaningful insight for future researches in this area.
\section{Conclusion}
\label{sec:conclusion}
We have introduced a simple and efficient model for transparent object matting, and proposed a CNN architecture, called TOM-Net, that takes a single image as input and predicts environment matte as an object mask, an attenuation mask, and a refractive flow field in a fast feed-forward pass. Besides, we created a large-scale synthetic dataset and a real dataset as a benchmark for learning transparent object matting. We have also shown that TOM-Net can perform better by incorporating a trimap or a background image in the input. Promising results have been achieved on both synthetic and real data, which clearly demonstrate the feasibility and effectiveness of the proposed approach. We consider exploring better models and architectures for transparent object matting as our future work.

\paragraph{\bf Acknowledgments}
This project is supported by a grant from the Research Grant Council of the Hong Kong (SAR), China, under the project HKU 718113E. We gratefully acknowledge the support of NVIDIA Corporation with the donation of the Titan X Pascal GPU used for this research.

\bibliographystyle{spmpsci}      
\bibliography{gychen}   

\end{document}

%% file: syn_data_samples.tex
     \makebox[0.075\textwidth]{\scriptsize Bg} 
     \makebox[0.075\textwidth]{\scriptsize Image} 
     \makebox[0.075\textwidth]{\scriptsize Flow Vis} 
     \makebox[0.075\textwidth]{\scriptsize Ref. Flow} 
     \makebox[0.075\textwidth]{\scriptsize Obj. Mask} 
     \makebox[0.075\textwidth]{\scriptsize Att. Mask} 
     \\
     \includegraphics[width=0.075\textwidth]{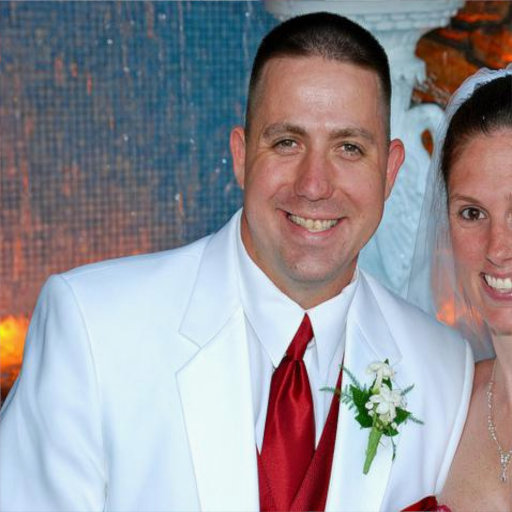}
     \includegraphics[width=0.075\textwidth]{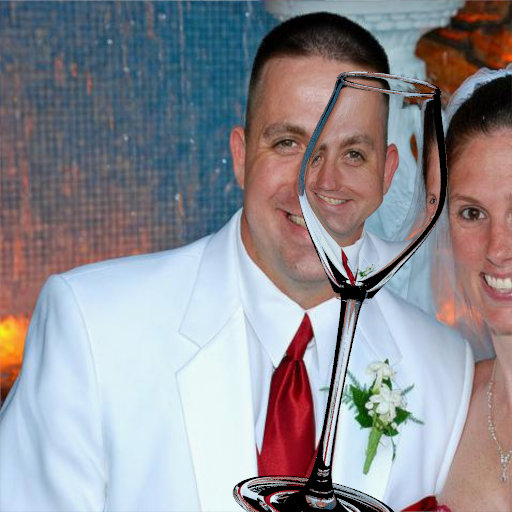}
     \includegraphics[width=0.075\textwidth]{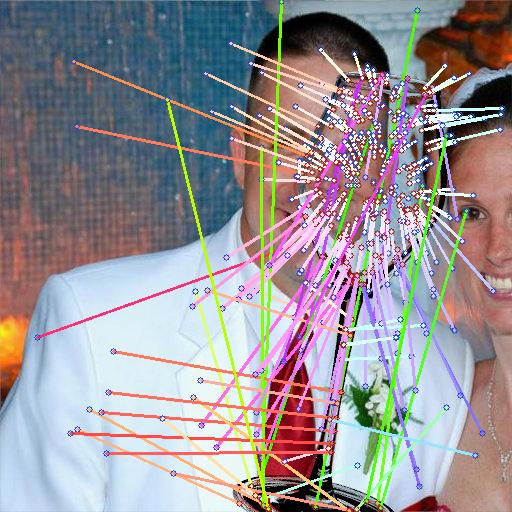}
     \includegraphics[width=0.075\textwidth]{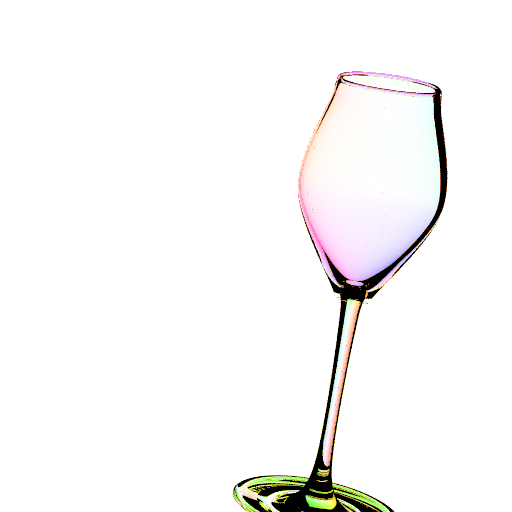}
     \includegraphics[width=0.075\textwidth]{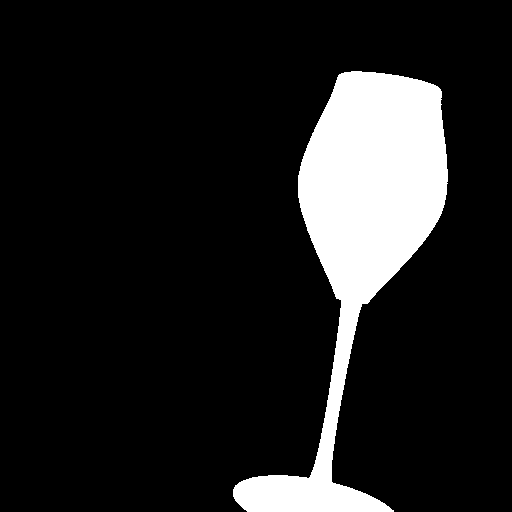}
     \includegraphics[width=0.075\textwidth]{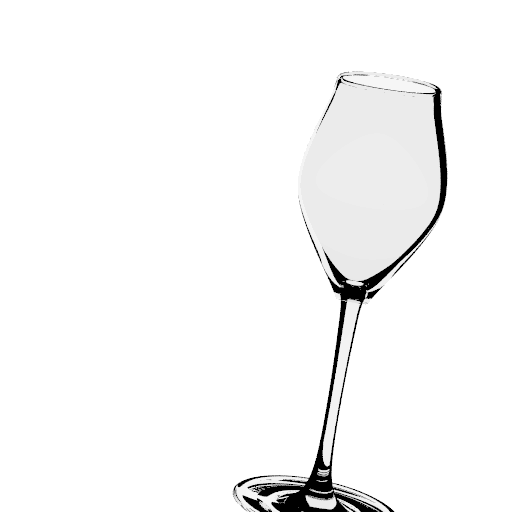}
     \\
     \includegraphics[width=0.075\textwidth]{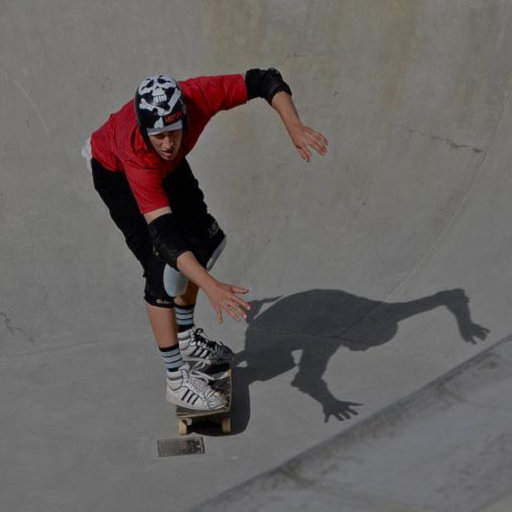}
     \includegraphics[width=0.075\textwidth]{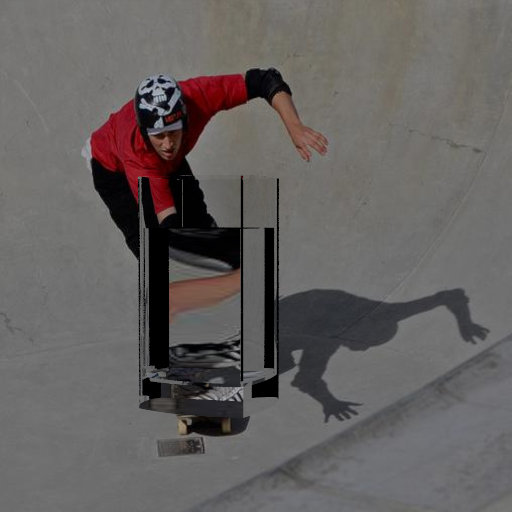}
     \includegraphics[width=0.075\textwidth]{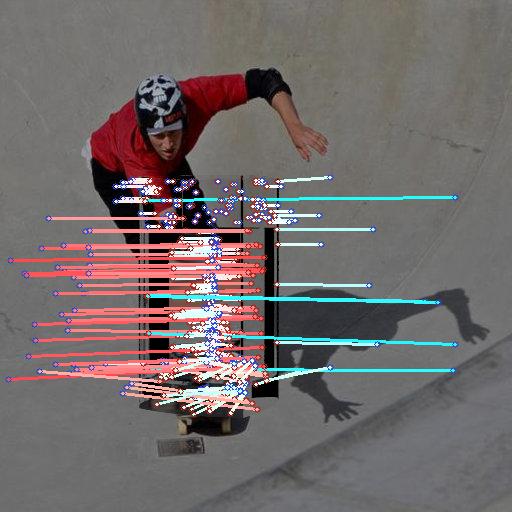}
     \includegraphics[width=0.075\textwidth]{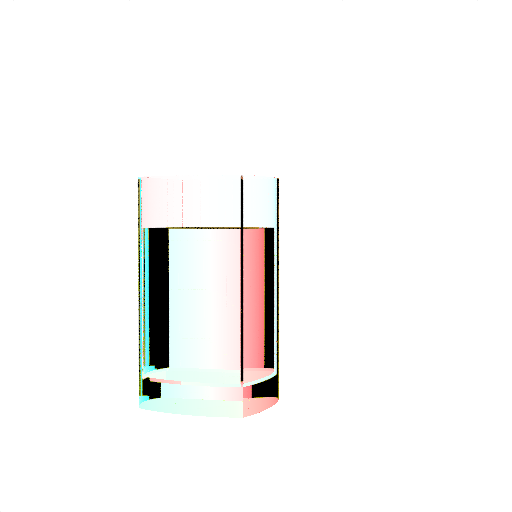}
     \includegraphics[width=0.075\textwidth]{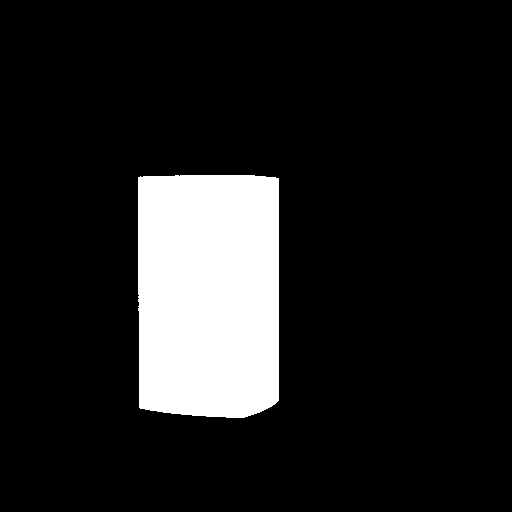}
     \includegraphics[width=0.075\textwidth]{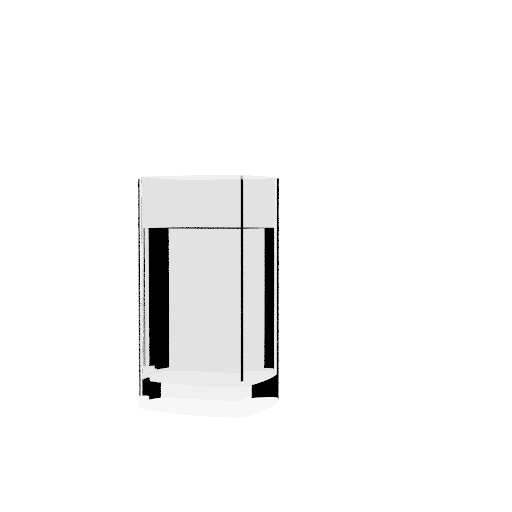}
     \\
     \includegraphics[width=0.075\textwidth]{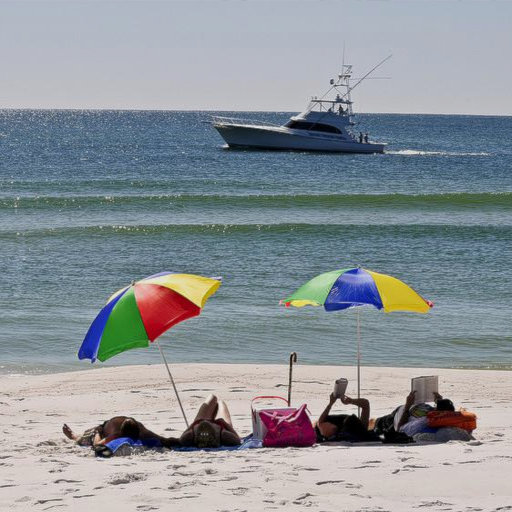}
     \includegraphics[width=0.075\textwidth]{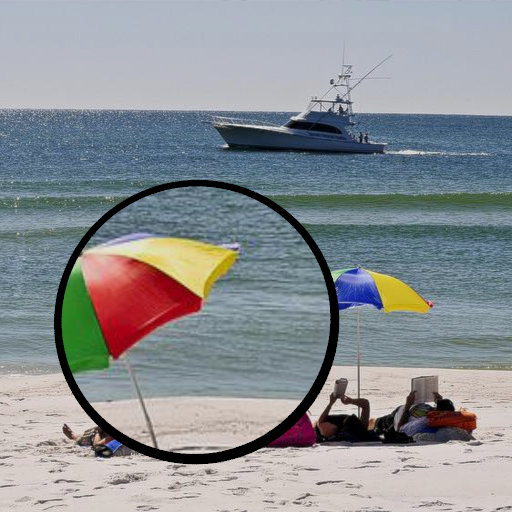}
     \includegraphics[width=0.075\textwidth]{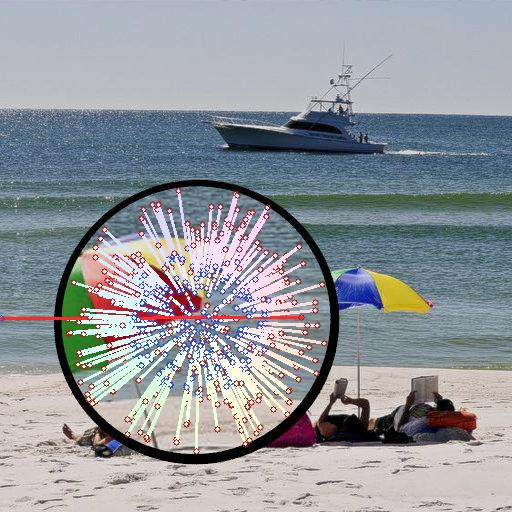}
     \includegraphics[width=0.075\textwidth]{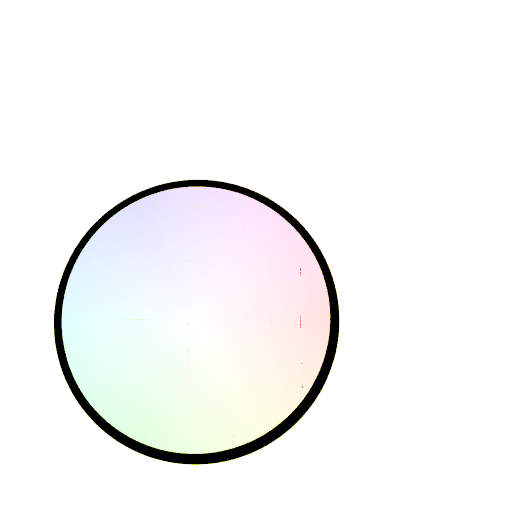}
     \includegraphics[width=0.075\textwidth]{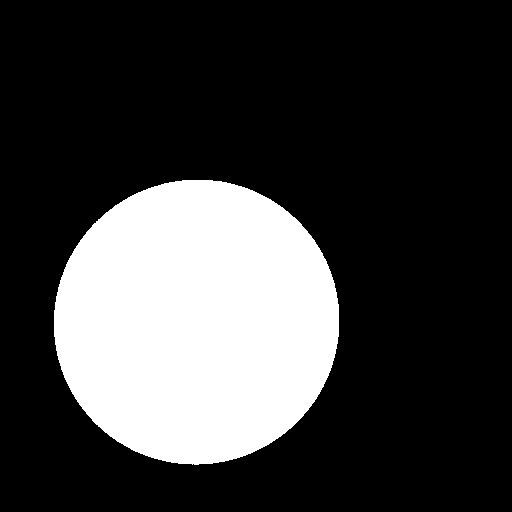}
     \includegraphics[width=0.075\textwidth]{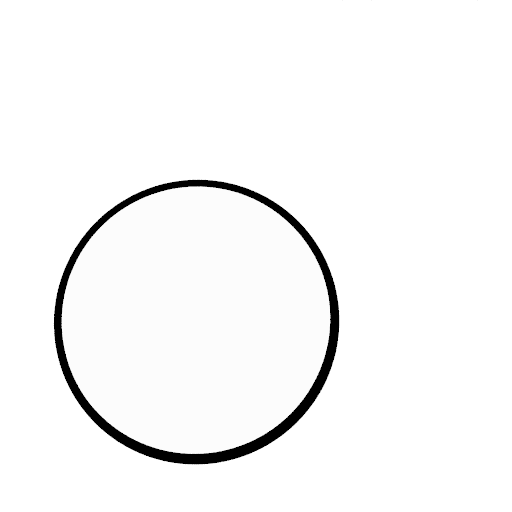}
     \\
     \includegraphics[width=0.075\textwidth]{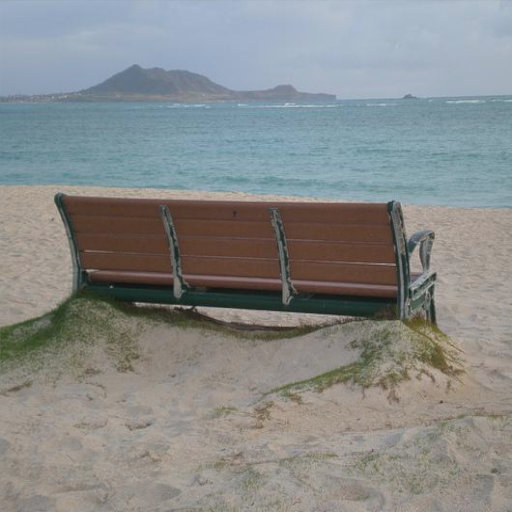}
     \includegraphics[width=0.075\textwidth]{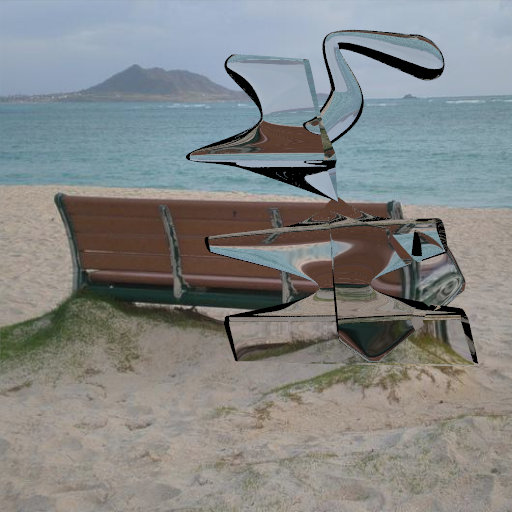}
     \includegraphics[width=0.075\textwidth]{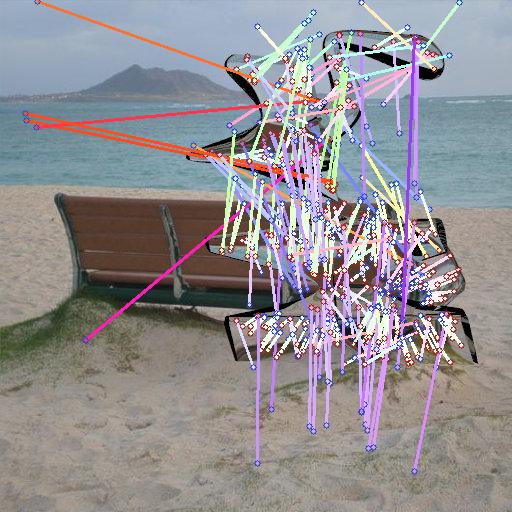}
     \includegraphics[width=0.075\textwidth]{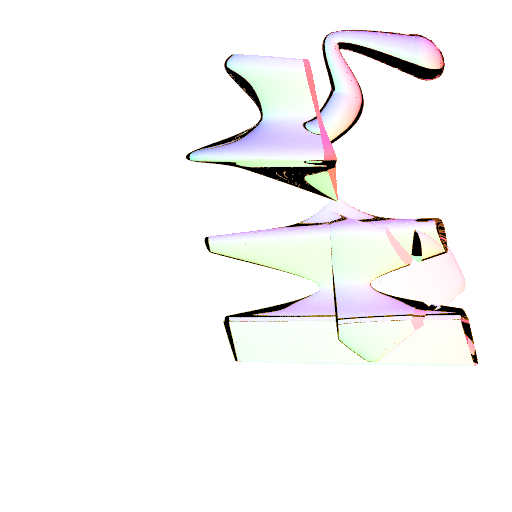}
     \includegraphics[width=0.075\textwidth]{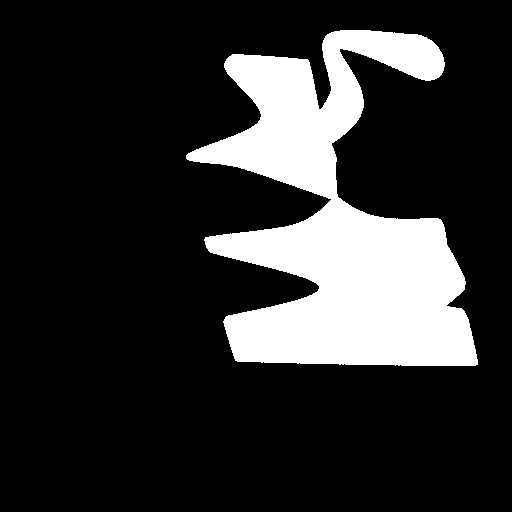}
     \includegraphics[width=0.075\textwidth]{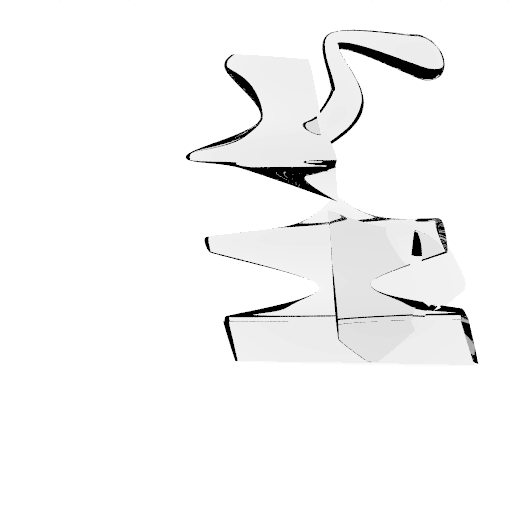}

%% file: real_data_samples.tex
    \makebox[0.092\textwidth]{\scriptsize Glass} 
    \makebox[0.092\textwidth]{\scriptsize Glass \& Water} 
    \makebox[0.092\textwidth]{\scriptsize Lens} 
    \makebox[0.092\textwidth]{\scriptsize Complex}
    \makebox[0.092\textwidth]{\scriptsize Complex}
    \\
    \includegraphics[width=0.092\textwidth]{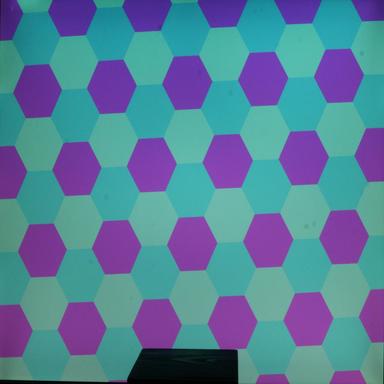}
    \includegraphics[width=0.092\textwidth]{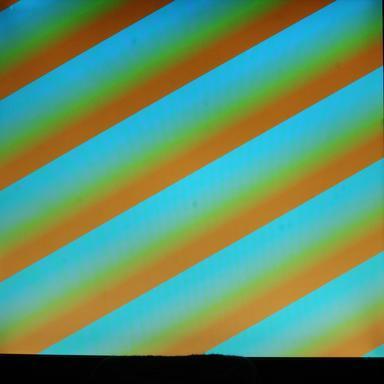}
    \includegraphics[width=0.092\textwidth]{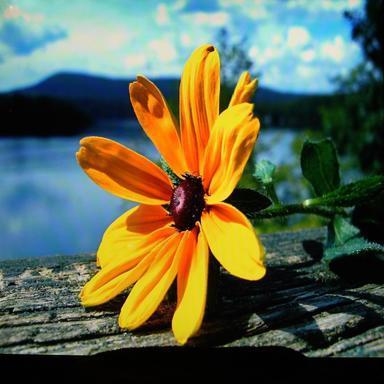}
    \includegraphics[width=0.092\textwidth]{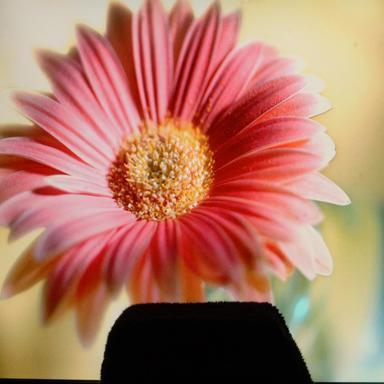}
    \includegraphics[width=0.092\textwidth]{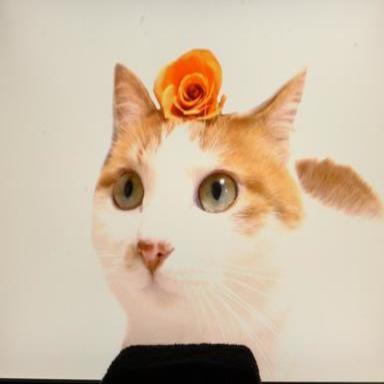}
    \\
    \includegraphics[width=0.092\textwidth]{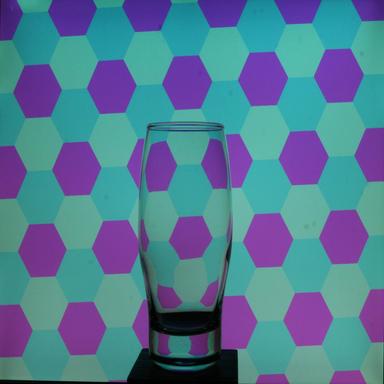}
    \includegraphics[width=0.092\textwidth]{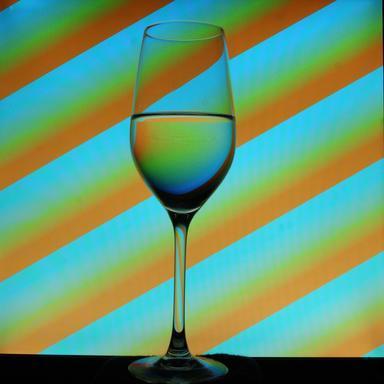}
    \includegraphics[width=0.092\textwidth]{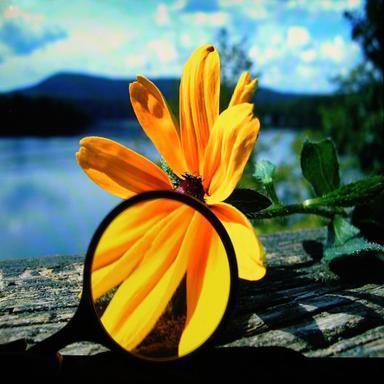}
    \includegraphics[width=0.092\textwidth]{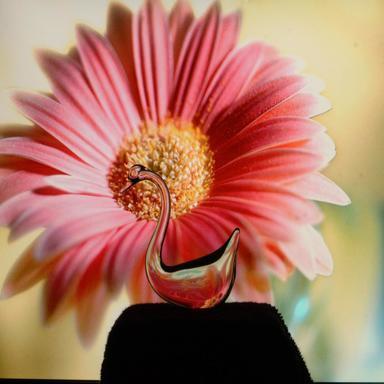}
    \includegraphics[width=0.092\textwidth]{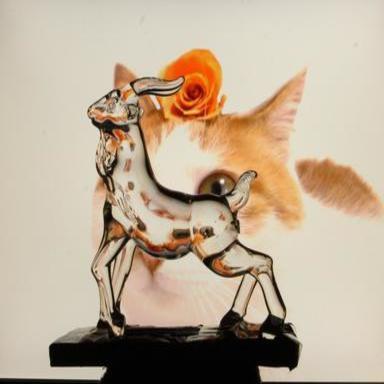}
    \\

%% file: syn_ablation_study.tex
    \makebox[0.162\textwidth]{\footnotesize Background} 
    \makebox[0.162\textwidth]{\footnotesize Input} 
    \makebox[0.162\textwidth]{\footnotesize GT Flow} 
    \makebox[0.162\textwidth]{\footnotesize GT Mask} 
    \makebox[0.162\textwidth]{\footnotesize GT Attenuation} 
    \\
    \includegraphics[width=0.162\textwidth]{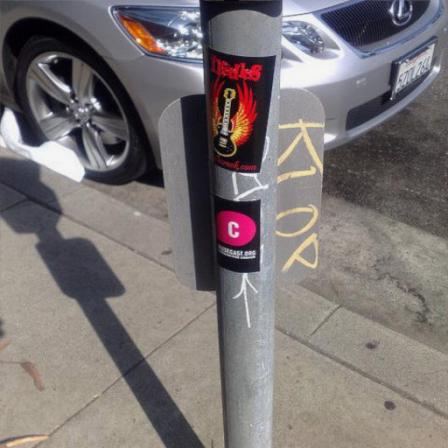}
    \includegraphics[width=0.162\textwidth]{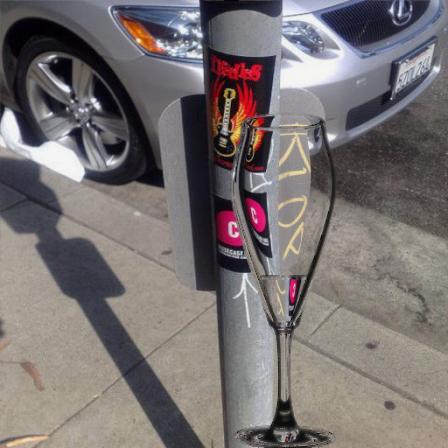}
    \includegraphics[width=0.162\textwidth]{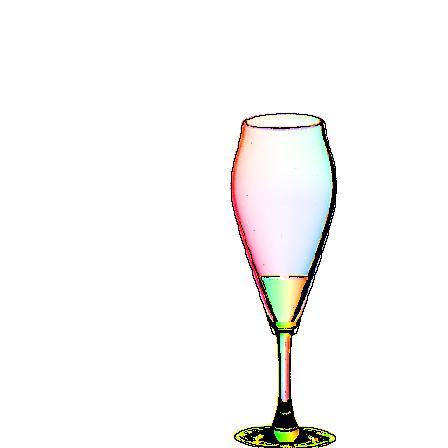}
    \includegraphics[width=0.162\textwidth]{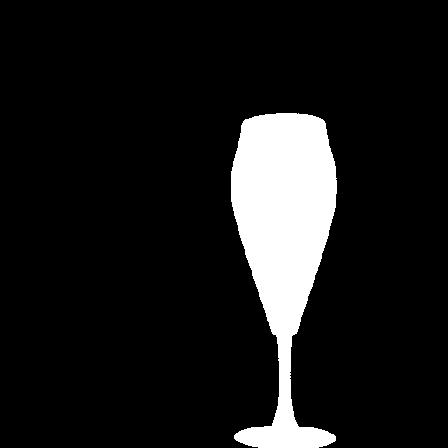}
    \includegraphics[width=0.162\textwidth]{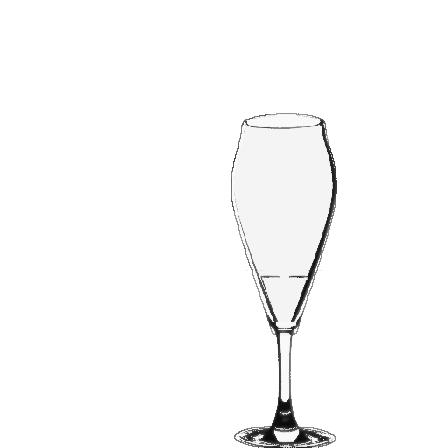}
    \\ \vspace{-0.4em}\ \hrule\  \\
    \makebox[0.162\textwidth]{\footnotesize (a) \textbf{CNet} - ($\mathcal{L}^c_{fr}$)} 
    \makebox[0.162\textwidth]{\footnotesize (b) \textbf{CNet} - (c-l)} 
    \makebox[0.162\textwidth]{\footnotesize (c) \textbf{CNet} - (m-s)} 
    \makebox[0.162\textwidth]{\footnotesize (d) \textbf{CNet} - ($\mathcal{L}^c_{ir}$)} 
    \makebox[0.162\textwidth]{\footnotesize (e) \textbf{CNet}} 
    \makebox[0.162\textwidth]{\footnotesize (f) \textbf{CNet} + \textbf{RNet}} 
    \\
    \includegraphics[width=0.162\textwidth]{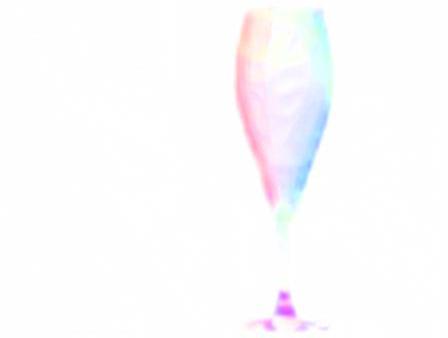}
    \includegraphics[width=0.162\textwidth]{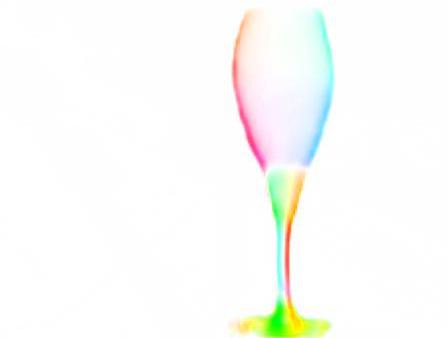}
    \includegraphics[width=0.162\textwidth]{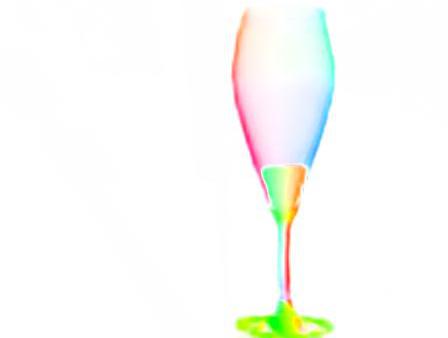}
    \includegraphics[width=0.162\textwidth]{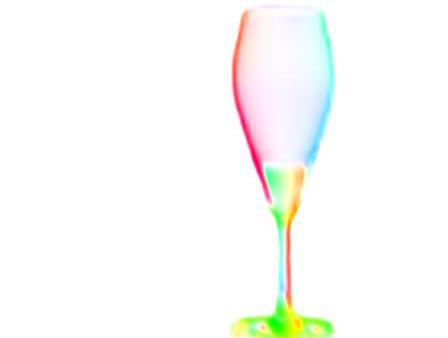}
    \includegraphics[width=0.162\textwidth]{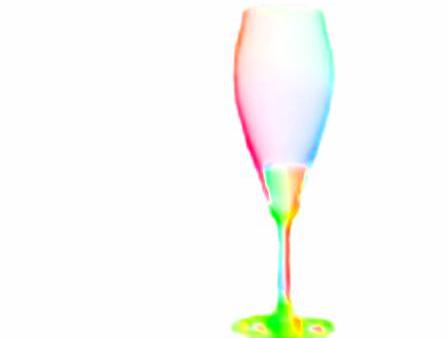}
    \includegraphics[width=0.162\textwidth]{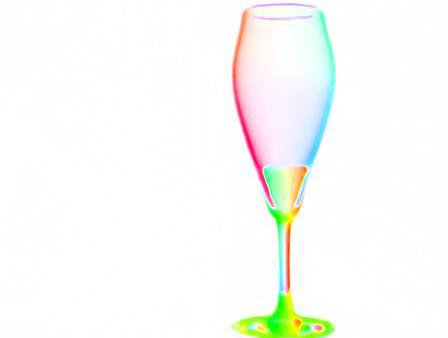}
    \\
    \vspace{-0.3em}
    \makebox[0.162\textwidth]{\tiny F-EPE = $4.5 / 39.7$} 
    \makebox[0.162\textwidth]{\tiny F-EPE = $2.9 / 24.5$} 
    \makebox[0.162\textwidth]{\tiny F-EPE = $2.5 / 20.7$} 
    \makebox[0.162\textwidth]{\tiny F-EPE = $2.6 / 21.4$} 
    \makebox[0.162\textwidth]{\tiny F-EPE = $2.4 / 19.4$} 
    \makebox[0.162\textwidth]{\tiny \textbf{F-EPE =} $\mathbf{2.1 / 16.9}$} 
    \\
    \includegraphics[width=0.162\textwidth]{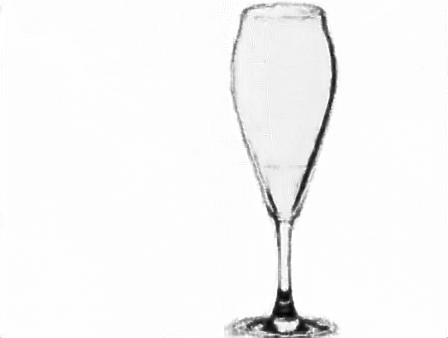}
    \includegraphics[width=0.162\textwidth]{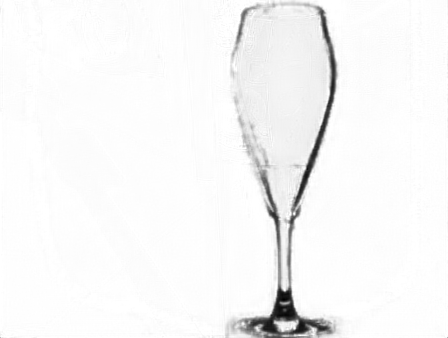}
    \includegraphics[width=0.162\textwidth]{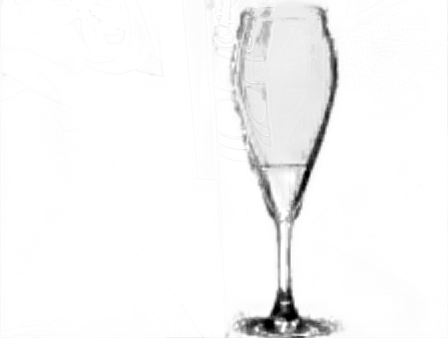}
    \includegraphics[width=0.162\textwidth]{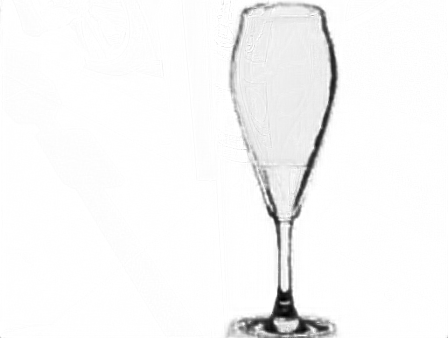}
    \includegraphics[width=0.162\textwidth]{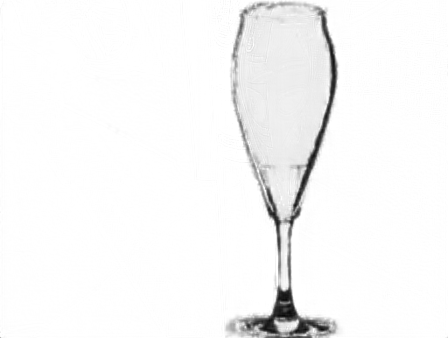}
    \includegraphics[width=0.162\textwidth]{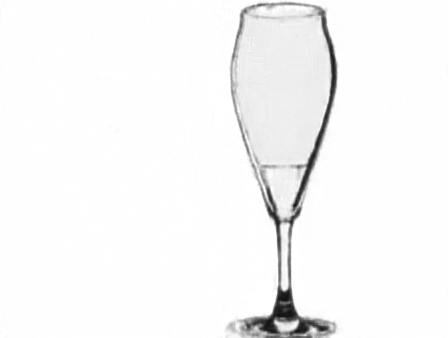}
    \\
    \vspace{-0.3em}
    \makebox[0.162\textwidth]{\tiny A-MSE = $0.19\times 10^{-2}$} 
    \makebox[0.162\textwidth]{\tiny A-MSE = $0.23\times 10^{-2}$} 
    \makebox[0.162\textwidth]{\tiny A-MSE = $0.32\times 10^{-2}$} 
    \makebox[0.162\textwidth]{\tiny A-MSE = $0.21\times 10^{-2}$} 
    \makebox[0.162\textwidth]{\tiny A-MSE = $0.22\times 10^{-2}$} 
    \makebox[0.162\textwidth]{\tiny \textbf{A-MSE =} $\mathbf{0.19\times 10^{-2}}$} 
    \\

%% file: refine_vis.tex
    \makebox[0.09\textwidth]{\scriptsize Input} 
    \makebox[0.09\textwidth]{\scriptsize Coarse Flow} 
    \makebox[0.09\textwidth]{\scriptsize Refined Flow} 
    \makebox[0.09\textwidth]{\scriptsize Coarse Att.} 
    \makebox[0.09\textwidth]{\scriptsize Refined Att.} 
    \\
    \includegraphics[width=0.090\textwidth]{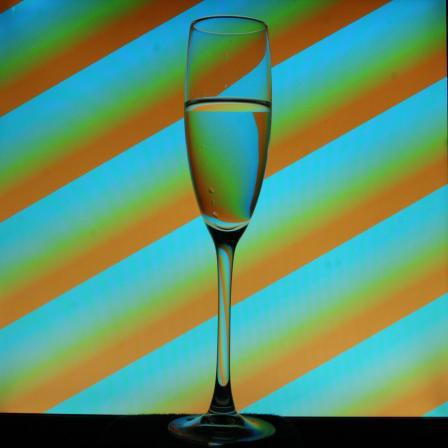}
    \includegraphics[width=0.090\textwidth]{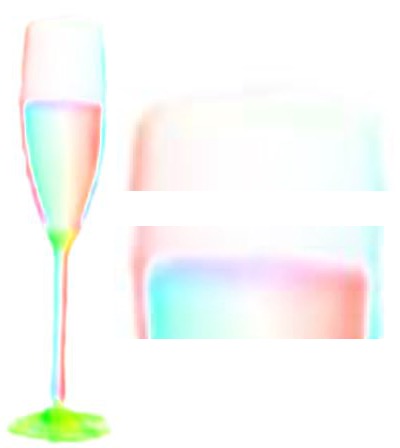}
    \includegraphics[width=0.090\textwidth]{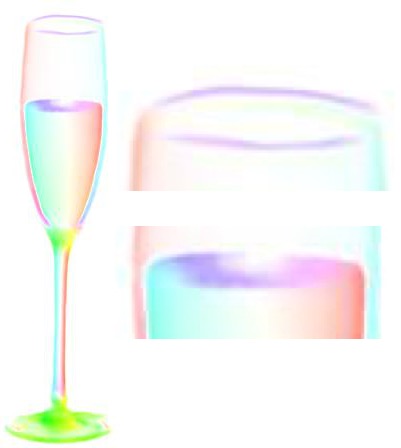}
    \includegraphics[width=0.090\textwidth]{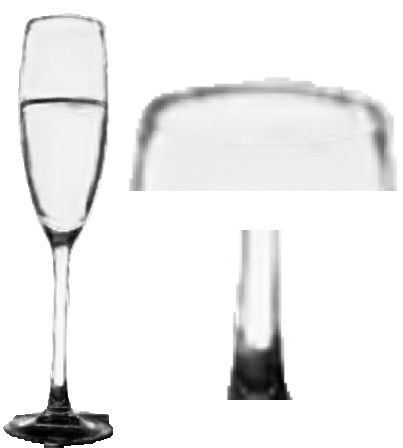}
    \includegraphics[width=0.090\textwidth]{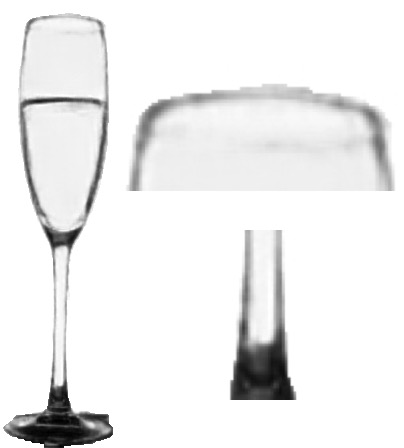}
    \\
    \includegraphics[width=0.090\textwidth]{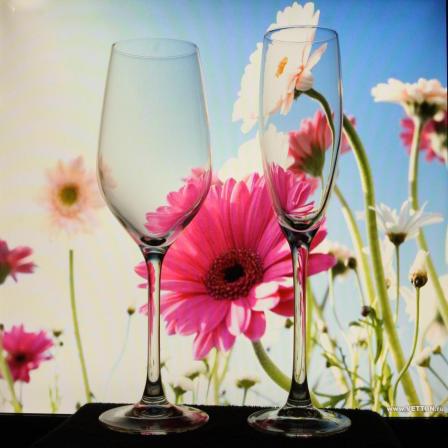}
    \includegraphics[width=0.090\textwidth]{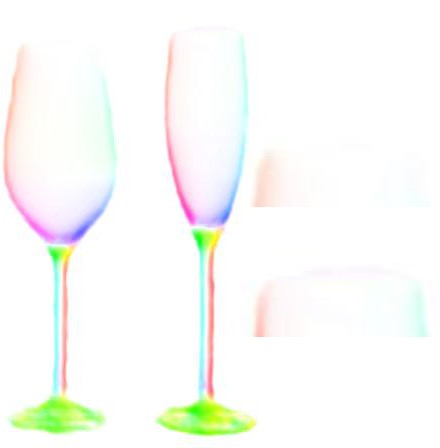}
    \includegraphics[width=0.090\textwidth]{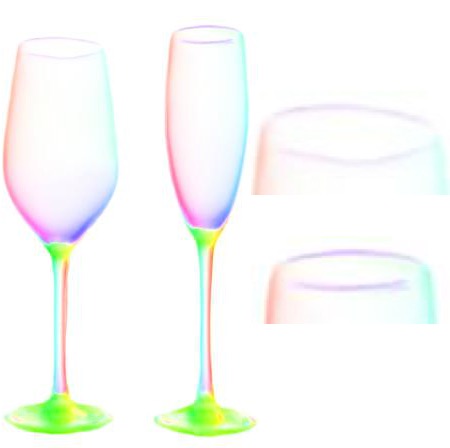}
    \includegraphics[width=0.090\textwidth]{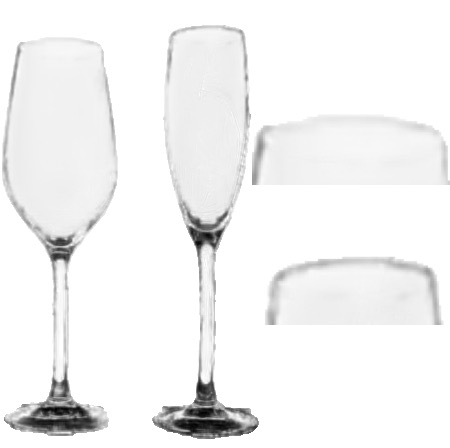}
    \includegraphics[width=0.090\textwidth]{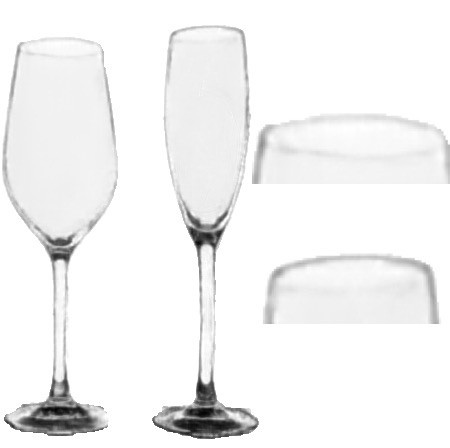}
    \\

%% file: syn_quant.tex
    \begin{minipage}{0.93\textwidth}
    \resizebox{\textwidth}{!}{
    \Huge
        \begin{tabular}{c|*{4}{c}|*{4}{c}|*{4}{c}|*{4}{c}|*{4}{c}}
        \toprule
        \multirow{2}{*}{} & \multicolumn{4}{c}{Glass} 
                               & \multicolumn{4}{c}{Glass with Water} 
                               & \multicolumn{4}{c}{Lens} 
                               & \multicolumn{4}{c}{Complex Shape}  
                               & \multicolumn{4}{c}{Average}  \\
                               & \cellcolor{red!25} F-EPE & \cellcolor{red!25}A-MSE & \cellcolor{red!25}I-MSE & \cellcolor{blue!25} M-IoU 
                               & \cellcolor{red!25} F-EPE & \cellcolor{red!25}A-MSE & \cellcolor{red!25}I-MSE & \cellcolor{blue!25} M-IoU 
                               & \cellcolor{red!25} F-EPE & \cellcolor{red!25}A-MSE & \cellcolor{red!25}I-MSE & \cellcolor{blue!25} M-IoU 
                               & \cellcolor{red!25} F-EPE & \cellcolor{red!25}A-MSE & \cellcolor{red!25}I-MSE & \cellcolor{blue!25} M-IoU 
                               & \cellcolor{red!25} F-EPE & \cellcolor{red!25}A-MSE & \cellcolor{red!25}I-MSE & \cellcolor{blue!25} M-IoU \\
        \midrule
        Background    & 3.6 / 30.3 & 1.33 & 0.48 & 0.12 & 6.4 / 53.2 & 1.54 & 0.68 & 0.12 & 10.3 / 39.2 & 1.94 & 1.57 & 0.24 & 6.8 / 56.8 & 2.50 & 0.85 & 0.11 & 6.8 / 44.9 & 1.83 & 0.90 & 0.15 \\
        CoarseNet     & 2.1 / 15.8 & 0.22 & 0.14 & 0.97 & 3.1 / 23.5 & 0.31 & 0.23 & 0.97 & 2.0 / 6.7   & 0.17 & 0.28 & 0.99 & 4.5 / 34.4 & 0.38 & 0.33 & 0.92 & 2.9 / 20.1 & 0.27 & 0.24 & 0.96 \\
        \midrule
        TOM-Net       & 1.9 / 14.7 & 0.21 & 0.14 & 0.97 & 2.9 / 21.8 & 0.30 & 0.22 & 0.97 & 1.9 / 6.6   & 0.15 & 0.29 & 0.99 & 4.1 / 31.5 & 0.37 & 0.32 & 0.92 & 2.7 / 18.6 & 0.26 & 0.24 & 0.96 \\
        \bottomrule
    \end{tabular}
    }
    \end{minipage}
    \hspace{-0.6em}
    \begin{minipage}{0.06\textwidth}
        \resizebox{\textwidth}{!}{
        \begin{tabular}{cc}
            \midrule
            MSE ($\cdot10^{-2}$) \\
            \cellcolor{red!25} $\downarrow$ better \\ 
            \cellcolor{blue!25} $\uparrow$ better\\
            \midrule
        \end{tabular}
        }
    \end{minipage}

%% file: syn_qual.tex
    \makebox[0.095\textwidth]{\footnotesize Background} 
    \makebox[0.095\textwidth]{\footnotesize Input} 
    \makebox[0.095\textwidth]{\footnotesize Rec. Image} 
    \makebox[0.095\textwidth]{\footnotesize Rec. Error} 
    \makebox[0.195\textwidth]{\footnotesize Ref. Flow (GT / Est.)} 
    \makebox[0.195\textwidth]{\footnotesize Obj. Mask (GT / Est.)} 
    \makebox[0.195\textwidth]{\footnotesize Att. Mask (GT / Est.)} 
    \\
    \includegraphics[width=0.095\textwidth]{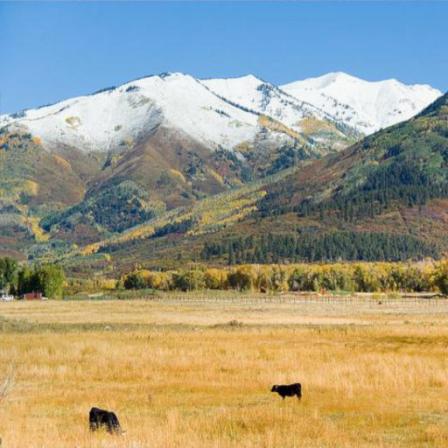}
    \includegraphics[width=0.095\textwidth]{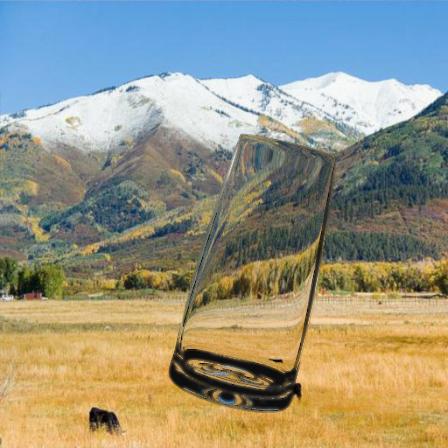}
    \includegraphics[width=0.095\textwidth]{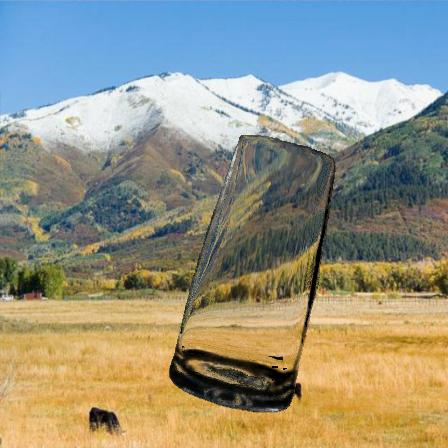}
    \includegraphics[width=0.095\textwidth]{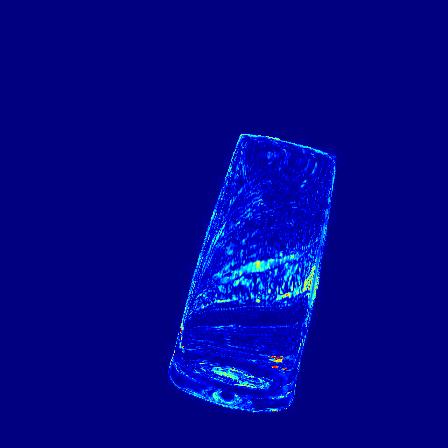}
    \includegraphics[width=0.095\textwidth]{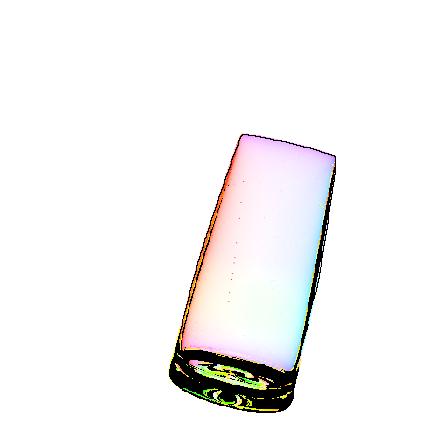}
    \includegraphics[width=0.095\textwidth]{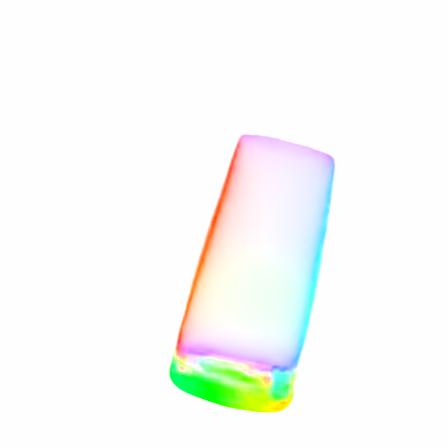}
    \includegraphics[width=0.095\textwidth]{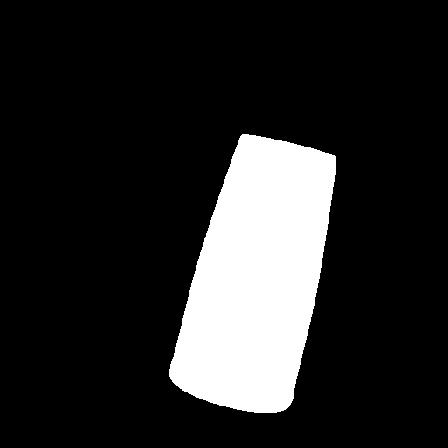}
    \includegraphics[width=0.095\textwidth]{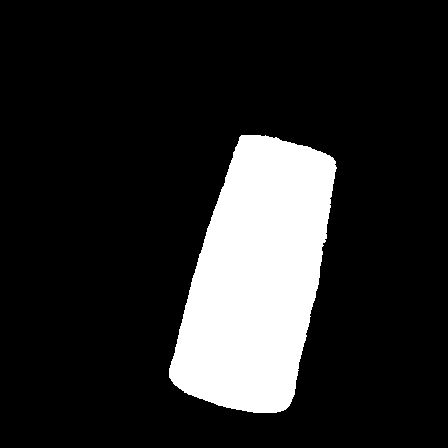}
    \includegraphics[width=0.095\textwidth]{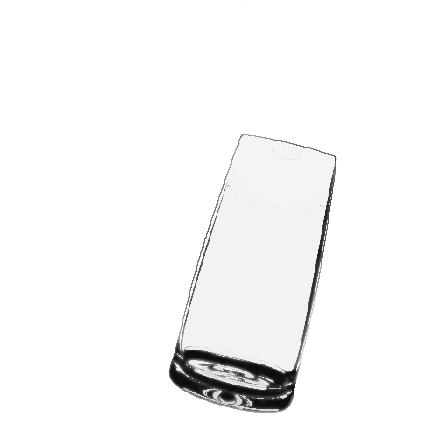}
    \includegraphics[width=0.095\textwidth]{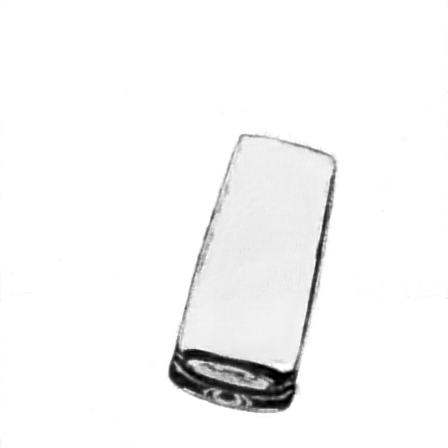}
    \\
    \vspace{-0.3em}\makebox[0.38\textwidth]{\scriptsize (a) Glass, I-MSE = $0.21 \times 10^{-2}$} 
    \makebox[0.19\textwidth]{\scriptsize F-EPE = 2.6 / 15.0} 
    \makebox[0.19\textwidth]{\scriptsize M-IoU = 0.99} 
    \makebox[0.19\textwidth]{\scriptsize A-MSE = $0.16 \times 10^{-2}$} 
    \\
    \includegraphics[width=0.095\textwidth]{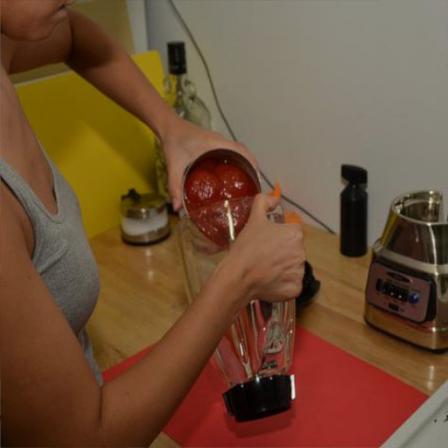}
    \includegraphics[width=0.095\textwidth]{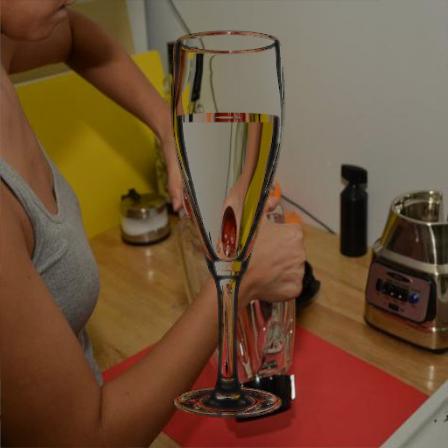}
    \includegraphics[width=0.095\textwidth]{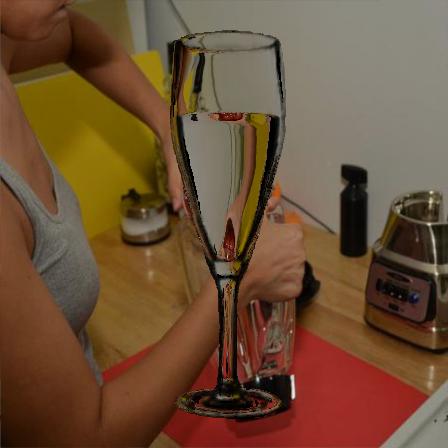}
   \includegraphics[width=0.095\textwidth]{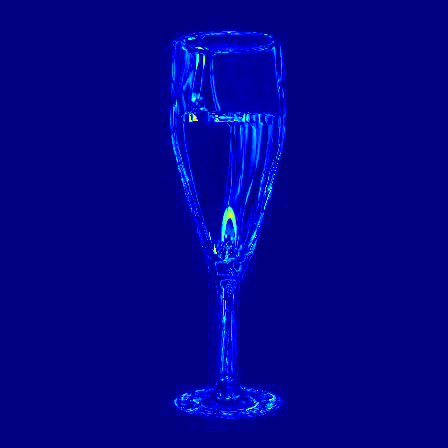}
    \includegraphics[width=0.095\textwidth]{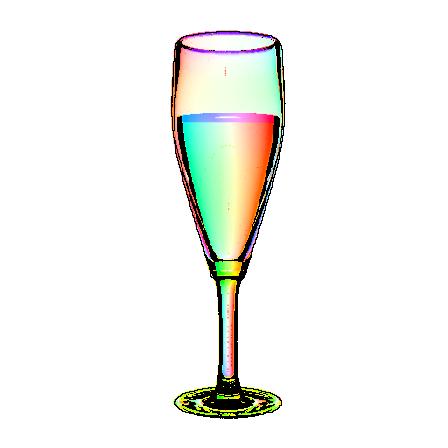}
    \includegraphics[width=0.095\textwidth]{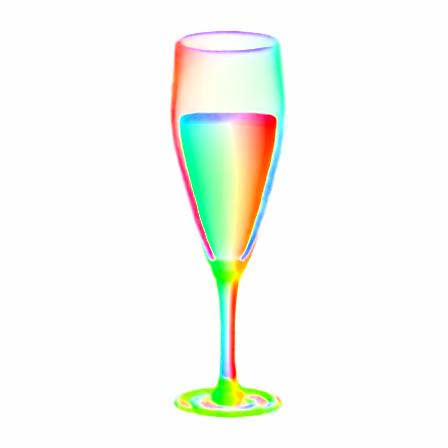}
    \includegraphics[width=0.095\textwidth]{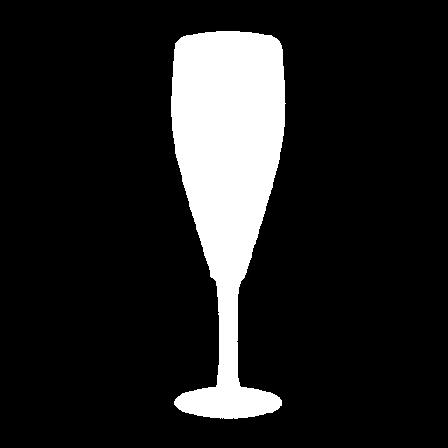}
    \includegraphics[width=0.095\textwidth]{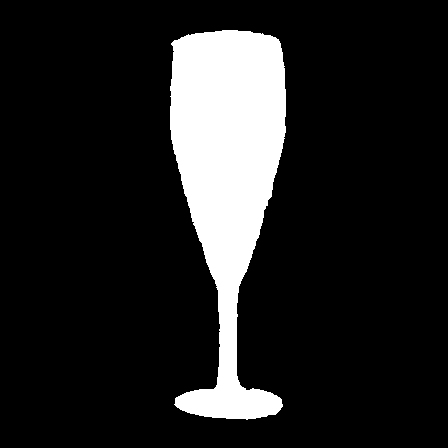}
    \includegraphics[width=0.095\textwidth]{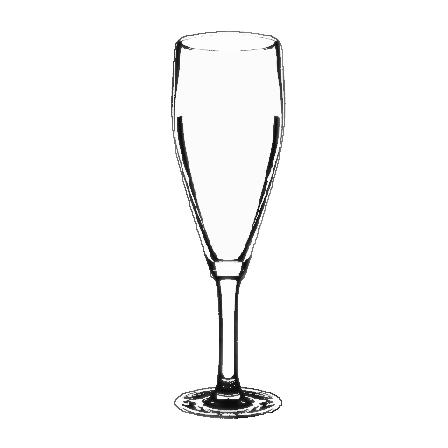}
    \includegraphics[width=0.095\textwidth]{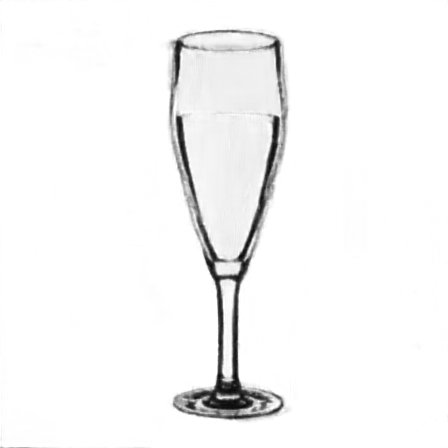}
    \\
    \vspace{-0.3em}\makebox[0.38\textwidth]{\scriptsize (b) Glass with Water, I-MSE = $0.15\times 10^{-2}$} 
    \makebox[0.19\textwidth]{\scriptsize F-EPE = 3.8 / 25.0} 
    \makebox[0.19\textwidth]{\scriptsize M-IoU = 0.97} 
    \makebox[0.19\textwidth]{\scriptsize A-MSE = 0.40 $\times 10^{-2}$} 
    \\
    \includegraphics[width=0.095\textwidth]{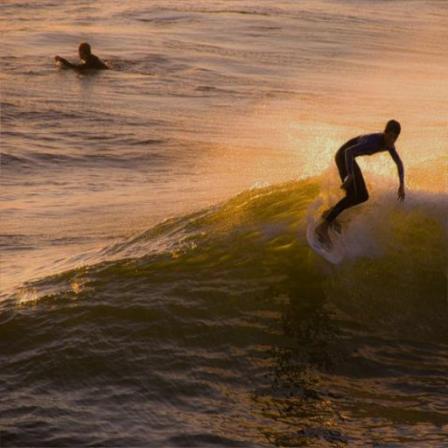}
    \includegraphics[width=0.095\textwidth]{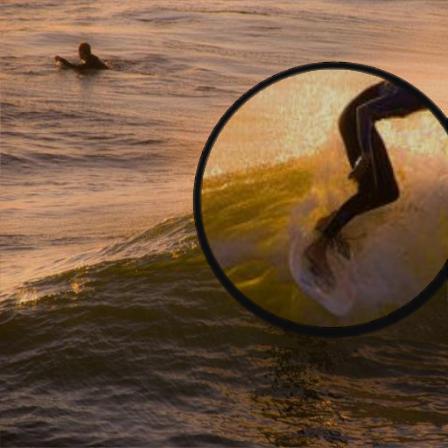}
    \includegraphics[width=0.095\textwidth]{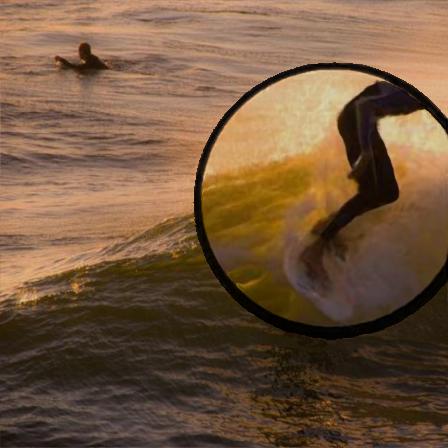}
   \includegraphics[width=0.095\textwidth]{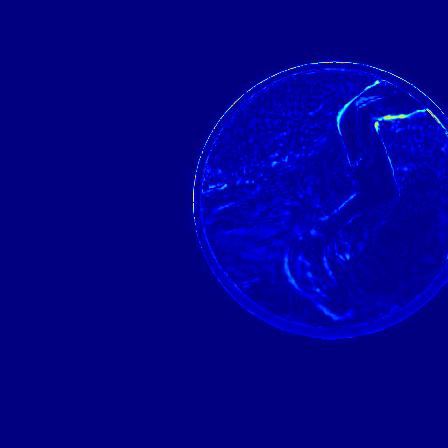}
    \includegraphics[width=0.095\textwidth]{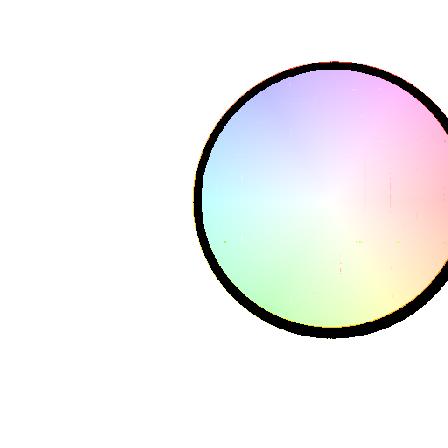}
    \includegraphics[width=0.095\textwidth]{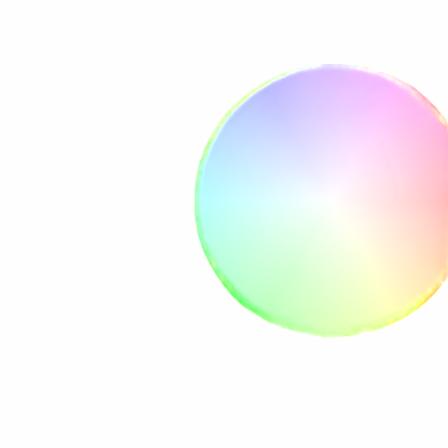}
    \includegraphics[width=0.095\textwidth]{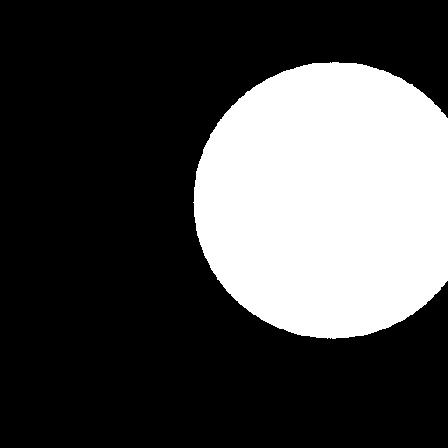}
    \includegraphics[width=0.095\textwidth]{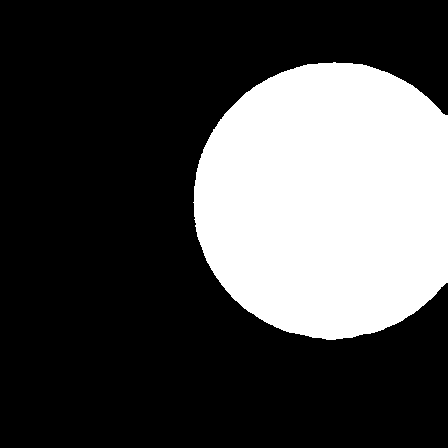}
    \includegraphics[width=0.095\textwidth]{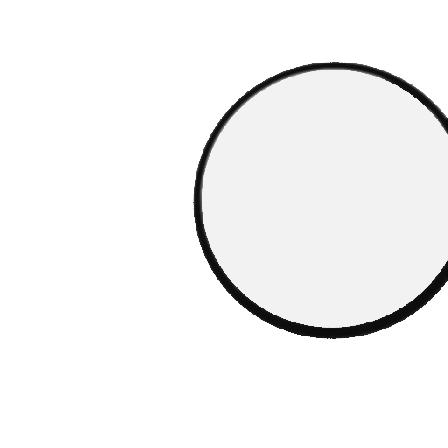}
    \includegraphics[width=0.095\textwidth]{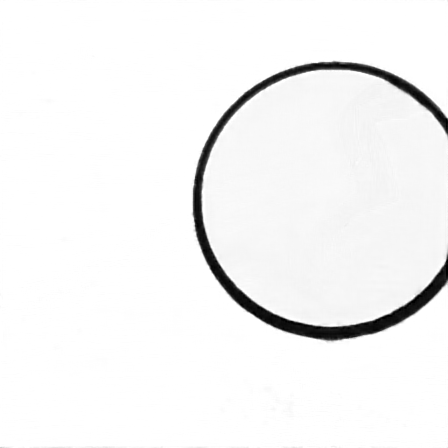}
    \\
    \vspace{-0.3em}\makebox[0.38\textwidth]{\scriptsize (c) Lens, I-MSE = $0.079\times 10^{-2}$}
    \makebox[0.19\textwidth]{\scriptsize F-EPE = 1.5 / 3.7} 
    \makebox[0.19\textwidth]{\scriptsize M-IoU = 1.00} 
    \makebox[0.19\textwidth]{\scriptsize A-MSE = $0.17\times 10^{-2}$} 
    \\
    \includegraphics[width=0.095\textwidth]{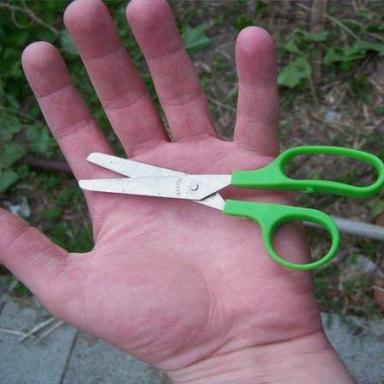}
    \includegraphics[width=0.095\textwidth]{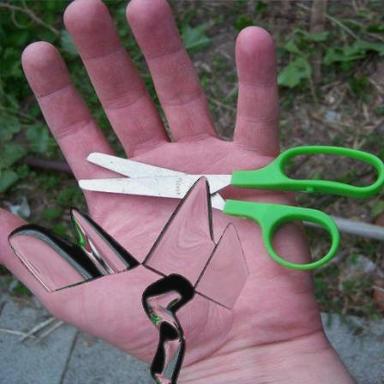}
    \includegraphics[width=0.095\textwidth]{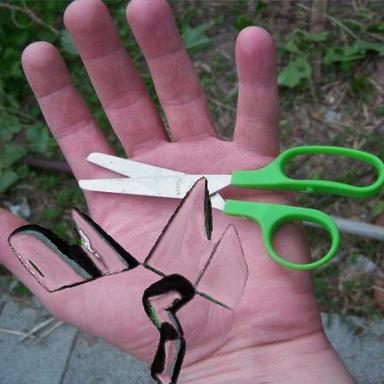}
    \includegraphics[width=0.095\textwidth]{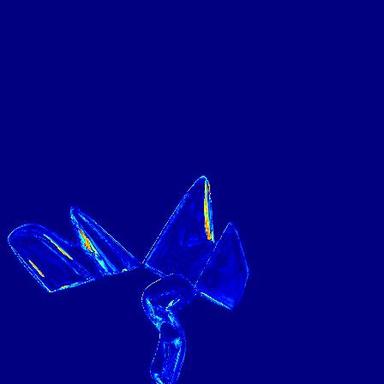}
    \includegraphics[width=0.095\textwidth]{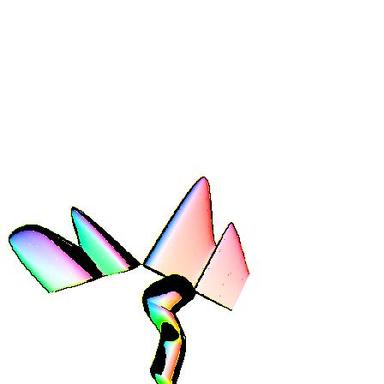}
    \includegraphics[width=0.095\textwidth]{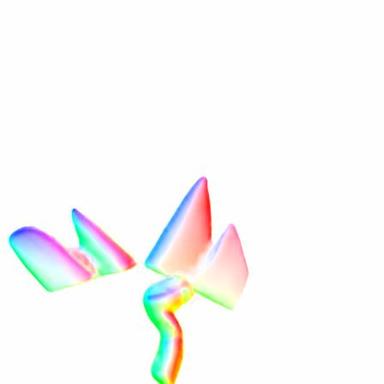}
    \includegraphics[width=0.095\textwidth]{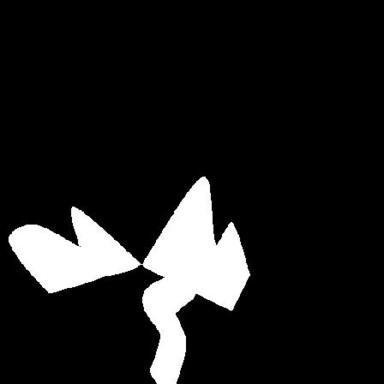}
    \includegraphics[width=0.095\textwidth]{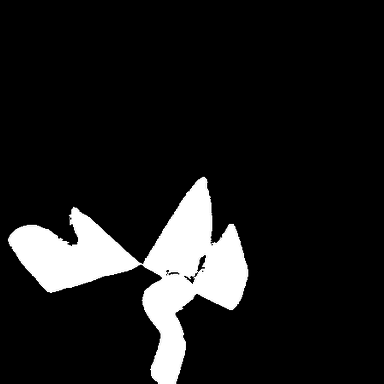}
    \includegraphics[width=0.095\textwidth]{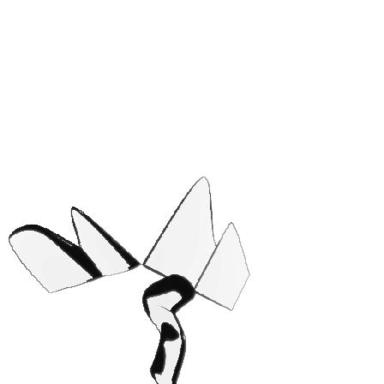}
    \includegraphics[width=0.095\textwidth]{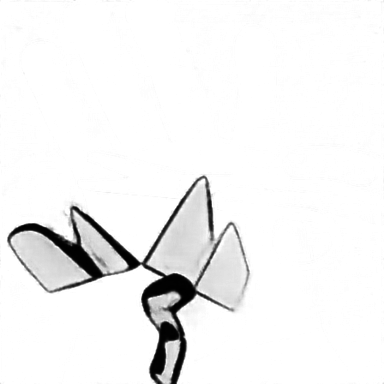}
    \\ \vspace{-0.3em}\makebox[0.38\textwidth]{\scriptsize (d) Complex Shape 1, I-MSE = $0.15 \times 10^{-2}$} 
    \makebox[0.19\textwidth]{\scriptsize F-EPE = 3.4 / 24.3} 
    \makebox[0.19\textwidth]{\scriptsize M-IoU = 0.97} 
    \makebox[0.19\textwidth]{\scriptsize A-MSE = $0.19 \times 10^{-2}$} 
    \\
    \includegraphics[width=0.095\textwidth]{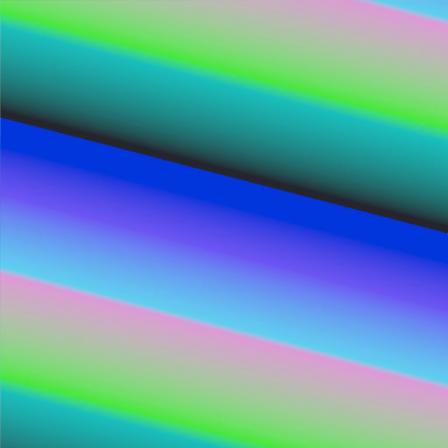}
    \includegraphics[width=0.095\textwidth]{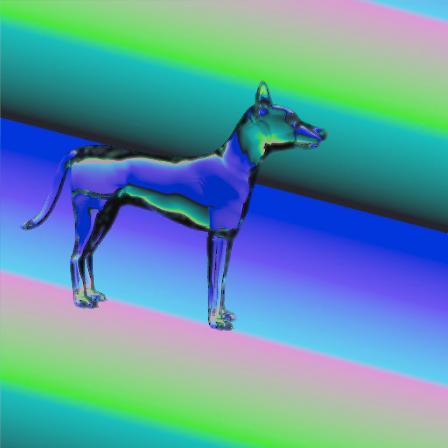}
    \includegraphics[width=0.095\textwidth]{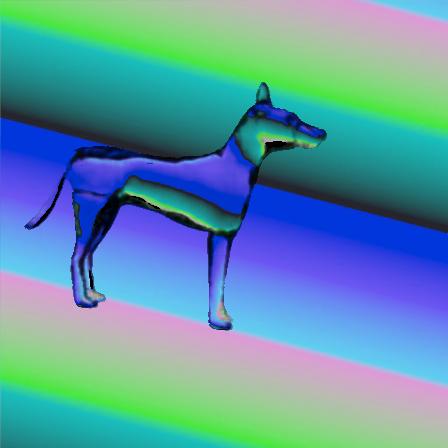}
    \includegraphics[width=0.095\textwidth]{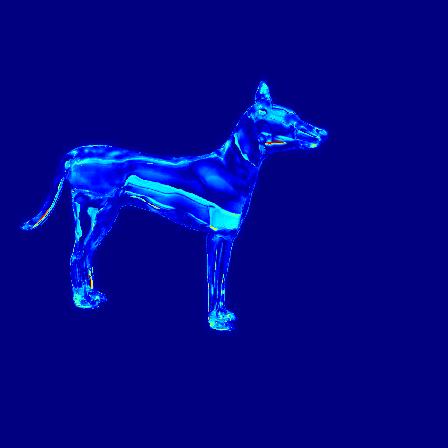}
    \includegraphics[width=0.095\textwidth]{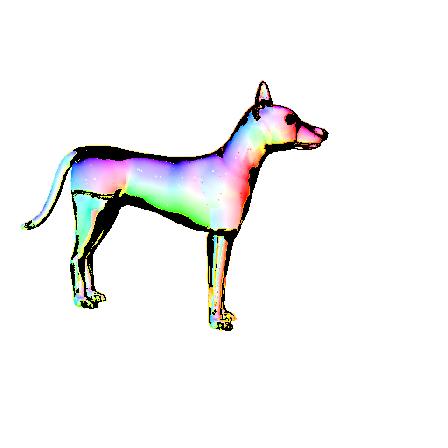}
    \includegraphics[width=0.095\textwidth]{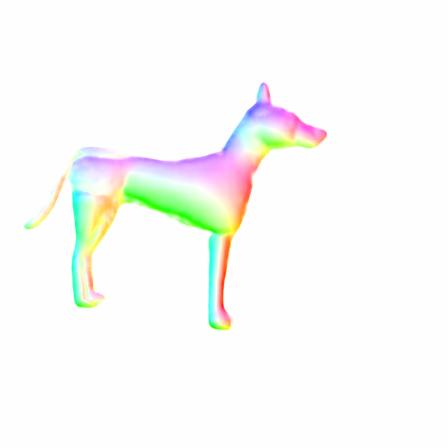}
    \includegraphics[width=0.095\textwidth]{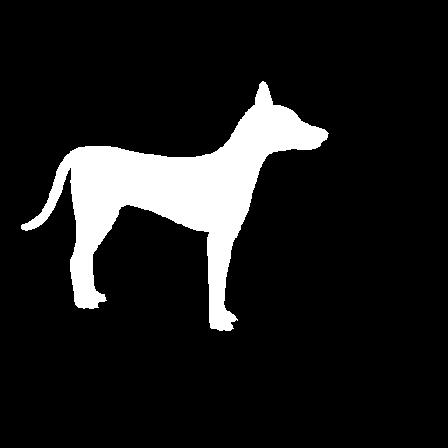}
    \includegraphics[width=0.095\textwidth]{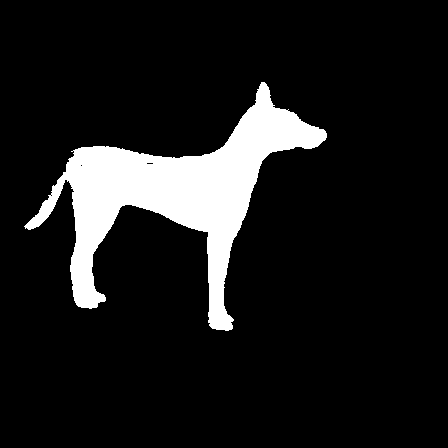}
    \includegraphics[width=0.095\textwidth]{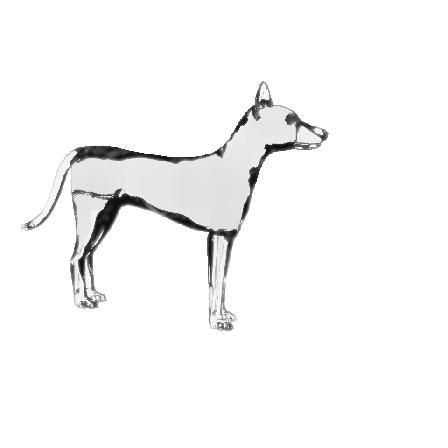}
    \includegraphics[width=0.095\textwidth]{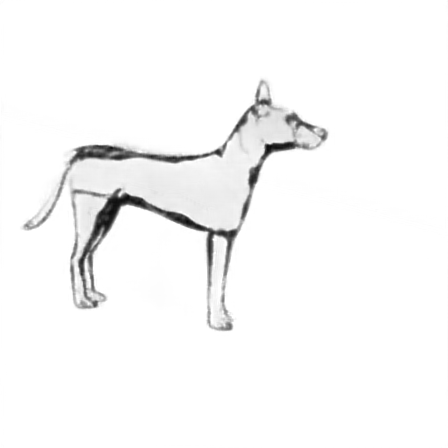}
    \\
    \vspace{-0.3em}\makebox[0.38\textwidth]{\scriptsize (e) Complex Dog, I-MSE = $0.28\times 10^{-2}$} 
    \makebox[0.19\textwidth]{\scriptsize F-EPE = 5.05 / 40.6} 
    \makebox[0.19\textwidth]{\scriptsize M-IoU = 0.96} 
    \makebox[0.19\textwidth]{\scriptsize A-MSE = $0.16\times 10^{-2}$} 

%% file: user_study_reason.tex
    \makebox[0.3\textwidth]{\footnotesize Predicted Environment Matte} \vspace{0.2em}\\
    \includegraphics[width=0.15\textwidth]{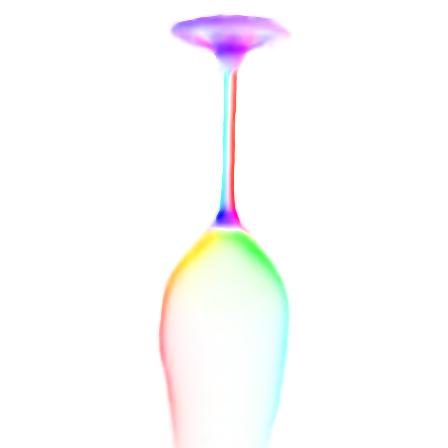}
    \includegraphics[width=0.15\textwidth]{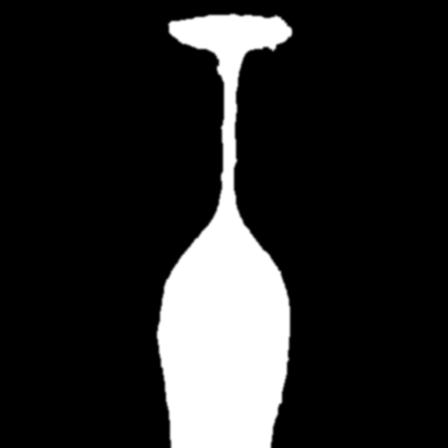}
    \includegraphics[width=0.15\textwidth]{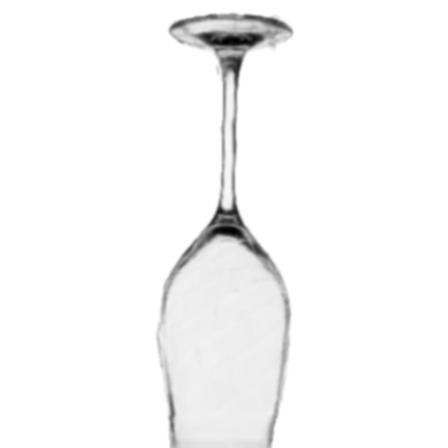}
    \\
    \makebox[0.23\textwidth]{\footnotesize Photograph} 
    \makebox[0.23\textwidth]{\footnotesize Composite} \\
    \includegraphics[width=0.238\textwidth]{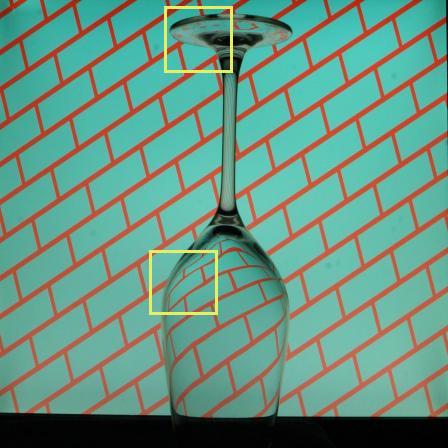}
    \includegraphics[width=0.238\textwidth]{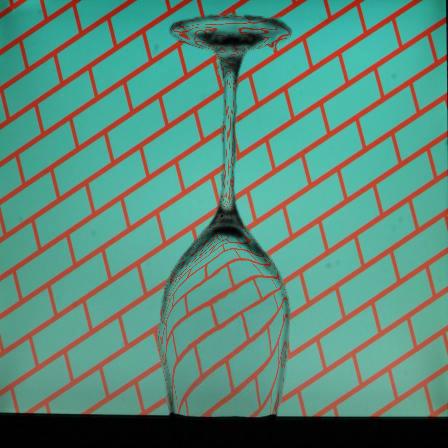}
    \\
    \includegraphics[width=0.116\textwidth]{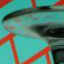}
    \includegraphics[width=0.116\textwidth]{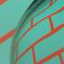}
    \includegraphics[width=0.116\textwidth]{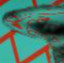}
    \includegraphics[width=0.116\textwidth]{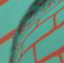}

%% file: real_qual.tex
    \makebox[0.12\textwidth]{\footnotesize Background} 
    \makebox[0.12\textwidth]{\footnotesize Input} 
    \makebox[0.12\textwidth]{\footnotesize Rec. Image} 
    \makebox[0.12\textwidth]{\footnotesize Rec. Error} 
    \makebox[0.12\textwidth]{\footnotesize Ref. Flow} 
    \makebox[0.12\textwidth]{\footnotesize Obj. Mask} 
    \makebox[0.12\textwidth]{\footnotesize Att. Mask} 
    \makebox[0.12\textwidth]{\footnotesize Composite} 
    \\
    \includegraphics[width=0.12\textwidth]{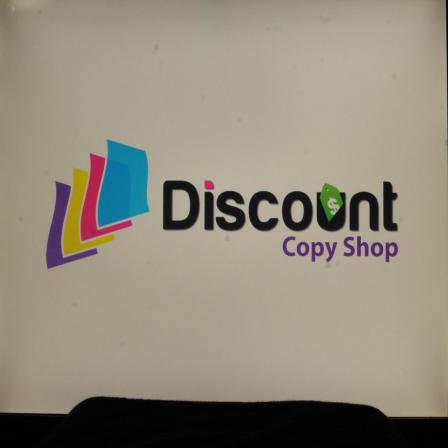}
    \includegraphics[width=0.12\textwidth]{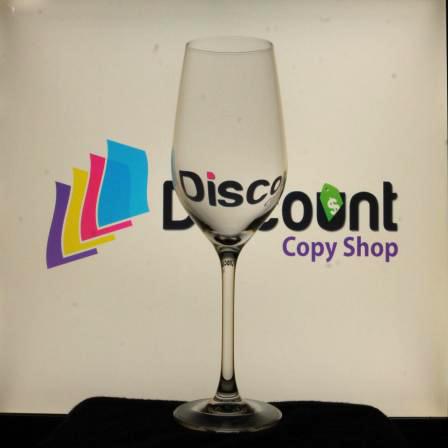}
    \includegraphics[width=0.12\textwidth]{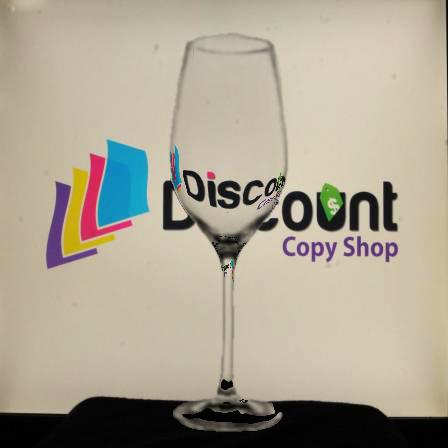}
    \includegraphics[width=0.12\textwidth]{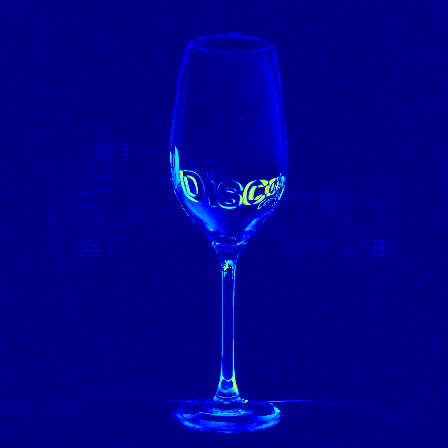}
    \includegraphics[width=0.12\textwidth]{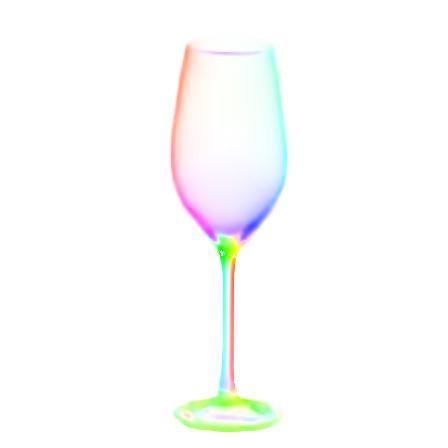}
    \includegraphics[width=0.12\textwidth]{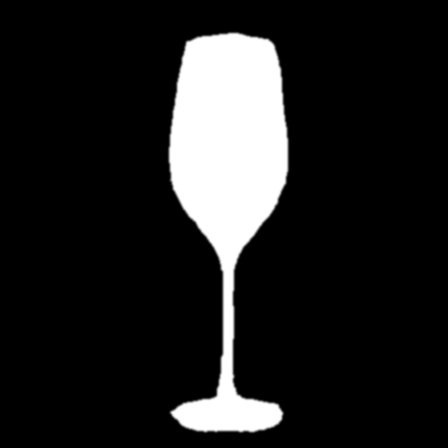}
    \includegraphics[width=0.12\textwidth]{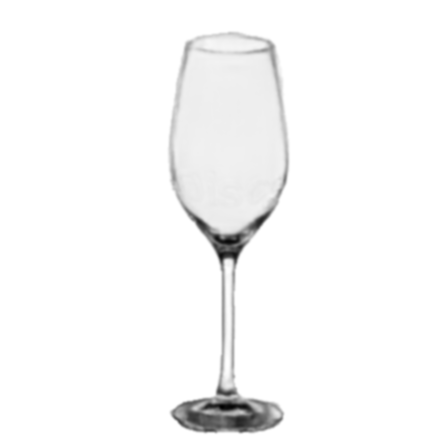}
    \includegraphics[width=0.12\textwidth]{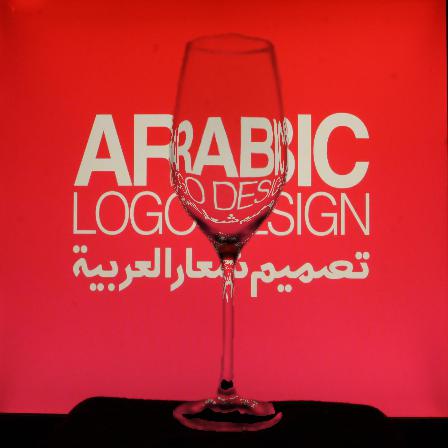}
    \\
    \vspace{-0.3em}\makebox[0.25\textwidth]{\scriptsize (a) Glass}
    \makebox[0.3\textwidth]{\scriptsize PSNR = 27.6, SSIM = 0.95}
    \makebox[0.4\textwidth]{\scriptsize }
    \\
    \includegraphics[width=0.12\textwidth]{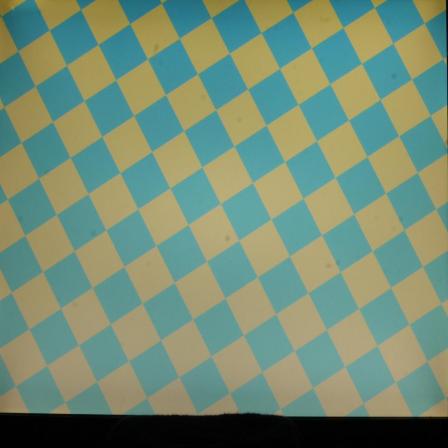}
    \includegraphics[width=0.12\textwidth]{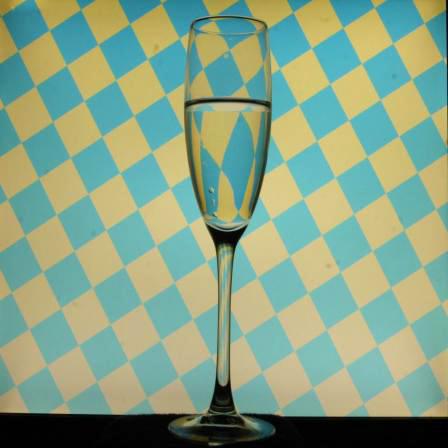}
    \includegraphics[width=0.12\textwidth]{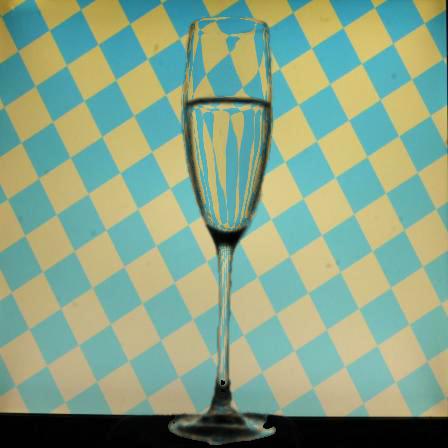}
    \includegraphics[width=0.12\textwidth]{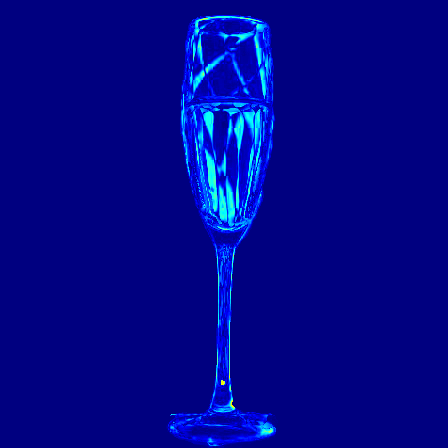}
    \includegraphics[width=0.12\textwidth]{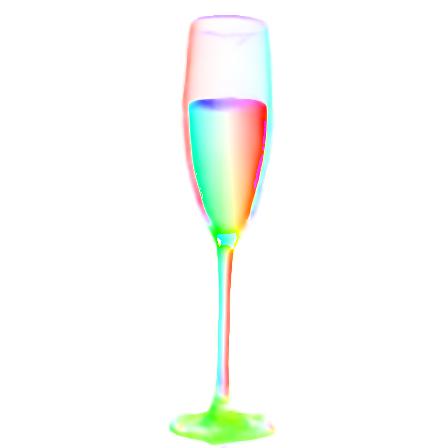}
    \includegraphics[width=0.12\textwidth]{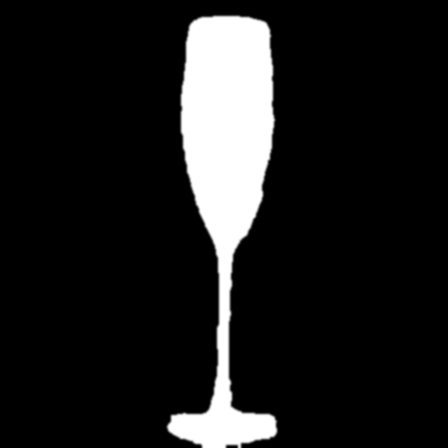}
    \includegraphics[width=0.12\textwidth]{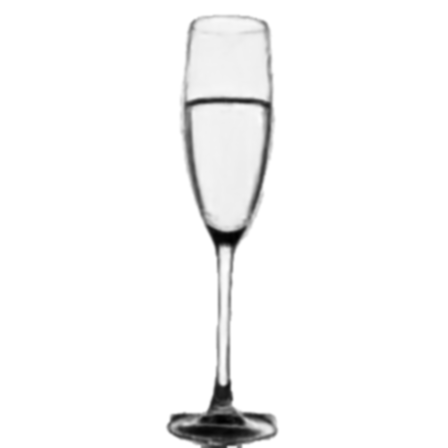}
    \includegraphics[width=0.12\textwidth]{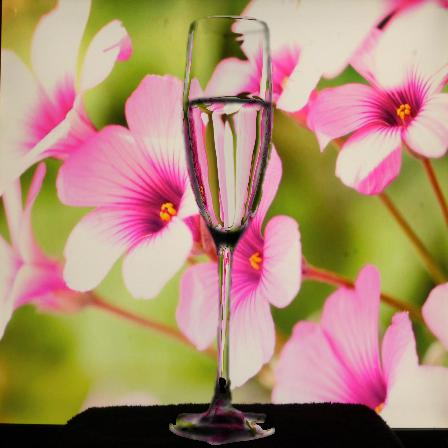}
    \\
    \vspace{-0.3em}\makebox[0.25\textwidth]{\scriptsize (b) Glass with Water}
    \makebox[0.3\textwidth]{\scriptsize PSNR = 27.3, SSIM = 0.95}
    \makebox[0.4\textwidth]{\scriptsize }
    \\
    \includegraphics[width=0.12\textwidth]{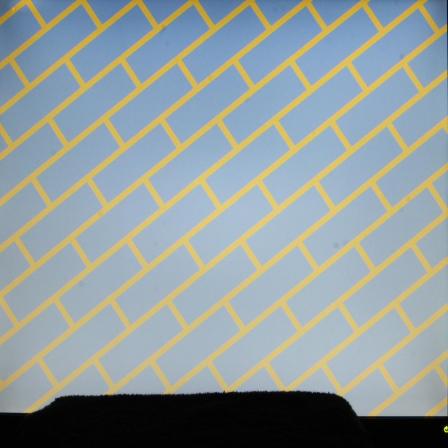}
    \includegraphics[width=0.12\textwidth]{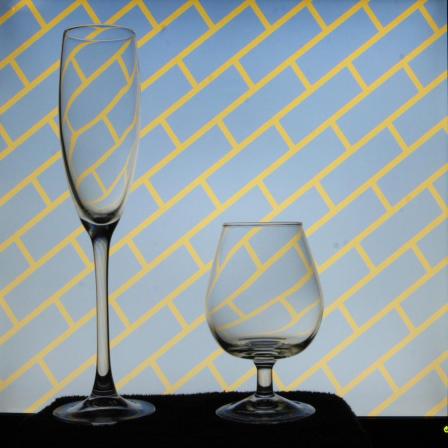}
    \includegraphics[width=0.12\textwidth]{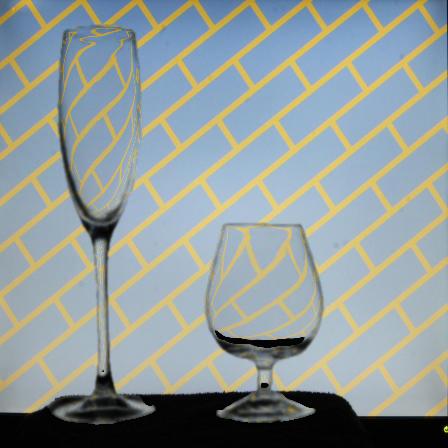}
    \includegraphics[width=0.12\textwidth]{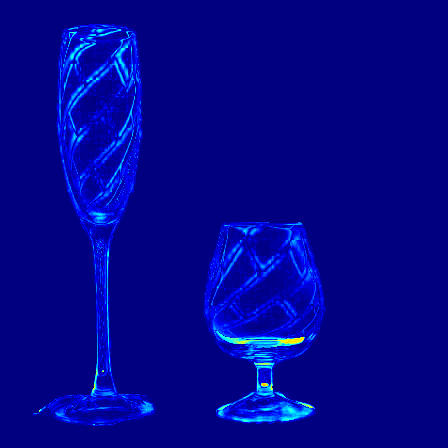}
    \includegraphics[width=0.12\textwidth]{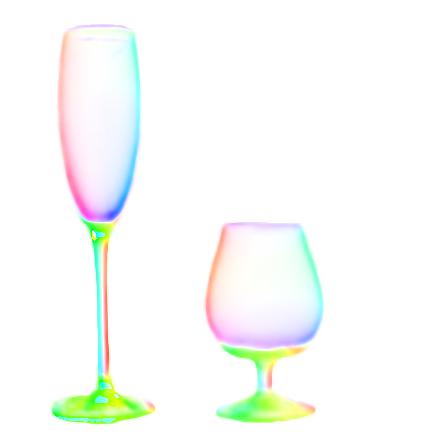}
    \includegraphics[width=0.12\textwidth]{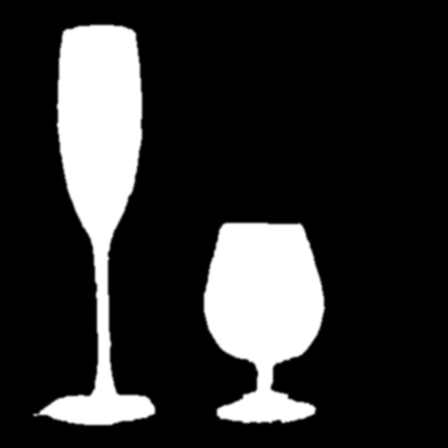}
    \includegraphics[width=0.12\textwidth]{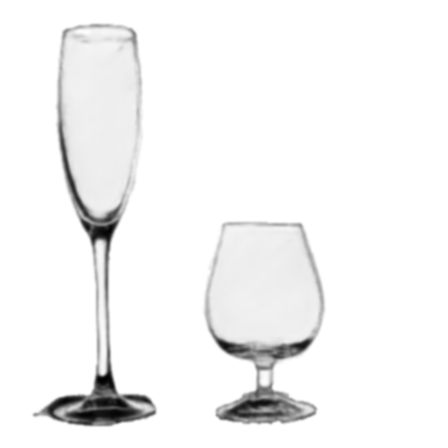}
    \includegraphics[width=0.12\textwidth]{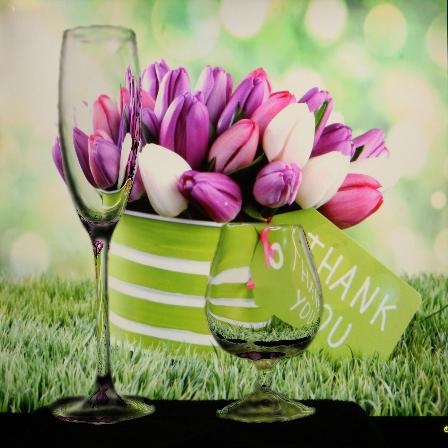}
    \\
    \vspace{-0.3em}\makebox[0.25\textwidth]{\scriptsize (c) Multi-objects}
    \makebox[0.3\textwidth]{\scriptsize PSNR = 25.37, SSIM = 0.93}
    \makebox[0.4\textwidth]{\scriptsize }
    \\
    \includegraphics[width=0.12\textwidth]{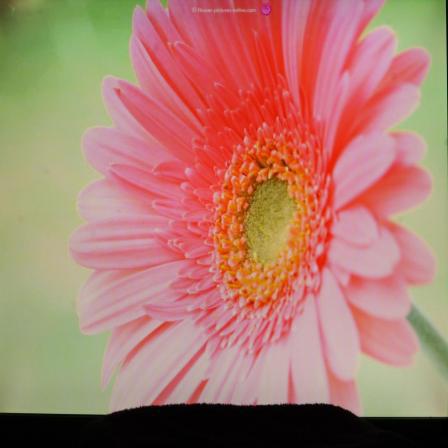}
    \includegraphics[width=0.12\textwidth]{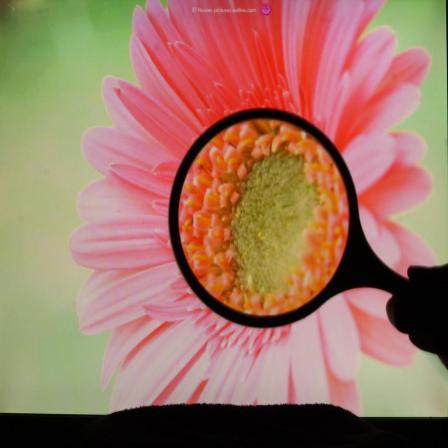}
    \includegraphics[width=0.12\textwidth]{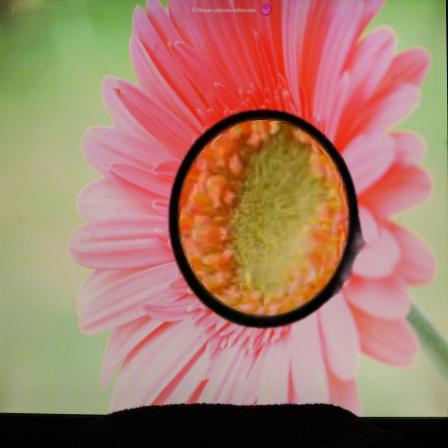}
    \includegraphics[width=0.12\textwidth]{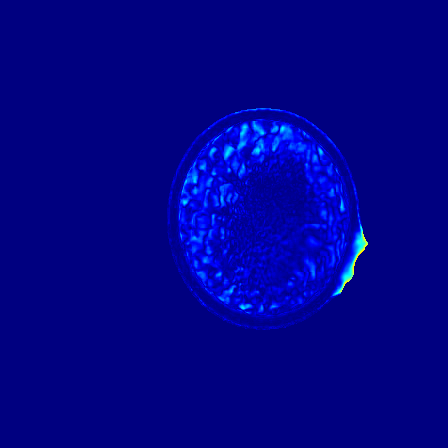}
    \includegraphics[width=0.12\textwidth]{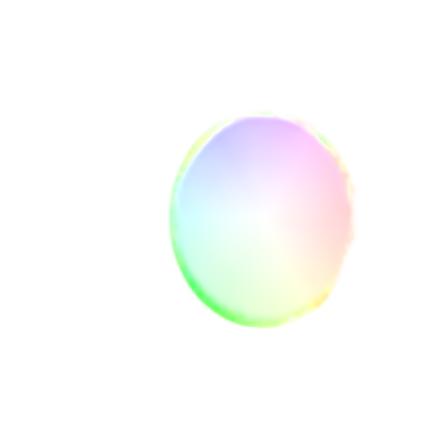}
    \includegraphics[width=0.12\textwidth]{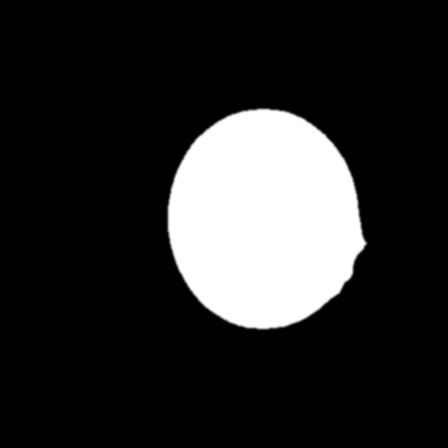}
    \includegraphics[width=0.12\textwidth]{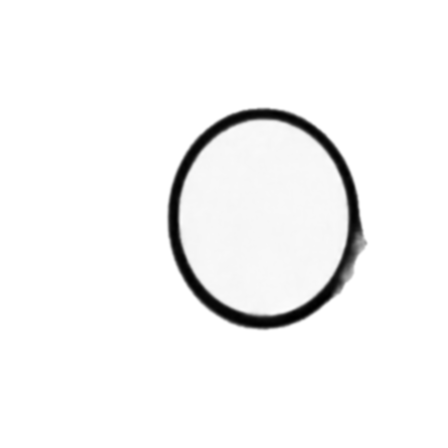}
    \includegraphics[width=0.12\textwidth]{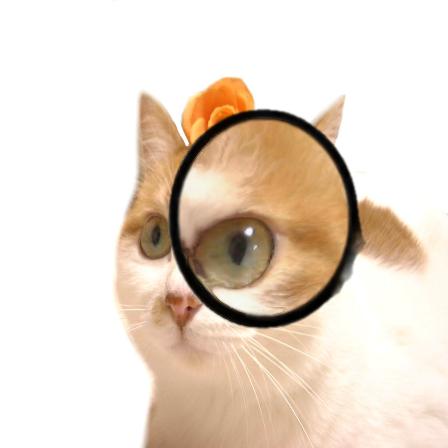}
    \\
    \vspace{-0.3em}\makebox[0.25\textwidth]{\scriptsize (d) Lens}
    \makebox[0.3\textwidth]{\scriptsize PSNR = 27.15, SSIM = 0.91}
    \makebox[0.4\textwidth]{\scriptsize }
    \\
    \includegraphics[width=0.12\textwidth]{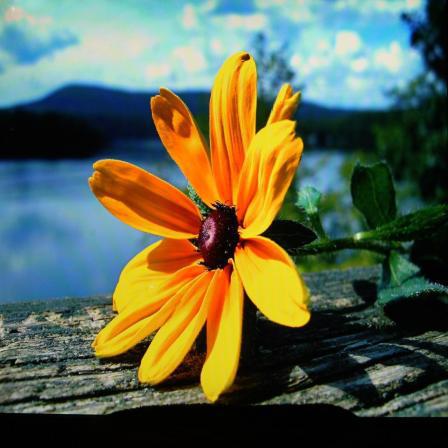}
    \includegraphics[width=0.12\textwidth]{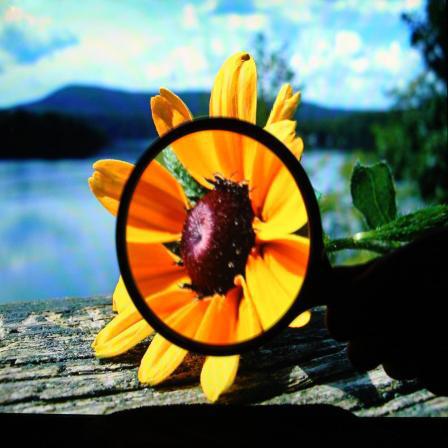}
    \includegraphics[width=0.12\textwidth]{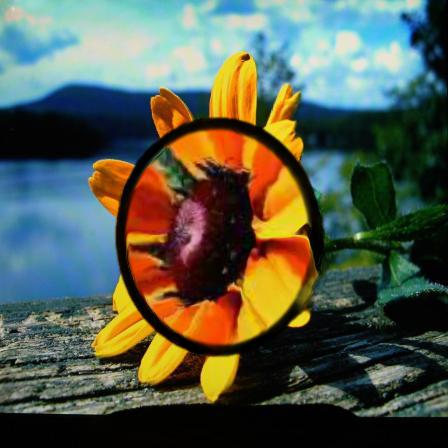}
    \includegraphics[width=0.12\textwidth]{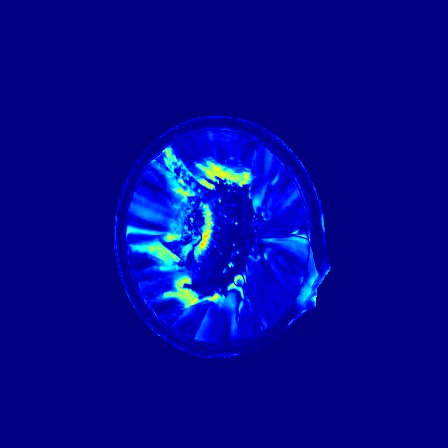}
    \includegraphics[width=0.12\textwidth]{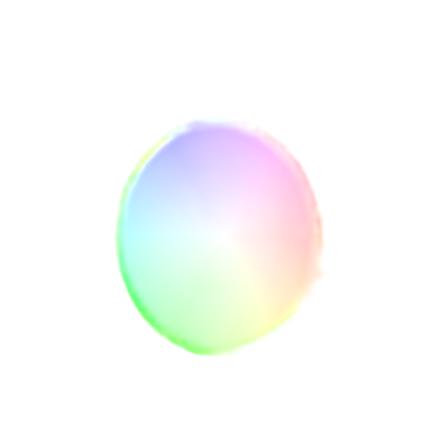}
    \includegraphics[width=0.12\textwidth]{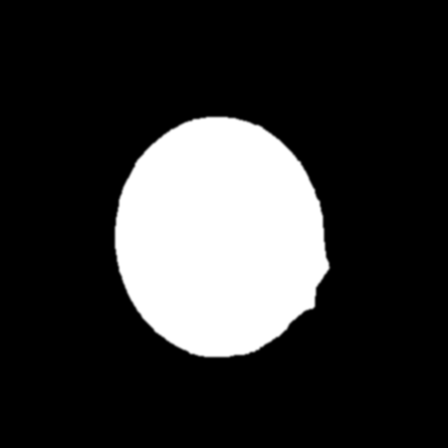}
    \includegraphics[width=0.12\textwidth]{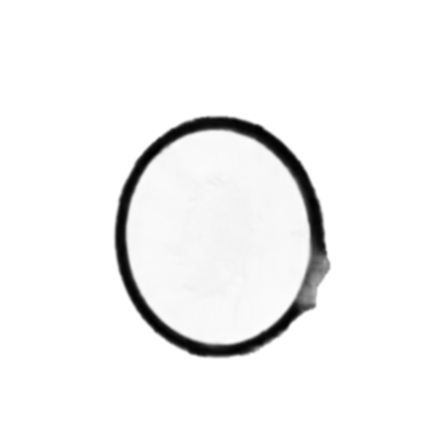}
    \includegraphics[width=0.12\textwidth]{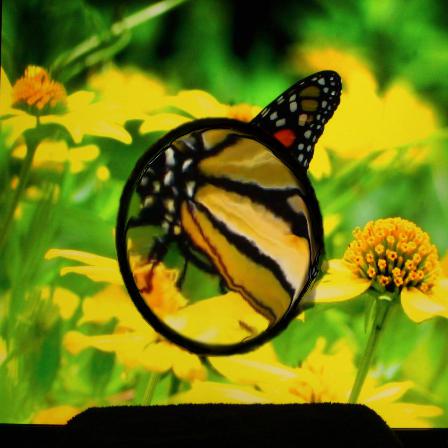}
    \\ 
    \vspace{-0.3em}\makebox[0.25\textwidth]{\scriptsize (e) Lens}
    \makebox[0.3\textwidth]{\scriptsize PSNR=21.22, SSIM=0.90}
    \makebox[0.4\textwidth]{\scriptsize }
    \\
    \includegraphics[width=0.12\textwidth]{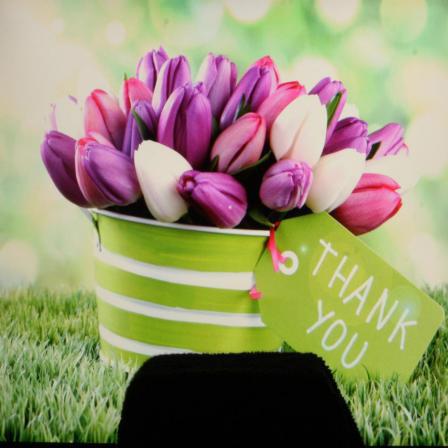}
    \includegraphics[width=0.12\textwidth]{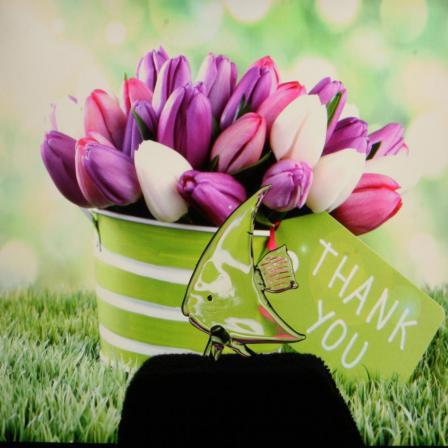}
    \includegraphics[width=0.12\textwidth]{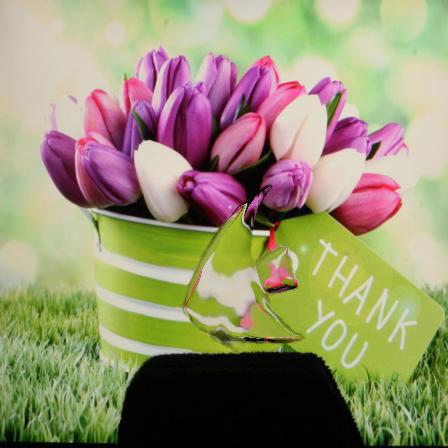}
    \includegraphics[width=0.12\textwidth]{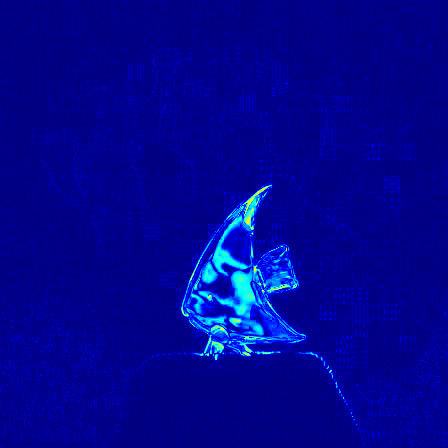}
    \includegraphics[width=0.12\textwidth]{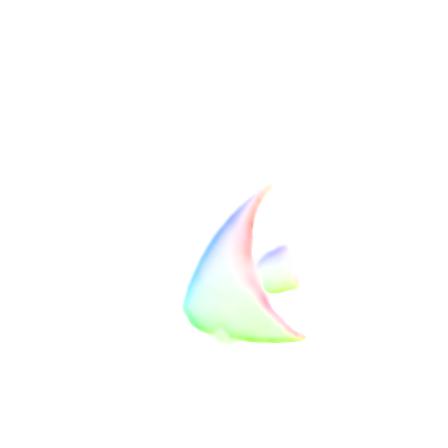}
    \includegraphics[width=0.12\textwidth]{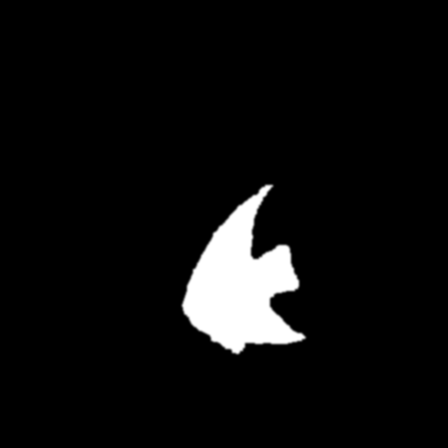}
    \includegraphics[width=0.12\textwidth]{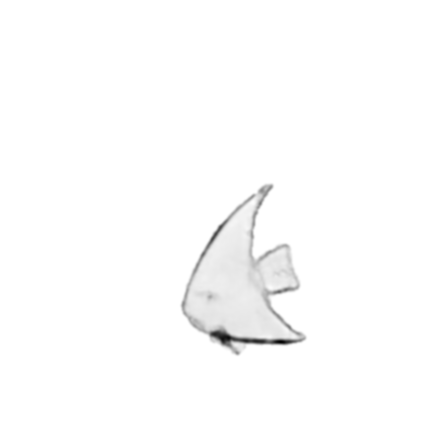}
    \includegraphics[width=0.12\textwidth]{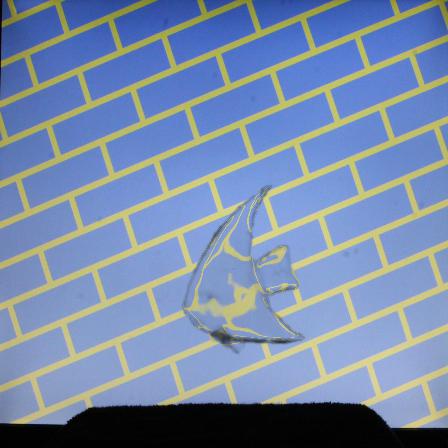}
    \\
    \vspace{-0.3em}\makebox[0.25\textwidth]{\scriptsize (f) Complex Fish}
    \makebox[0.3\textwidth]{\scriptsize PSNR=25.09, SSIM=0.94}
    \makebox[0.4\textwidth]{\scriptsize }
    \\
    \includegraphics[width=0.12\textwidth]{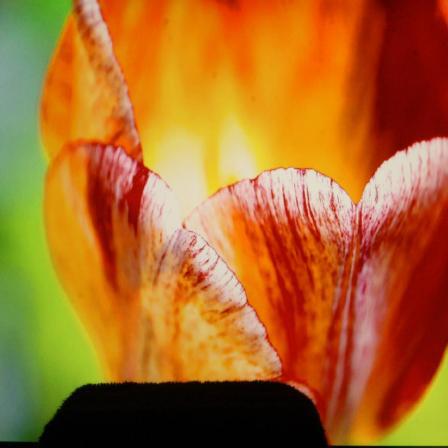}
    \includegraphics[width=0.12\textwidth]{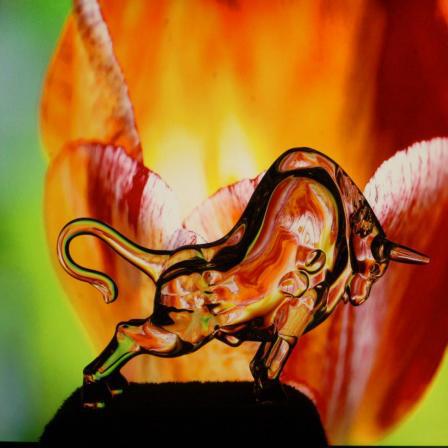}
    \includegraphics[width=0.12\textwidth]{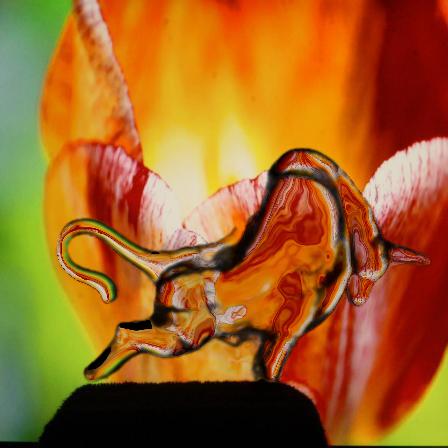}
    \includegraphics[width=0.12\textwidth]{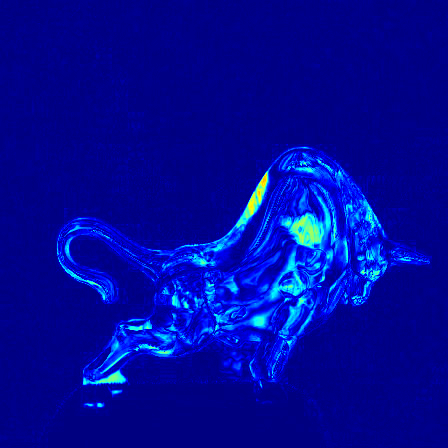}
    \includegraphics[width=0.12\textwidth]{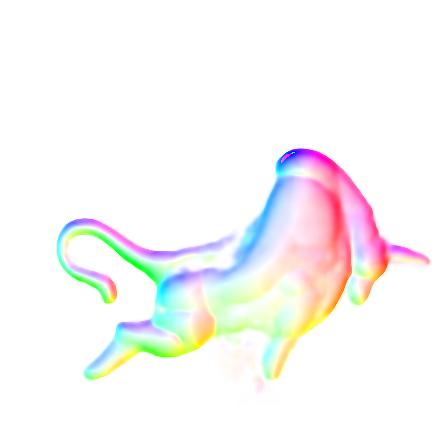}
    \includegraphics[width=0.12\textwidth]{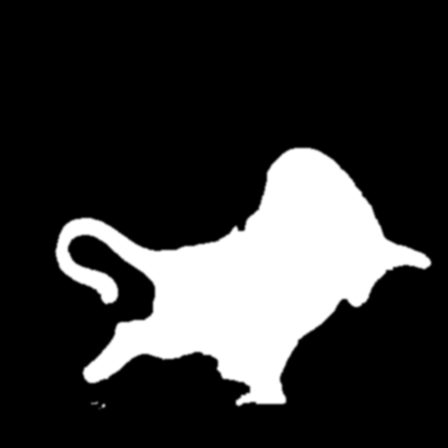}
    \includegraphics[width=0.12\textwidth]{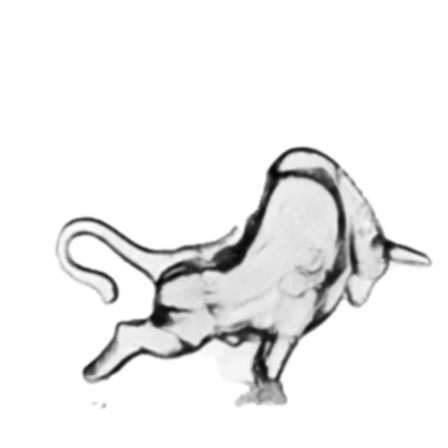}
    \includegraphics[width=0.12\textwidth]{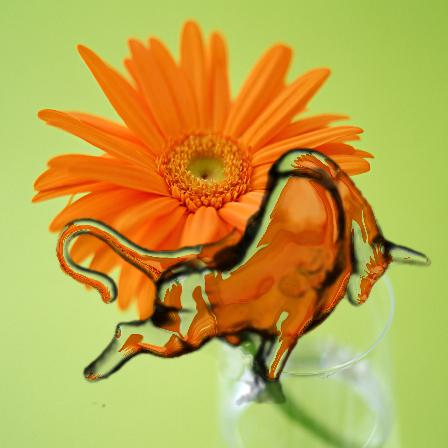}
    \\                                                                          
    \vspace{-0.3em}\makebox[0.25\textwidth]{\scriptsize (g) Complex Bull}
    \makebox[0.3\textwidth]{\scriptsize PSNR = 20.31, SSIM = 0.84}
    \makebox[0.4\textwidth]{\scriptsize }
    \\
    \includegraphics[width=0.12\textwidth]{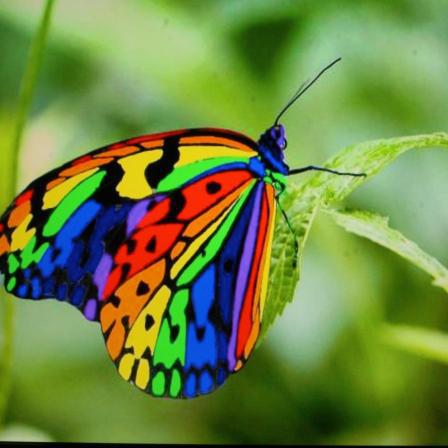}
    \includegraphics[width=0.12\textwidth]{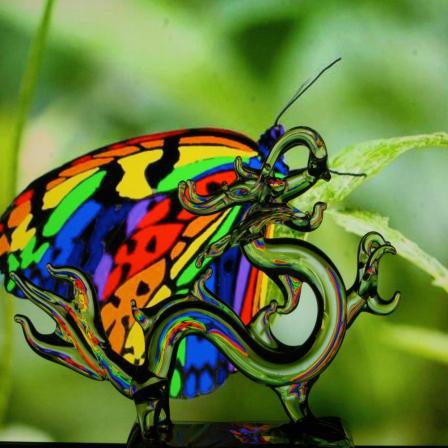}
    \includegraphics[width=0.12\textwidth]{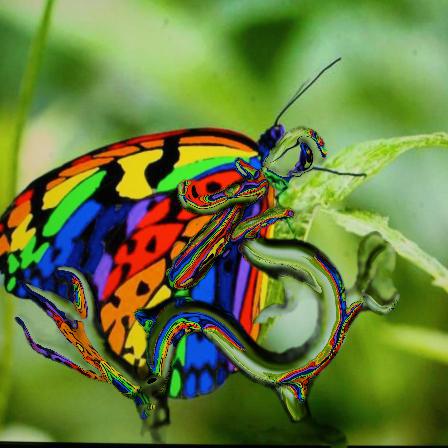}
    \includegraphics[width=0.12\textwidth]{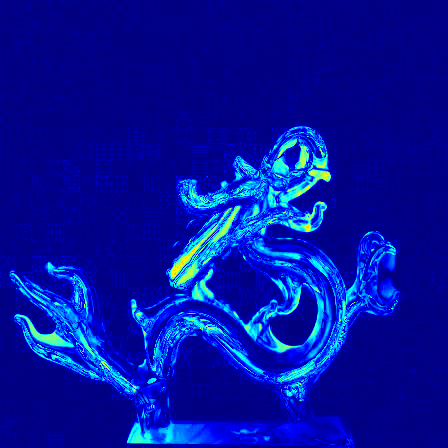}
    \includegraphics[width=0.12\textwidth]{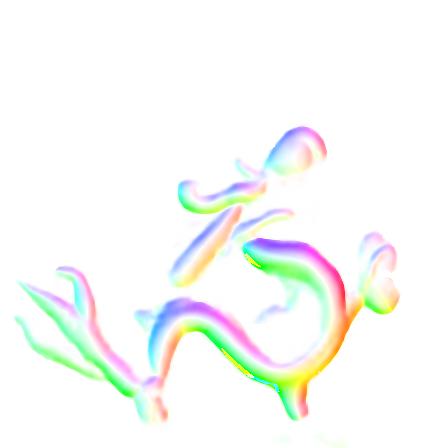}
    \includegraphics[width=0.12\textwidth]{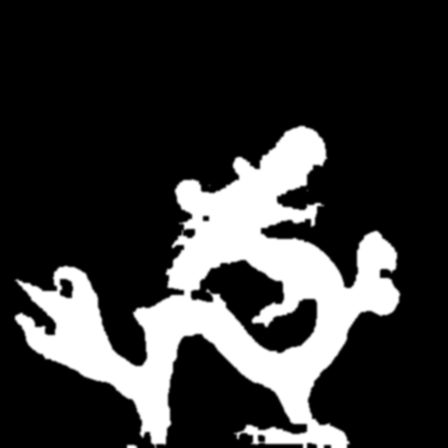}
    \includegraphics[width=0.12\textwidth]{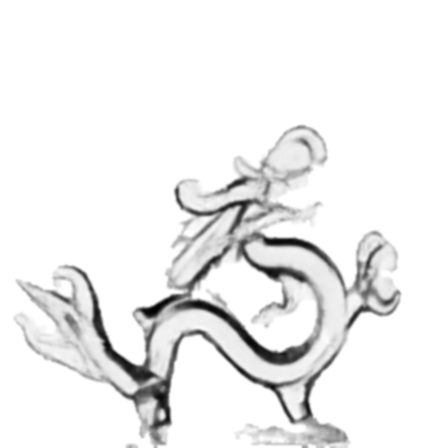}
    \includegraphics[width=0.12\textwidth]{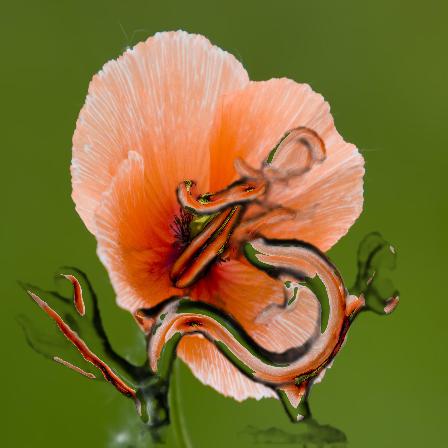}
    \\
    \vspace{-0.3em}\makebox[0.25\textwidth]{\scriptsize (h) Complex Dragon}
    \makebox[0.3\textwidth]{\scriptsize PSNR = 18.46, SSIM = 0.80}
    \makebox[0.4\textwidth]{\scriptsize }
    \\
    \includegraphics[width=0.12\textwidth]{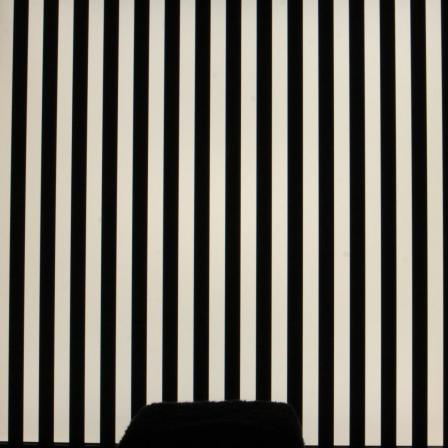}
    \includegraphics[width=0.12\textwidth]{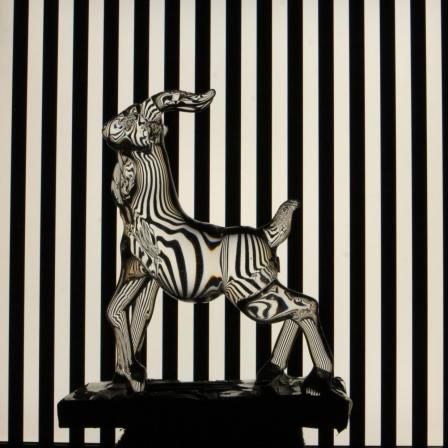}
    \includegraphics[width=0.12\textwidth]{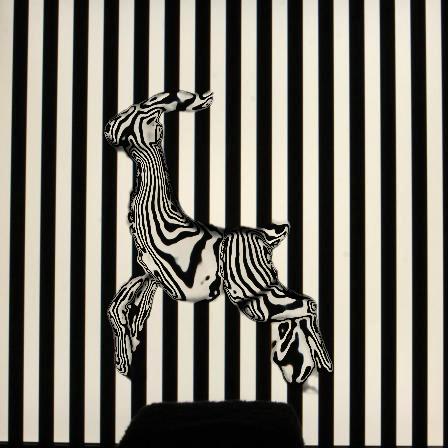}
    \includegraphics[width=0.12\textwidth]{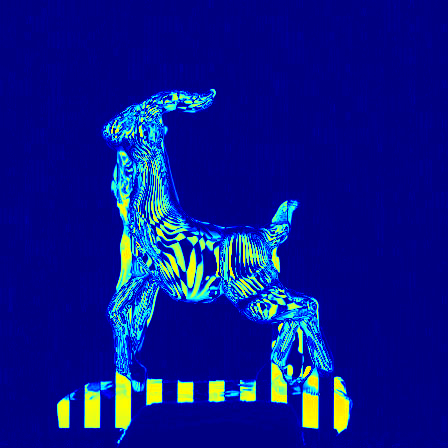}
    \includegraphics[width=0.12\textwidth]{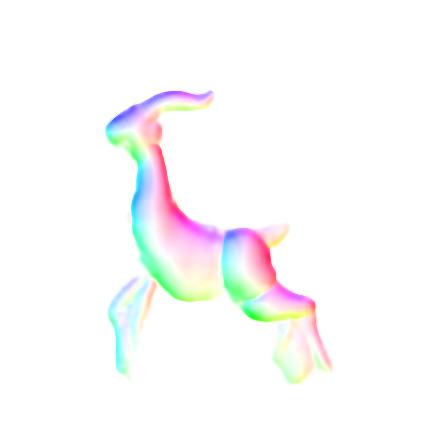}
    \includegraphics[width=0.12\textwidth]{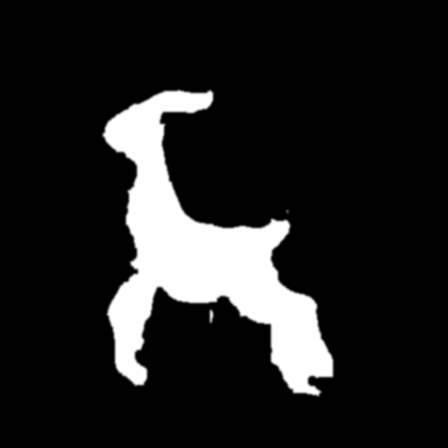}
    \includegraphics[width=0.12\textwidth]{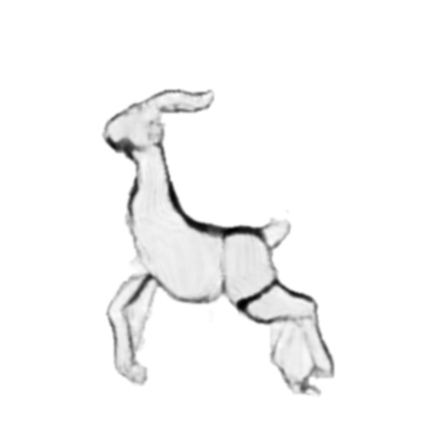}
    \includegraphics[width=0.12\textwidth]{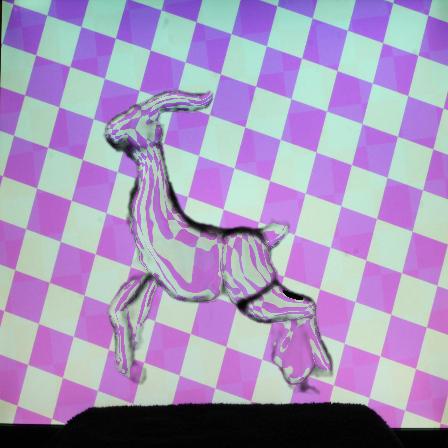}
    \\
    \vspace{-0.3em}\makebox[0.25\textwidth]{\scriptsize (i) Complex Sheep}
    \makebox[0.3\textwidth]{\scriptsize PSNR=14.48, SSIM=0.79}
    \makebox[0.4\textwidth]{\scriptsize }
    \\

%% file: sup_editFlow_example.tex
    \includegraphics[width=0.135\textwidth]{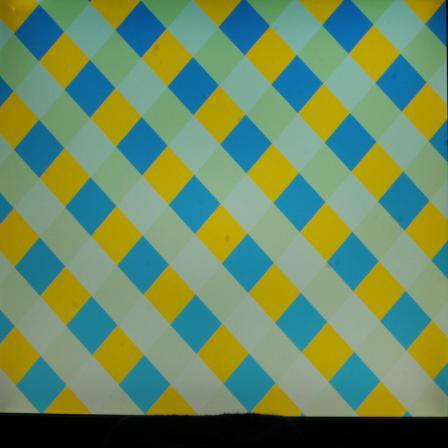}
    \includegraphics[width=0.135\textwidth]{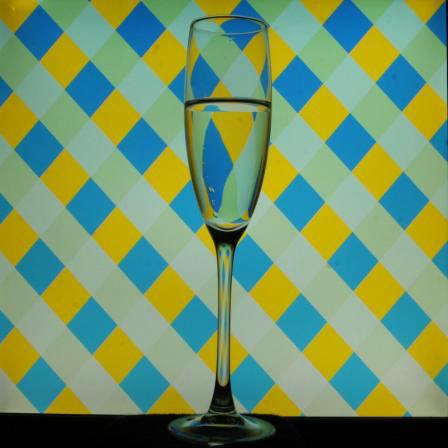}
    \includegraphics[width=0.135\textwidth]{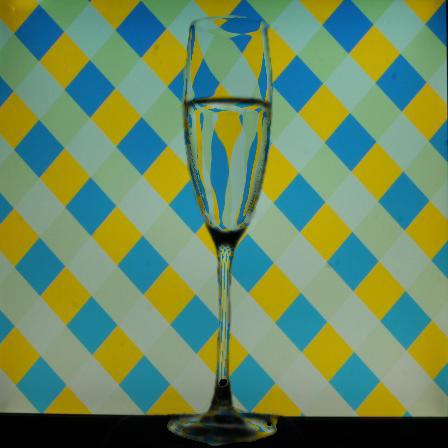}
    \includegraphics[width=0.135\textwidth]{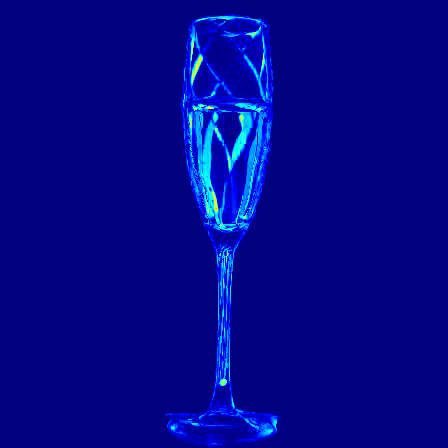}
    \includegraphics[width=0.135\textwidth]{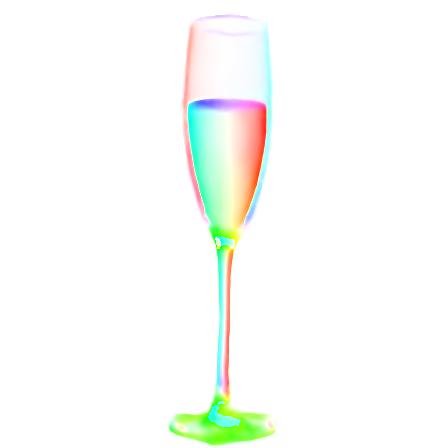}
    \includegraphics[width=0.135\textwidth]{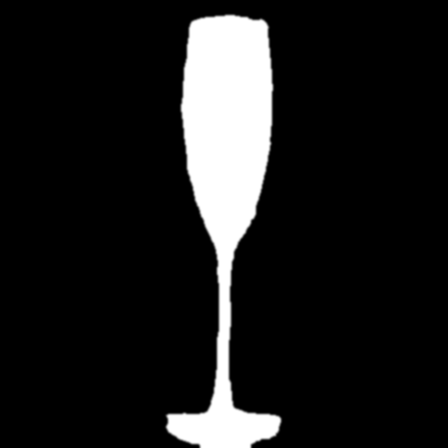}
    \includegraphics[width=0.135\textwidth]{{edit_flow_1__1.00_1.0_4_rho_r}.png}

%% file: sup_scaleFlow.tex
    \raisebox{0.1\height}{\rotatebox{90}{\scriptsize Ref. Flow}}
    \includegraphics[width=0.135\textwidth]{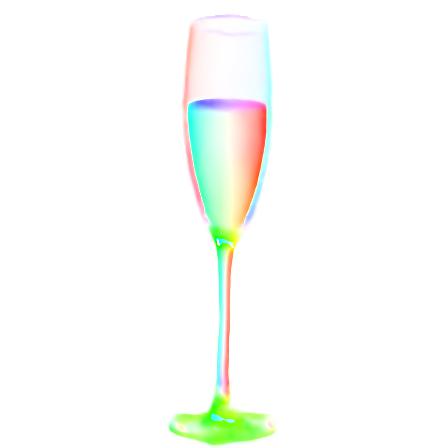}
    \includegraphics[width=0.135\textwidth]{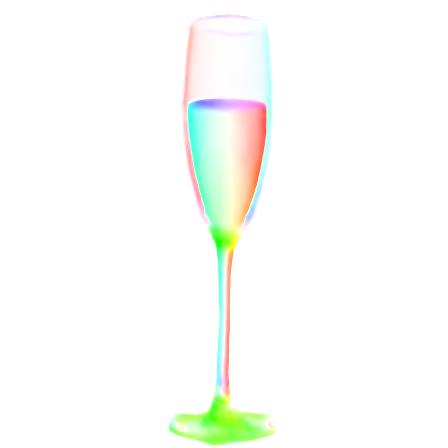}
    \includegraphics[width=0.135\textwidth]{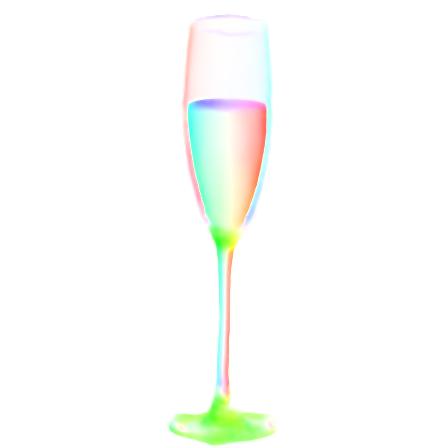}
    \makebox[0.135\textwidth]{}
    \includegraphics[width=0.135\textwidth]{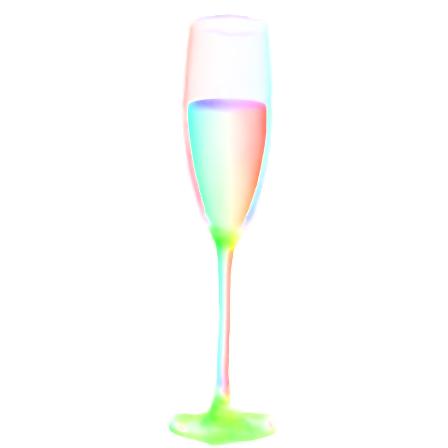}
    \includegraphics[width=0.135\textwidth]{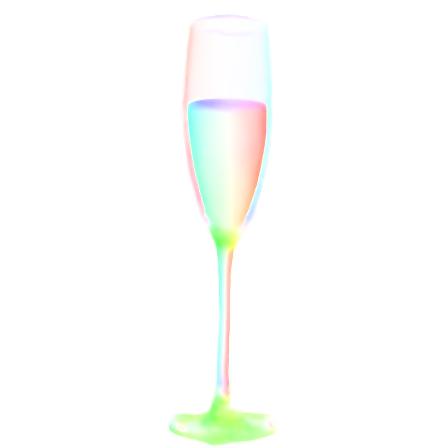}
    \includegraphics[width=0.135\textwidth]{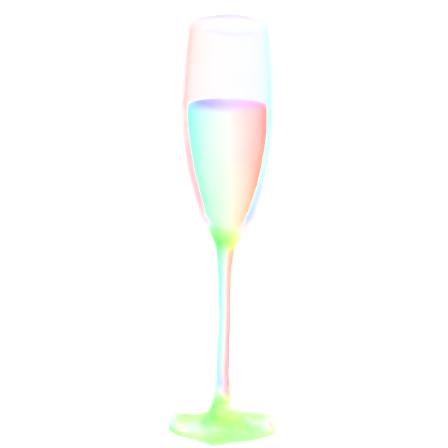} \\
    \vspace{-0.4em}
    \makebox[0.135\textwidth]{\scriptsize Flow $\times 0.9$}
    \makebox[0.135\textwidth]{\scriptsize Flow $\times 0.8$}
    \makebox[0.135\textwidth]{\scriptsize Flow $\times 0.7$}
    \makebox[0.135\textwidth]{\scriptsize GT}
    \makebox[0.135\textwidth]{\scriptsize Flow $\times 0.6$}
    \makebox[0.135\textwidth]{\scriptsize Flow $\times 0.5$}
    \makebox[0.135\textwidth]{\scriptsize Flow $\times 0.4$}
    \\
    \raisebox{0.3\height}{\rotatebox{90}{\scriptsize Rec. Image}}
    \includegraphics[width=0.135\textwidth]{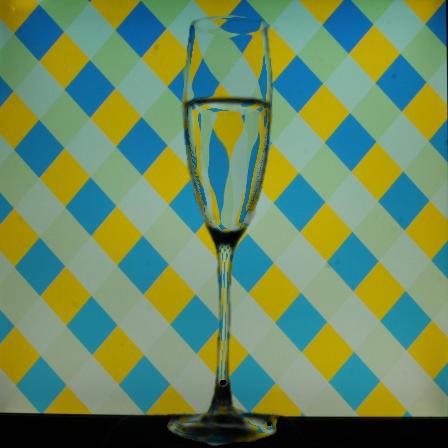}
    \includegraphics[width=0.135\textwidth]{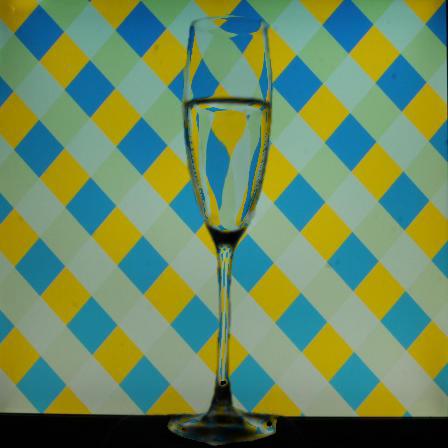}
    \includegraphics[width=0.135\textwidth]{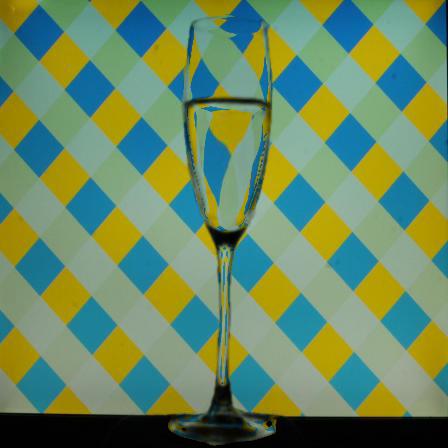}
    \includegraphics[width=0.135\textwidth]{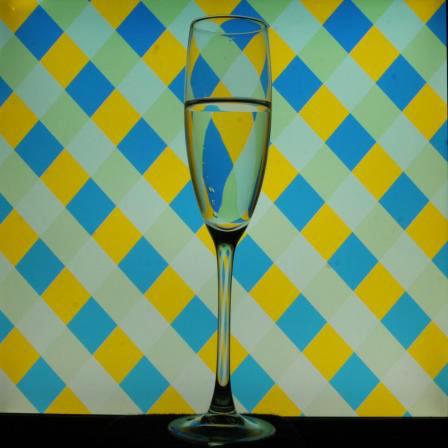}
    \includegraphics[width=0.135\textwidth]{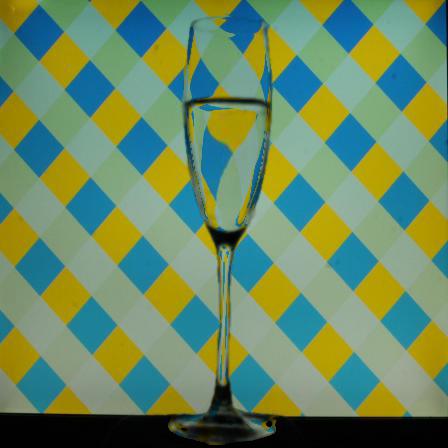}
    \includegraphics[width=0.135\textwidth]{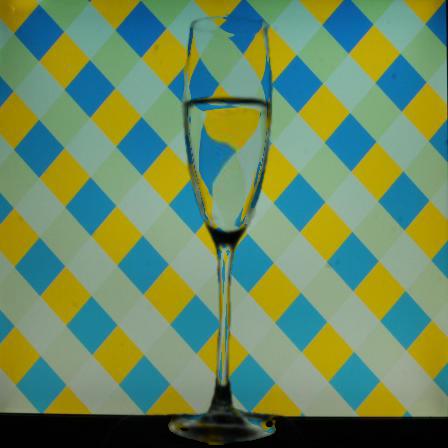}
    \includegraphics[width=0.135\textwidth]{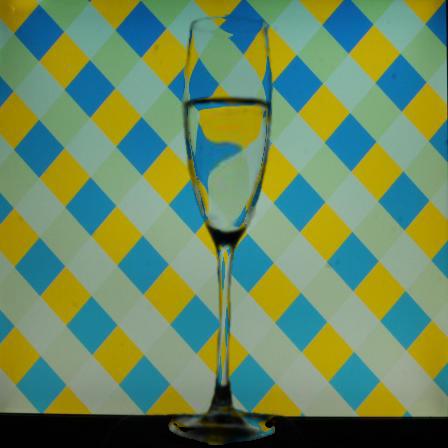} \\
    \vspace{-0.4em}
    \makebox[0.01\textwidth]{\scriptsize }
    \makebox[0.135\textwidth]{\scriptsize P=26.06, S=0.95}
    \makebox[0.135\textwidth]{\scriptsize P=26.54, S=0.96}
    \makebox[0.135\textwidth]{\scriptsize P=27.07, S=0.96}
    \makebox[0.135\textwidth]{\scriptsize }
    \makebox[0.135\textwidth]{\scriptsize \textbf{P=27.18, S=0.96}}
    \makebox[0.135\textwidth]{\scriptsize P=26.70, S=0.96}
    \makebox[0.135\textwidth]{\scriptsize P=25.94, S=0.96}
    \\
    \raisebox{0.3\height}{\rotatebox{90}{\scriptsize Rec. Error}}
    \includegraphics[width=0.135\textwidth]{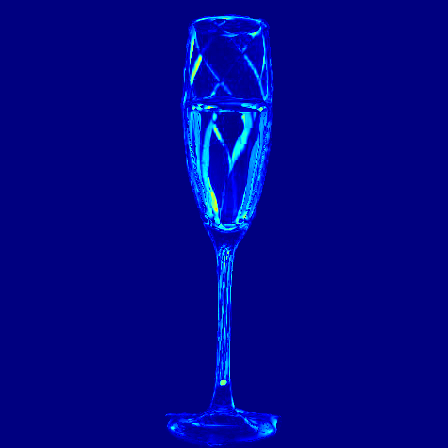}
    \includegraphics[width=0.135\textwidth]{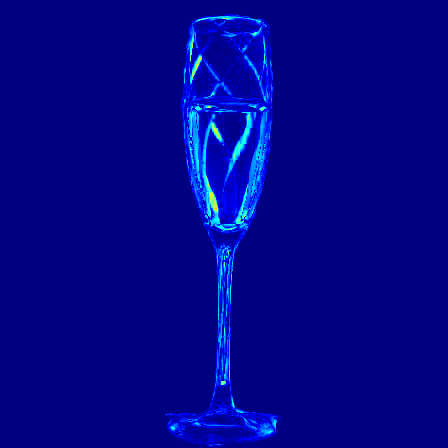}
    \includegraphics[width=0.135\textwidth]{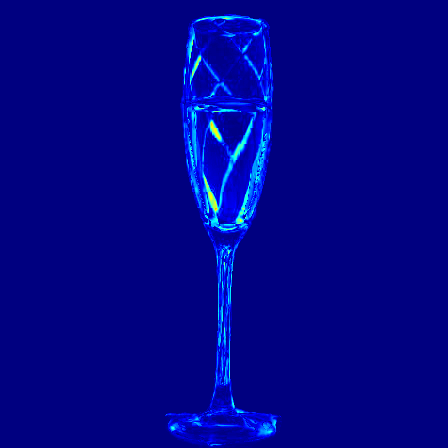}
    \makebox[0.135\textwidth]{}
    \includegraphics[width=0.135\textwidth]{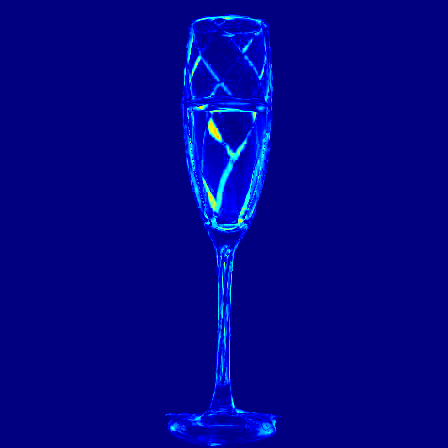}
    \includegraphics[width=0.135\textwidth]{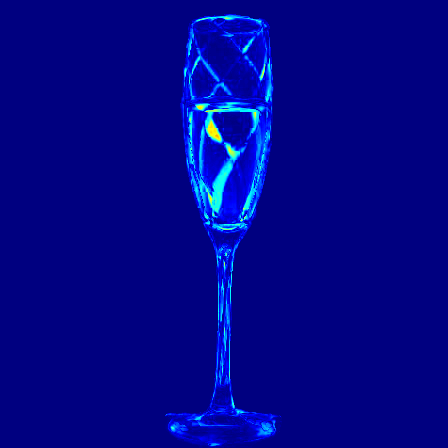}
    \includegraphics[width=0.135\textwidth]{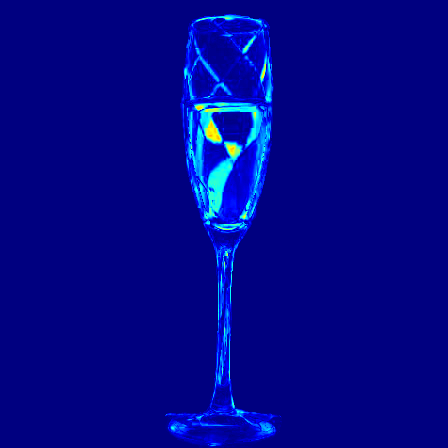}

%% file: sup_editFlow.tex
    \makebox[0.01\textwidth]{\footnotesize }
    \makebox[0.135\textwidth]{\footnotesize No Editing}
    \makebox[0.135\textwidth]{\footnotesize Translate to left}
    \makebox[0.135\textwidth]{\footnotesize Translate to right}
    \makebox[0.135\textwidth]{\footnotesize Rotate 40$\degree$}
    \makebox[0.135\textwidth]{\footnotesize Rotate -40$\degree$}
    \makebox[0.135\textwidth]{\footnotesize Rescale $\times 0.78$}
    \makebox[0.135\textwidth]{\footnotesize Rescale $\times 1.35$}
    \\
    \raisebox{0.3\height}{\rotatebox{90}{\scriptsize Composite}}
    \includegraphics[width=0.135\textwidth]{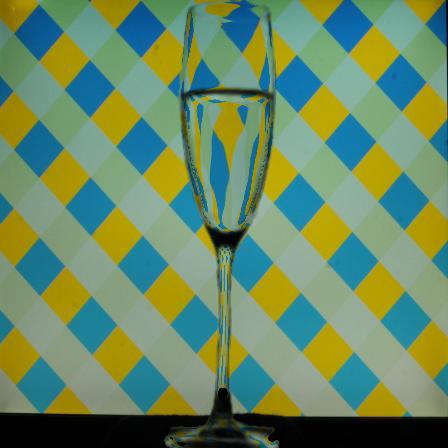}
    \includegraphics[width=0.135\textwidth]{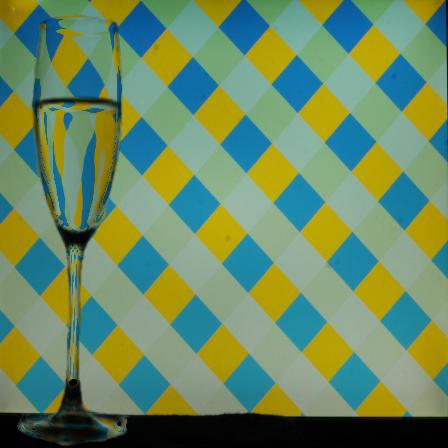}
    \includegraphics[width=0.135\textwidth]{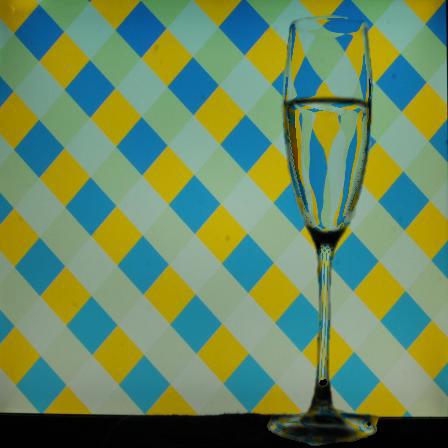} 
    \includegraphics[width=0.135\textwidth]{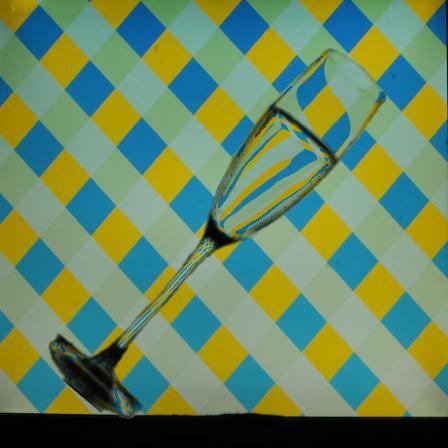}
    \includegraphics[width=0.135\textwidth]{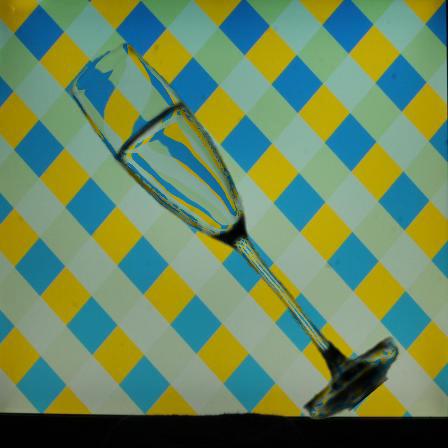}
    \includegraphics[width=0.135\textwidth]{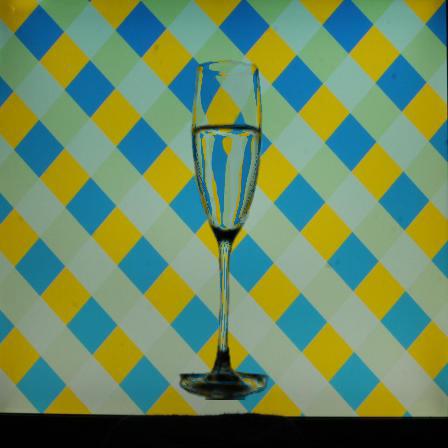}
    \includegraphics[width=0.135\textwidth]{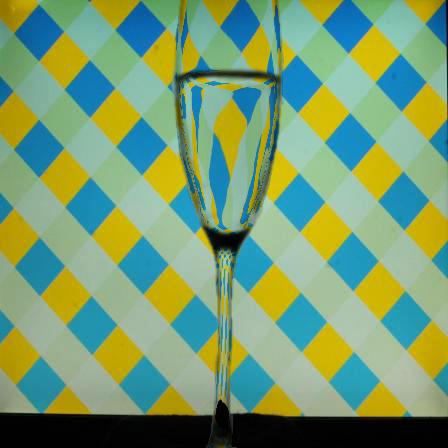}

%% file: syn_quant_trimap_bg.tex
    \begin{minipage}{0.93\textwidth}
    \resizebox{\textwidth}{!}{
    \Huge
        \begin{tabular}{c|*{4}{c}|*{4}{c}|*{4}{c}|*{4}{c}|*{4}{c}}
        \toprule
        \multirow{2}{*}{} & \multicolumn{4}{c}{Glass} 
                               & \multicolumn{4}{c}{Glass with Water} 
                               & \multicolumn{4}{c}{Lens} 
                               & \multicolumn{4}{c}{Complex Shape}  
                               & \multicolumn{4}{c}{Average}  \\
                               & \cellcolor{red!25} F-EPE & \cellcolor{red!25}A-MSE & \cellcolor{red!25}I-MSE & \cellcolor{blue!25} M-IoU 
                               & \cellcolor{red!25} F-EPE & \cellcolor{red!25}A-MSE & \cellcolor{red!25}I-MSE & \cellcolor{blue!25} M-IoU 
                               & \cellcolor{red!25} F-EPE & \cellcolor{red!25}A-MSE & \cellcolor{red!25}I-MSE & \cellcolor{blue!25} M-IoU 
                               & \cellcolor{red!25} F-EPE & \cellcolor{red!25}A-MSE & \cellcolor{red!25}I-MSE & \cellcolor{blue!25} M-IoU 
                               & \cellcolor{red!25} F-EPE & \cellcolor{red!25}A-MSE & \cellcolor{red!25}I-MSE & \cellcolor{blue!25} M-IoU \\
        \midrule
        Background    & 3.6 / 30.3 & 1.33 & 0.48 & 0.12 & 6.4 / 53.2 & 1.54 & 0.68 & 0.12 & 10.3 / 39.2 & 1.94 & 1.57 & 0.24 & 6.8 / 56.8 & 2.50 & 0.85 & 0.11 & 6.8 / 44.9 & 1.83 & 0.90 & 0.15 \\
        \midrule
             TOM-Net       & 1.9 / 14.7 & 0.21 & 0.14 & 0.97 & 2.9 / 21.8 & 0.30 & 0.22 & 0.97 & 1.9 / 6.6   & 0.15 & 0.29 & 0.99 & 4.1 / 31.5 & 0.37 & 0.32 & 0.92 & 2.7 / 18.6 & \textbf{0.26} & 0.24 & 0.96 \\
        TOM-Net$^{\text{+Trimap}}$ & 1.8 / 14.4 & 0.21 & 0.14 & 0.98 & 2.6 / 20.7 & 0.29 & 0.20 & 0.98 & 1.7 / 6.1   & 0.15 & 0.27 & 1.00 & 3.7 / 29.4 & 0.37 & 0.29 & 0.95 & 2.5 / 17.7 & \textbf{0.26} & 0.23 & \textbf{0.98} \\
TOM-Net$^{\text{+Bg}}$     & 1.6 / 13.1 & 0.21 & 0.12 & 0.99 & 2.4 / 19.3 & 0.29 & 0.19 & 0.98 & 1.4 / 4.9   & 0.18 & 0.19 & 1.00 & 3.5 / 27.7 & 0.36 & 0.27 & 0.97 & \textbf{2.2 / 16.2} & \textbf{0.26} & \textbf{0.19} & \textbf{0.98}\\
        \bottomrule
    \end{tabular}
    }
    \end{minipage}
    \hspace{-0.6em}
    \begin{minipage}{0.06\textwidth}
        \resizebox{\textwidth}{!}{
        \begin{tabular}{cc}
            \midrule
            MSE ($\cdot10^{-2}$) \\
            \cellcolor{red!25} $\downarrow$ better \\ 
            \cellcolor{blue!25} $\uparrow$ better\\
            \midrule
        \end{tabular}
        }
    \end{minipage}

%% file: sup_failure.tex
    \includegraphics[width=0.113\textwidth]{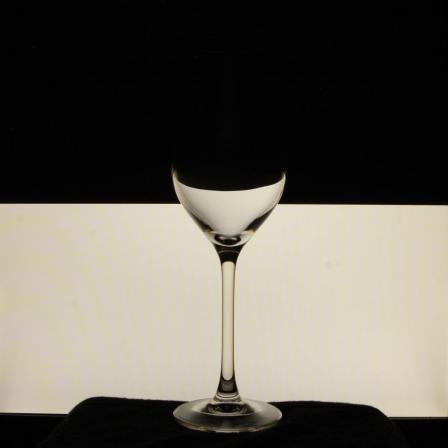}
    \includegraphics[width=0.113\textwidth]{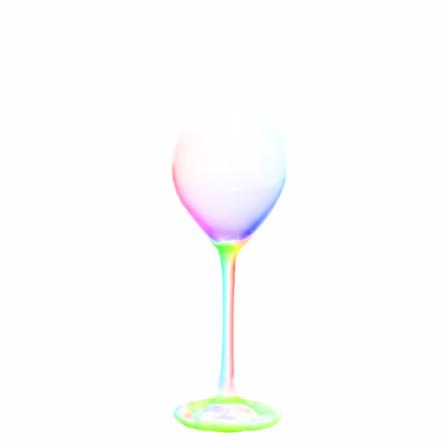}
    \includegraphics[width=0.113\textwidth]{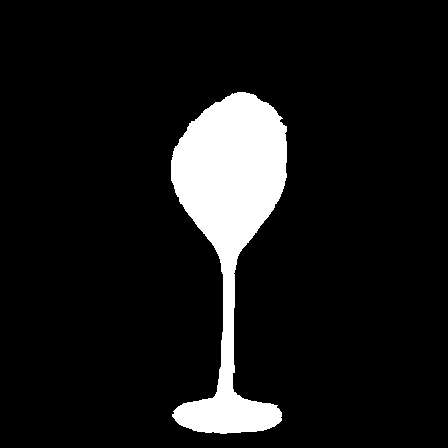}
    \includegraphics[width=0.113\textwidth]{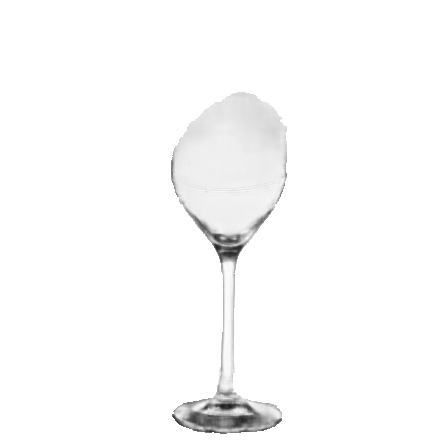}
    \\
    \makebox[0.1\textwidth]{\footnotesize (a)} 
    \\
    \includegraphics[width=0.113\textwidth]{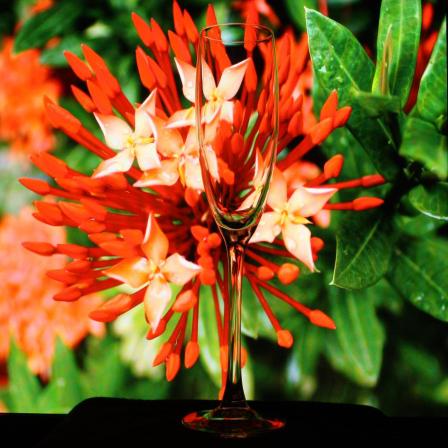}
    \includegraphics[width=0.113\textwidth]{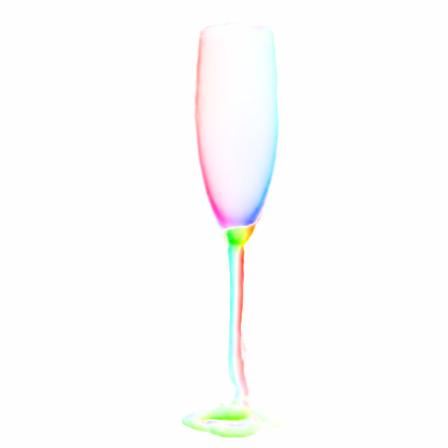}
    \includegraphics[width=0.113\textwidth]{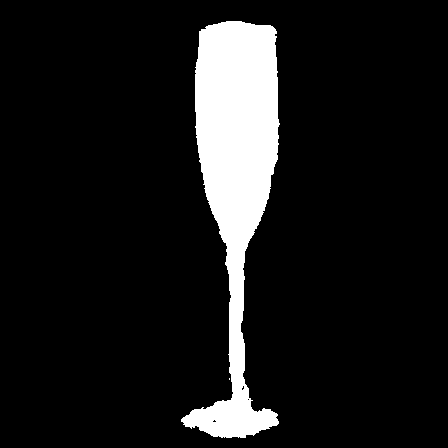}
    \includegraphics[width=0.113\textwidth]{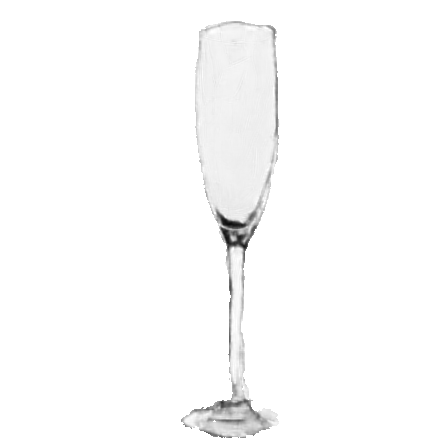}
    \\
    \makebox[0.1\textwidth]{\footnotesize (b)} 

%% file: qual_mono_trimap_stereo.tex
    \makebox[0.27\textwidth]{\scriptsize Testing Input} 
    \makebox[0.13\textwidth]{\scriptsize Rec. Image} 
    \makebox[0.13\textwidth]{\scriptsize Rec. Error} 
    \makebox[0.13\textwidth]{\scriptsize Refractive Flow} 
    \makebox[0.13\textwidth]{\scriptsize Object Mask} 
    \makebox[0.13\textwidth]{\scriptsize Attenuation Mask} 
    \\ 
    \makebox[0.13\textwidth]{} 
    \includegraphics[width=0.13\textwidth]{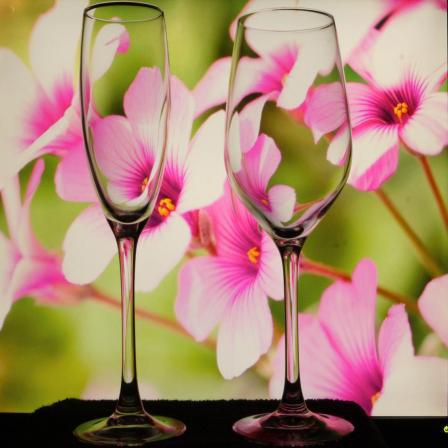}
    \includegraphics[width=0.13\textwidth]{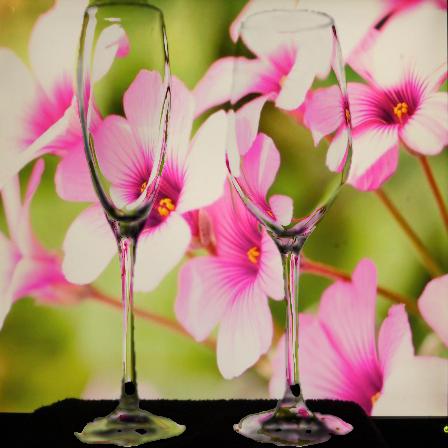}
    \includegraphics[width=0.13\textwidth]{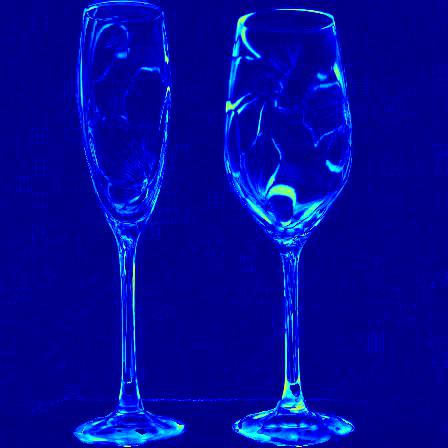}
    \raisebox{0.05\height}{\rotatebox{90}{\tiny P = 22.4, S = 0.87}}
    \includegraphics[width=0.13\textwidth]{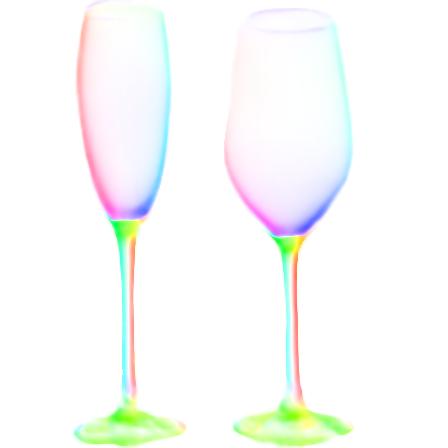}
    \includegraphics[width=0.13\textwidth]{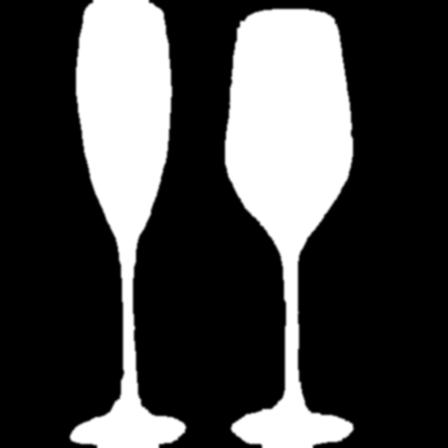}
    \includegraphics[width=0.13\textwidth]{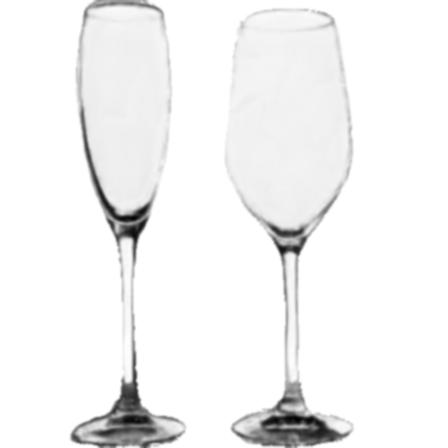}
    \\ 
    \includegraphics[width=0.13\textwidth]{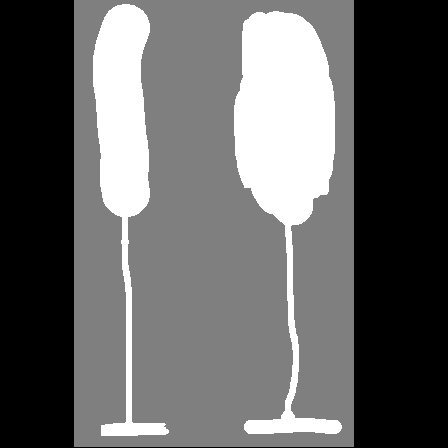}
    \includegraphics[width=0.13\textwidth]{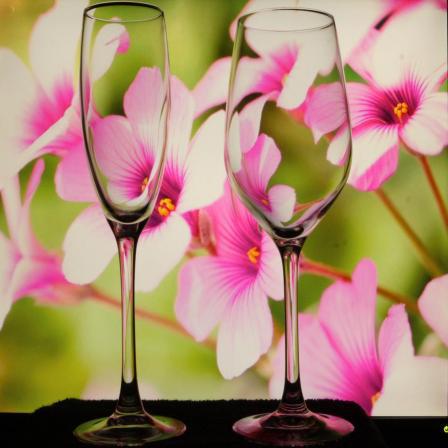}
    \includegraphics[width=0.13\textwidth]{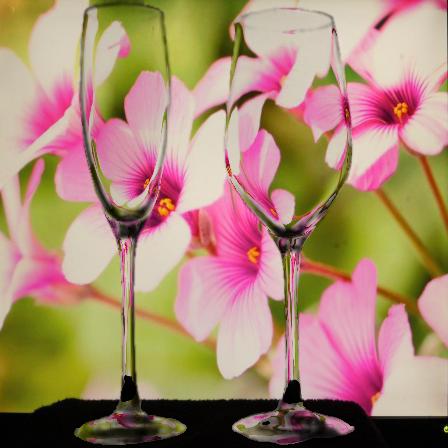}
    \includegraphics[width=0.13\textwidth]{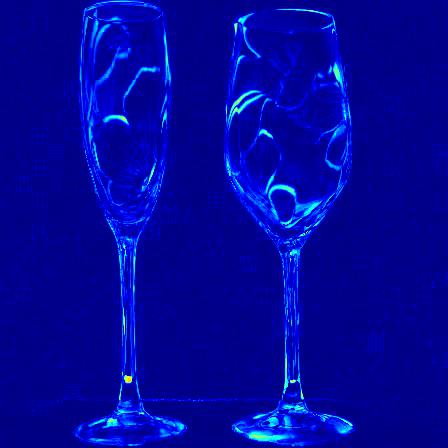}
    \raisebox{0.05\height}{\rotatebox{90}{\tiny P = 23.1, S = 0.87}}
    \includegraphics[width=0.13\textwidth]{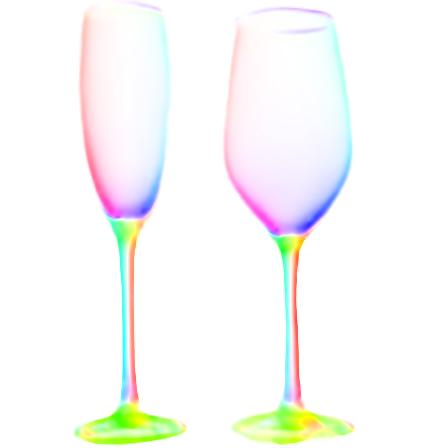}
    \includegraphics[width=0.13\textwidth]{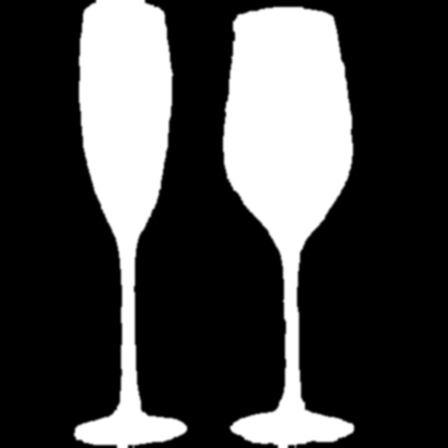}
    \includegraphics[width=0.13\textwidth]{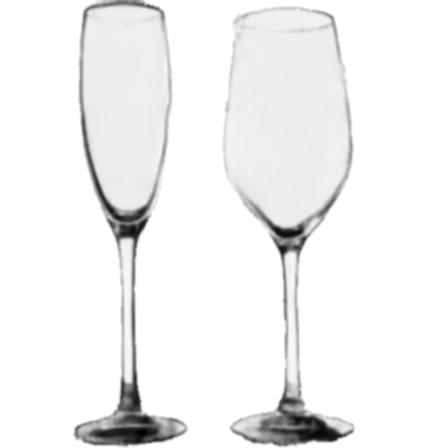}
    \\ 
    \includegraphics[width=0.13\textwidth]{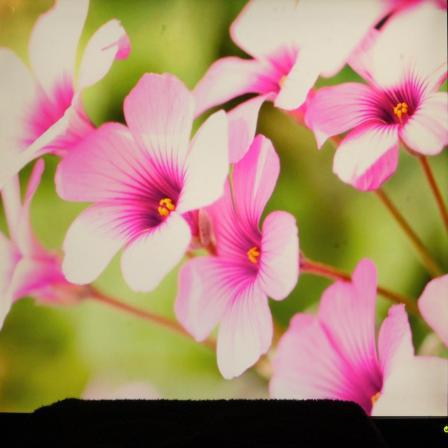}
    \includegraphics[width=0.13\textwidth]{{mono_251_run_model_1.00_1.0_7_tar}.jpg}
    \includegraphics[width=0.13\textwidth]{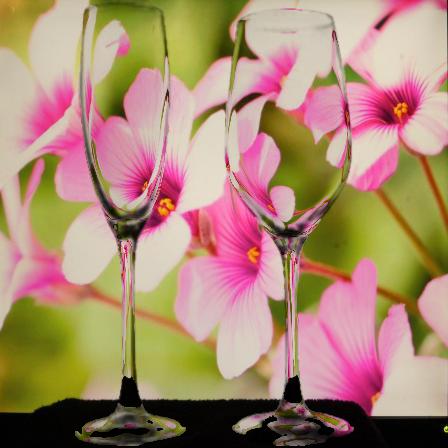}
    \includegraphics[width=0.13\textwidth]{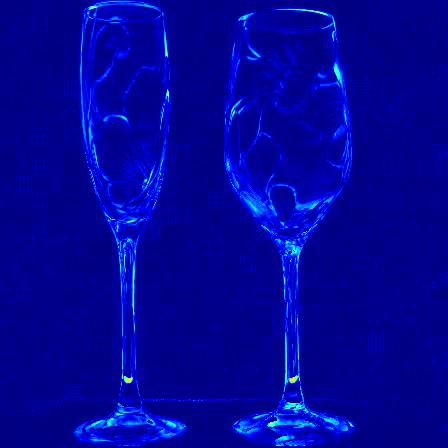}
    \raisebox{0.05\height}{\rotatebox{90}{\tiny P = 24.0, S = 0.89}}
    \includegraphics[width=0.13\textwidth]{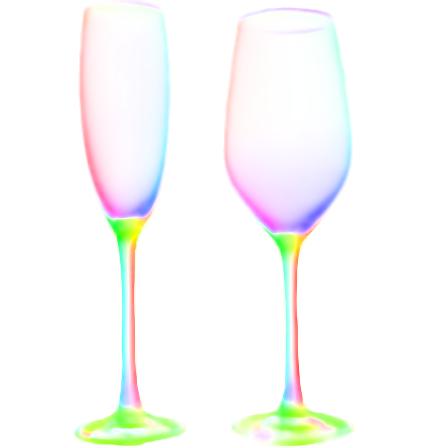}
    \includegraphics[width=0.13\textwidth]{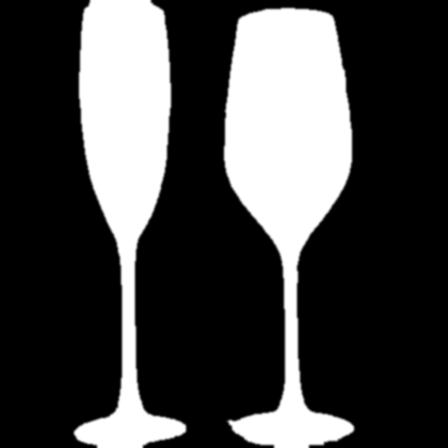}
    \includegraphics[width=0.13\textwidth]{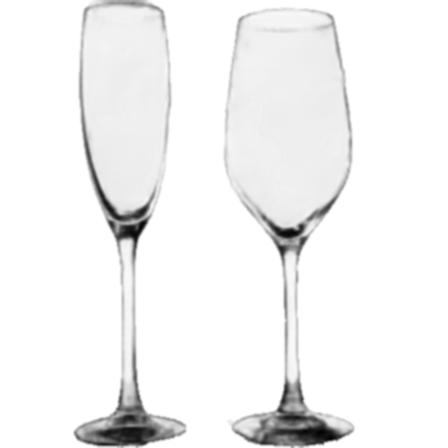}
    \\
    \vspace{-0.3em} \makebox[0.13\textwidth]{\scriptsize (a) Multi-objects}\vspace{0.2em}
    \\
    \makebox[0.13\textwidth]{} 
    \includegraphics[width=0.13\textwidth]{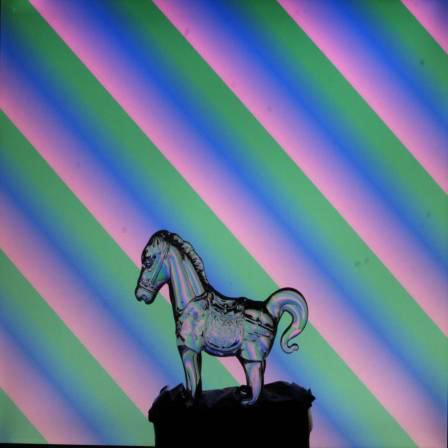}
    \includegraphics[width=0.13\textwidth]{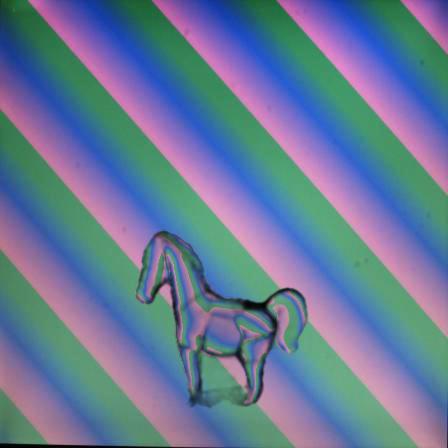}
    \includegraphics[width=0.13\textwidth]{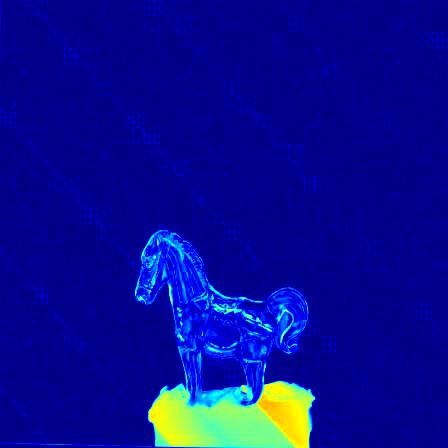}
    \raisebox{0.05\height}{\rotatebox{90}{\tiny P = 20.0, S = 0.89}}
    \includegraphics[width=0.13\textwidth]{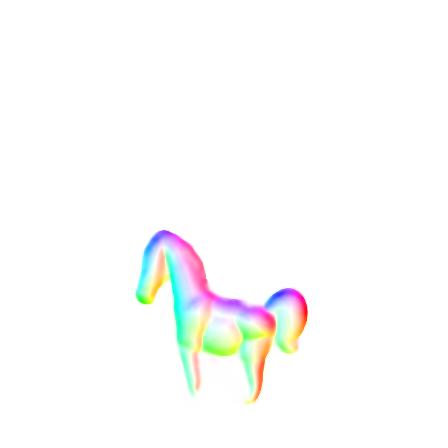}
    \includegraphics[width=0.13\textwidth]{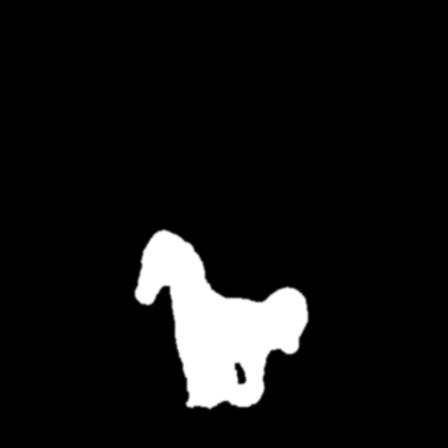}
    \includegraphics[width=0.13\textwidth]{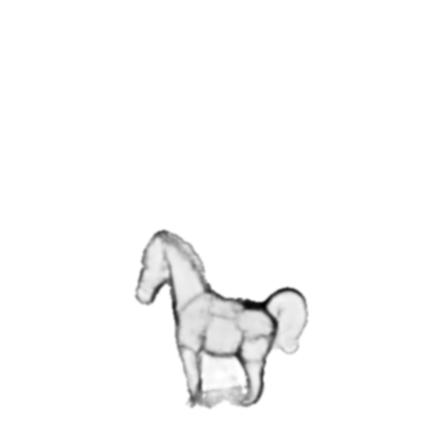}
    \\
    \includegraphics[width=0.13\textwidth]{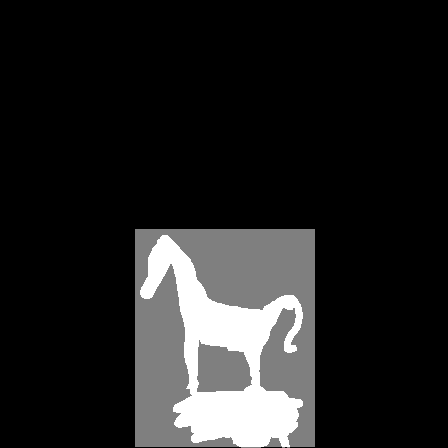}
    \includegraphics[width=0.13\textwidth]{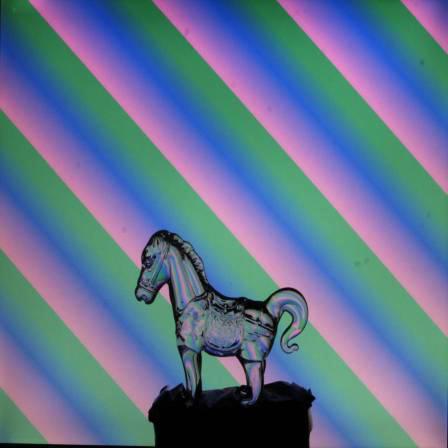}
    \includegraphics[width=0.13\textwidth]{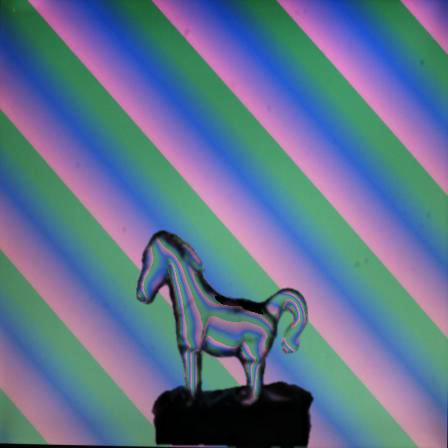}
    \includegraphics[width=0.13\textwidth]{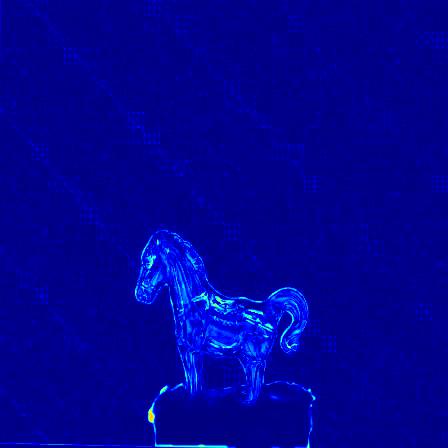}
    \raisebox{0.05\height}{\rotatebox{90}{\tiny P = 28.6, S = 0.94}}
    \includegraphics[width=0.13\textwidth]{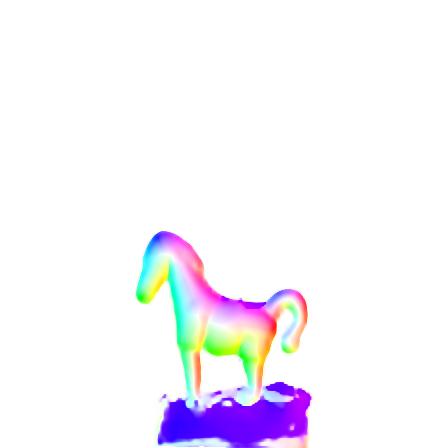}
    \includegraphics[width=0.13\textwidth]{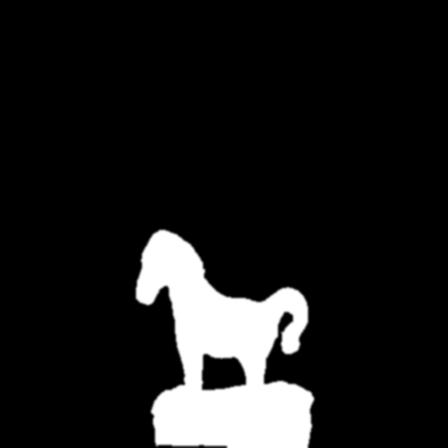}
    \includegraphics[width=0.13\textwidth]{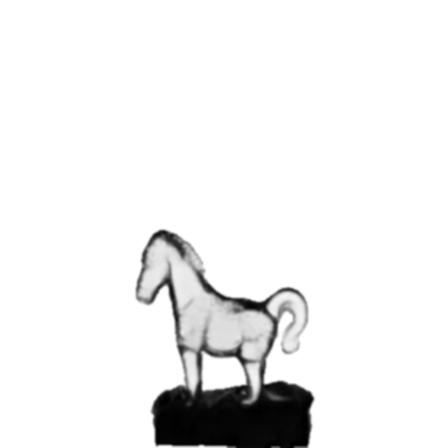}
    \\
    \includegraphics[width=0.13\textwidth]{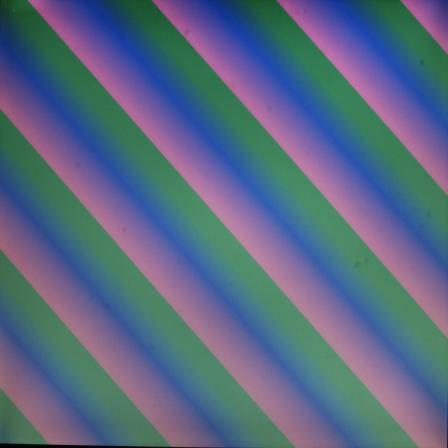}
    \includegraphics[width=0.13\textwidth]{{mono_35_run_model_1.00_1.0_7_tar}.jpg}
    \includegraphics[width=0.13\textwidth]{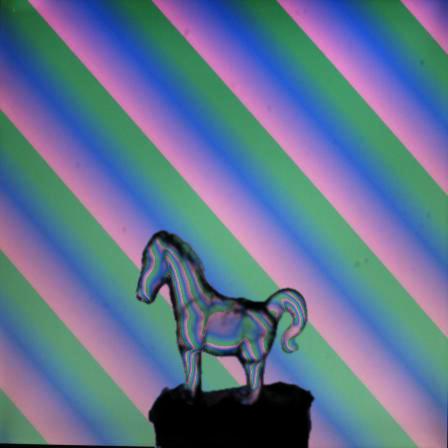}
    \includegraphics[width=0.13\textwidth]{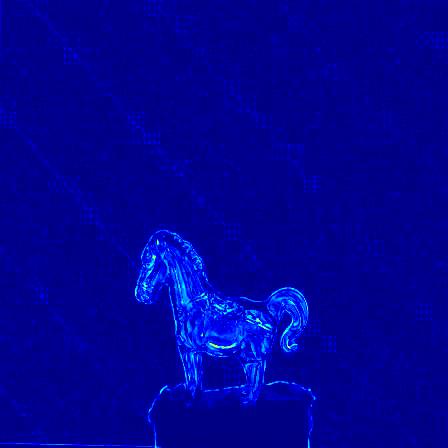}
    \raisebox{0.05\height}{\rotatebox{90}{\tiny P = 29.6, S = 0.94}}
    \includegraphics[width=0.13\textwidth]{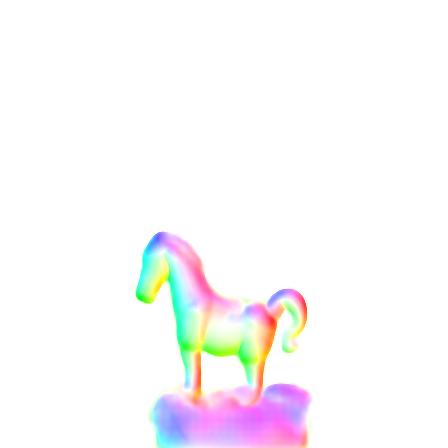}
    \includegraphics[width=0.13\textwidth]{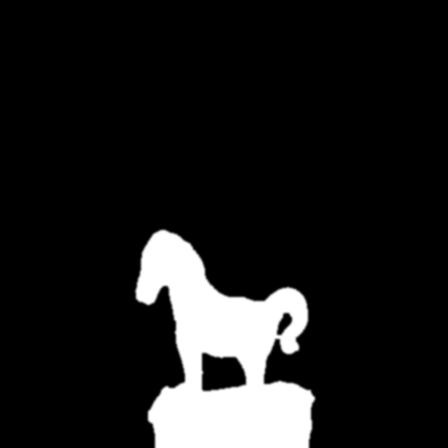}
    \includegraphics[width=0.13\textwidth]{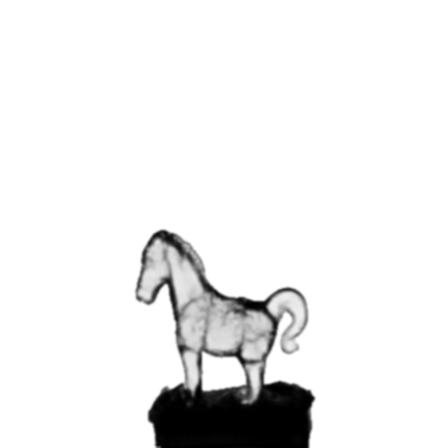}
    \\
    \vspace{-0.3em} \makebox[0.13\textwidth]{\scriptsize (b) Complex Horse} \vspace{0.2em}
    \\
    \makebox[0.13\textwidth]{} 
    \includegraphics[width=0.13\textwidth]{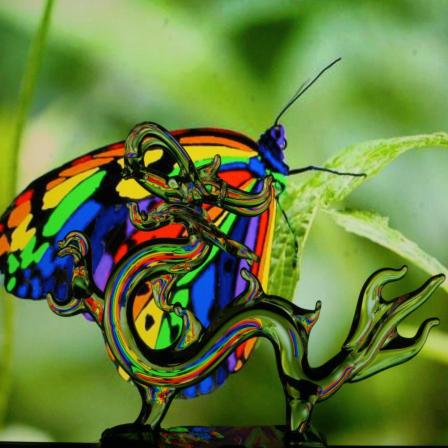}
    \includegraphics[width=0.13\textwidth]{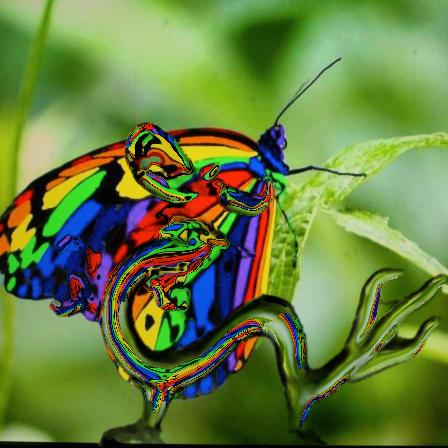}
    \includegraphics[width=0.13\textwidth]{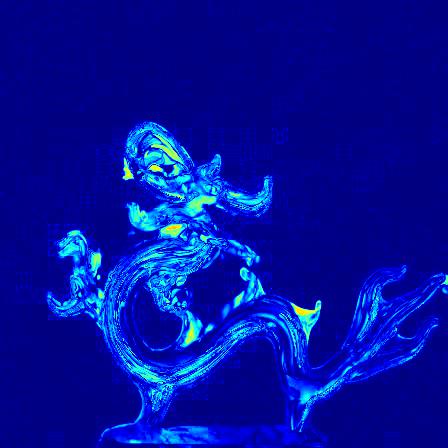}
    \raisebox{0.05\height}{\rotatebox{90}{\tiny P = 18.6, S = 0.81}}
    \includegraphics[width=0.13\textwidth]{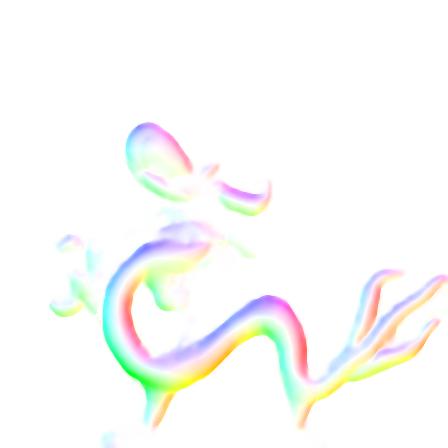}
    \includegraphics[width=0.13\textwidth]{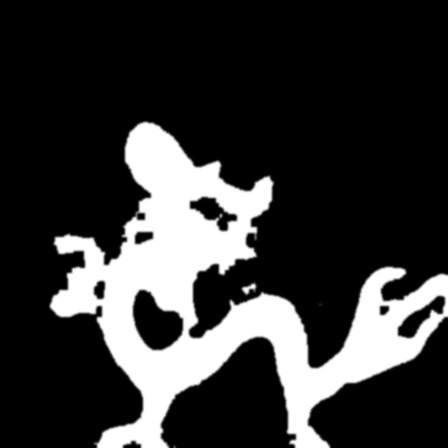}
    \includegraphics[width=0.13\textwidth]{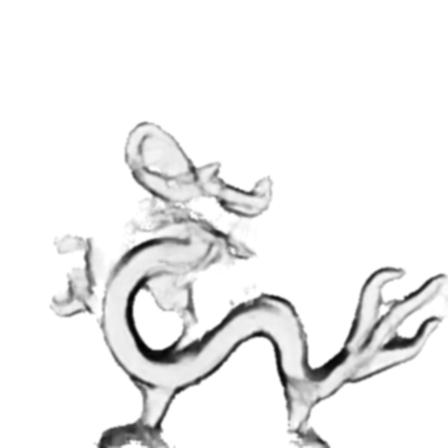}
    \\ 
    \includegraphics[width=0.13\textwidth]{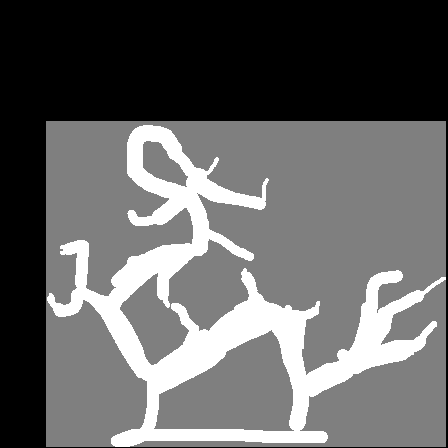}
    \includegraphics[width=0.13\textwidth]{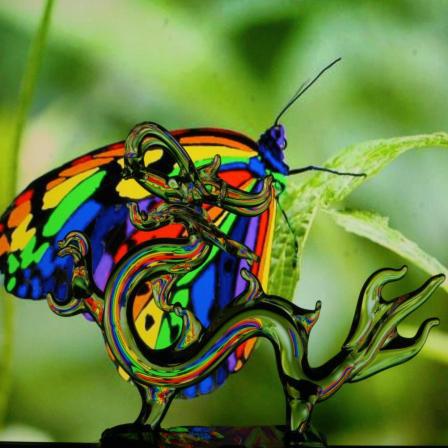}
    \includegraphics[width=0.13\textwidth]{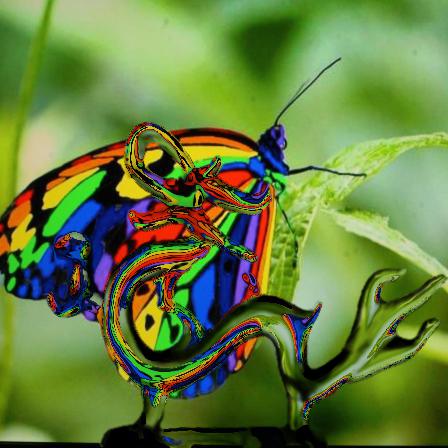}
    \includegraphics[width=0.13\textwidth]{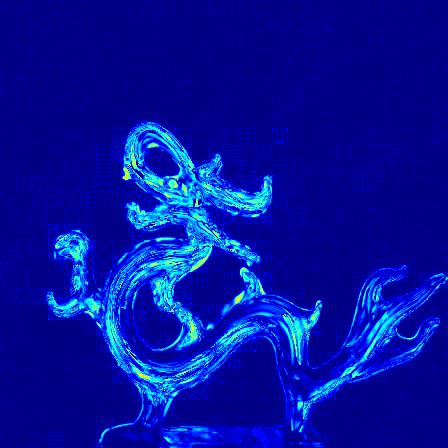}
    \raisebox{0.05\height}{\rotatebox{90}{\tiny P = 20.0, S = 0.81}}
    \includegraphics[width=0.13\textwidth]{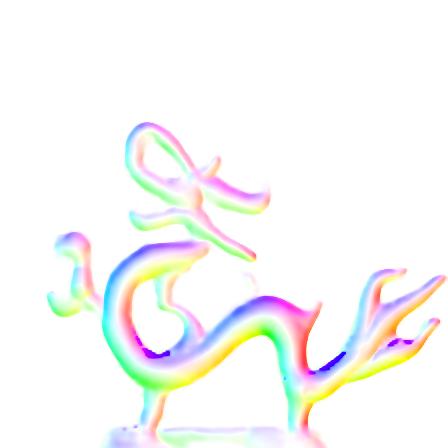}
    \includegraphics[width=0.13\textwidth]{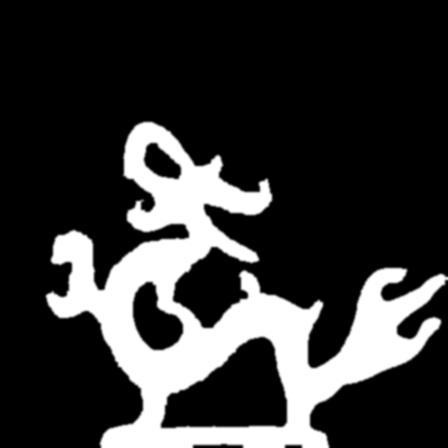}
    \includegraphics[width=0.13\textwidth]{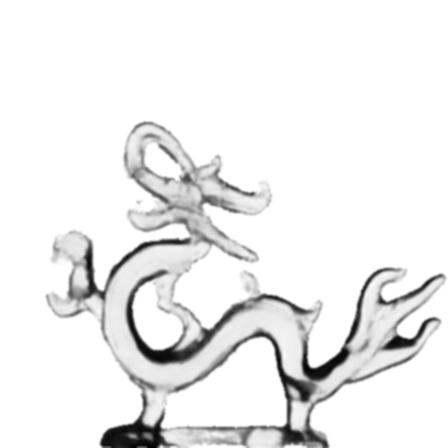}
    \\
    \includegraphics[width=0.13\textwidth]{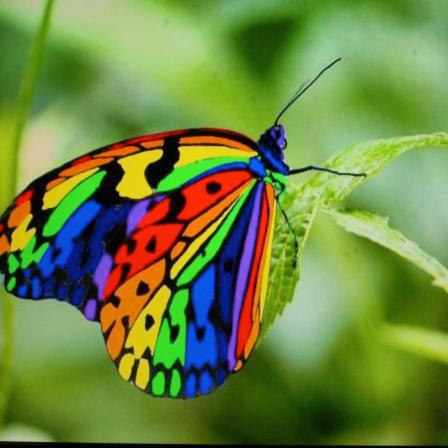}
    \includegraphics[width=0.13\textwidth]{{mono_214_run_model_1.00_1.0_7_tar}.jpg}
    \includegraphics[width=0.13\textwidth]{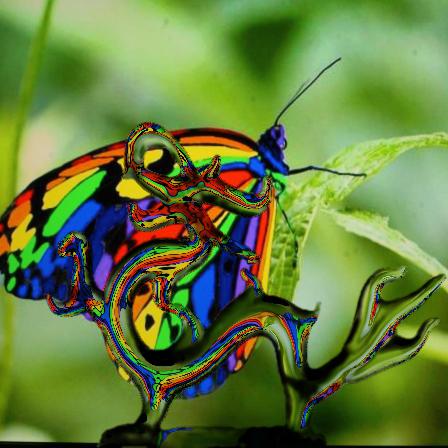}
    \includegraphics[width=0.13\textwidth]{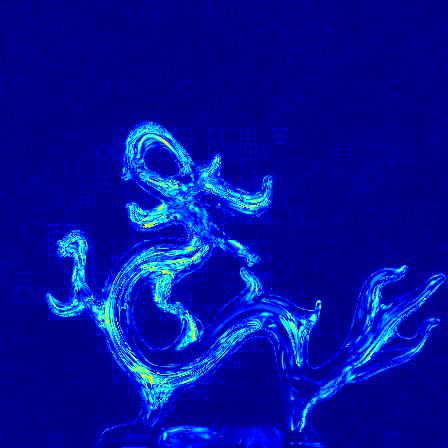}
    \raisebox{0.05\height}{\rotatebox{90}{\tiny P = 20.7, S = 0.84}}
    \includegraphics[width=0.13\textwidth]{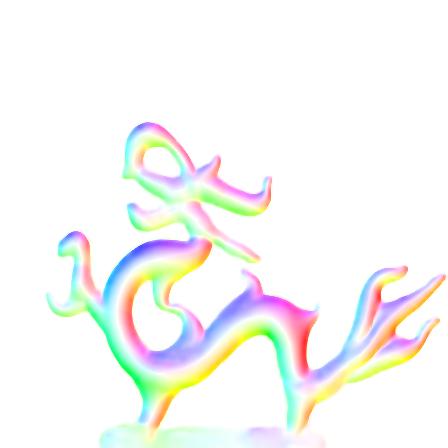}
    \includegraphics[width=0.13\textwidth]{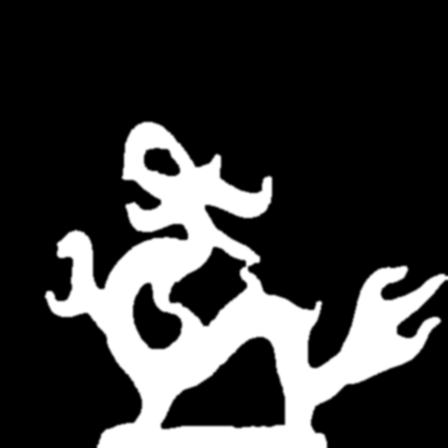}
    \includegraphics[width=0.13\textwidth]{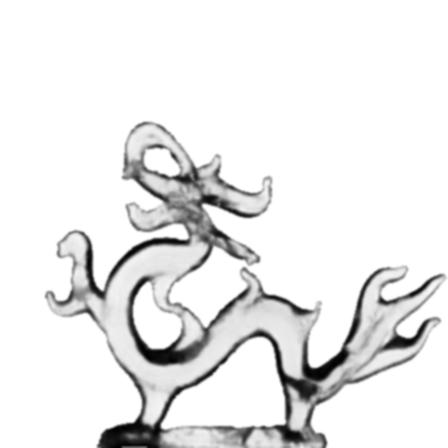}
    \\
    \vspace{-0.3em}\makebox[0.13\textwidth]{\scriptsize (c) Complex Dragon} 

%% file: limitation.tex
    \makebox[0.03\textwidth]{\scriptsize } 
    \makebox[0.15\textwidth]{\scriptsize Background} 
    \makebox[0.15\textwidth]{\scriptsize Input} 
    \makebox[0.15\textwidth]{\scriptsize Rec. Image} 
    \makebox[0.15\textwidth]{\scriptsize Refractive Flow} 
    \makebox[0.15\textwidth]{\scriptsize Object Mask} 
    \makebox[0.15\textwidth]{\scriptsize Attenuation Mask} 
    \\ 
    \raisebox{6\height}{\makebox[0.02\textwidth]{\footnotesize (a)}}
    \includegraphics[width=0.15\textwidth]{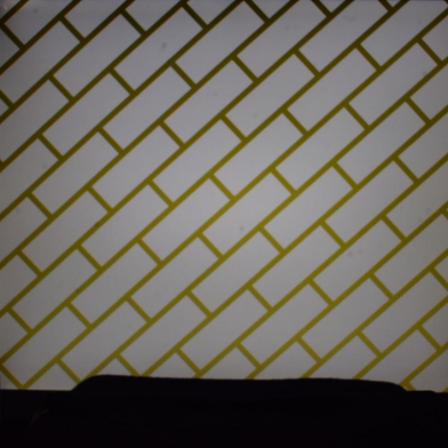}
    \includegraphics[width=0.15\textwidth]{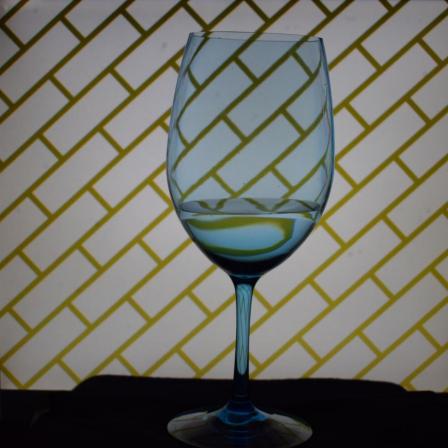}
    \includegraphics[width=0.15\textwidth]{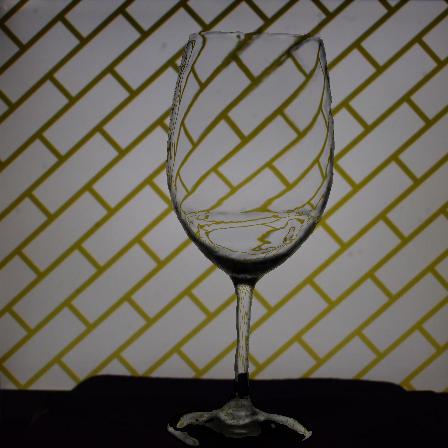}
    \includegraphics[width=0.15\textwidth]{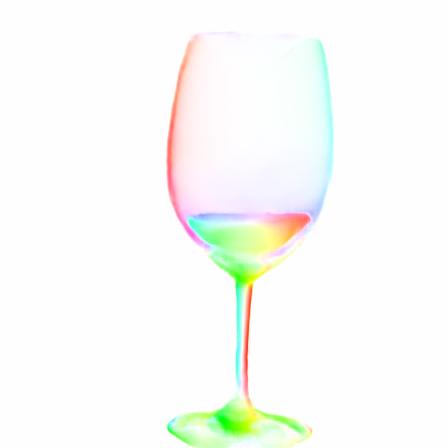}
    \includegraphics[width=0.15\textwidth]{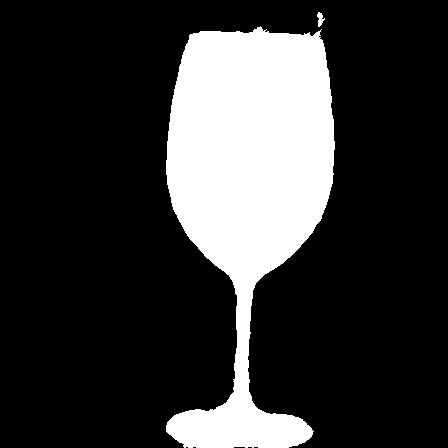}
    \includegraphics[width=0.15\textwidth]{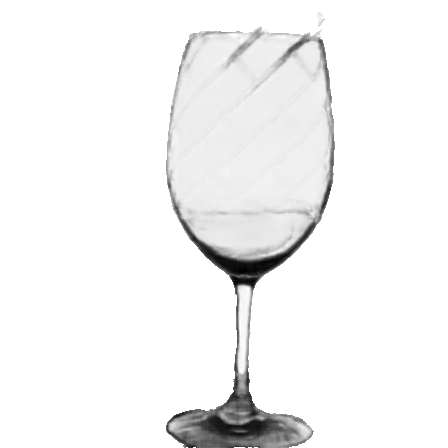}
\\
    \raisebox{6\height}{\makebox[0.02\textwidth]{\footnotesize (b)}}
    \includegraphics[width=0.15\textwidth]{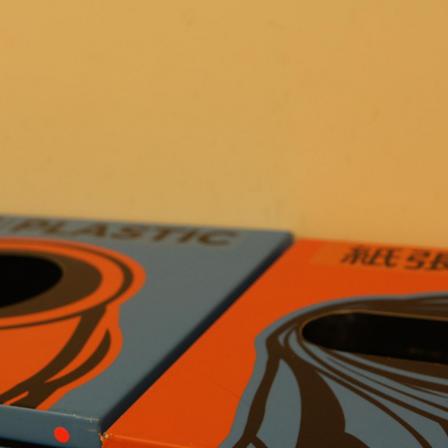}
    \includegraphics[width=0.15\textwidth]{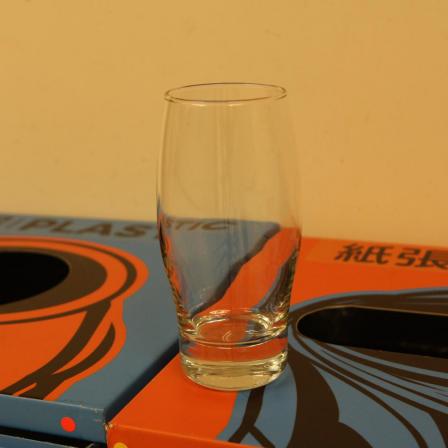}
    \includegraphics[width=0.15\textwidth]{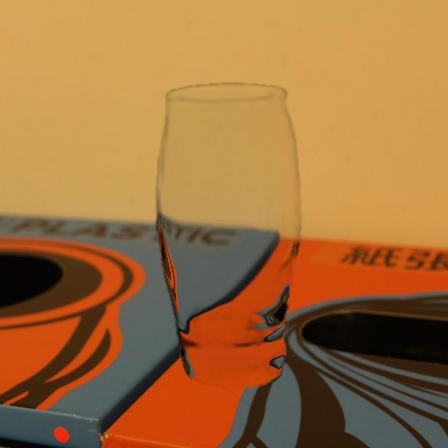}
    \includegraphics[width=0.15\textwidth]{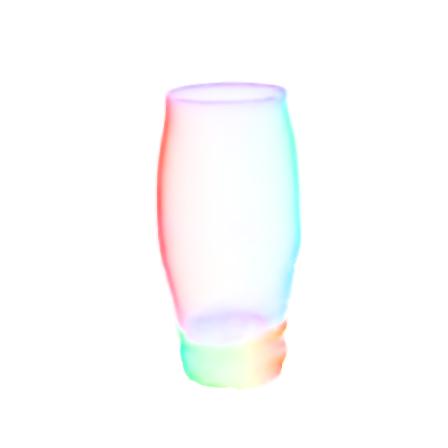}
    \includegraphics[width=0.15\textwidth]{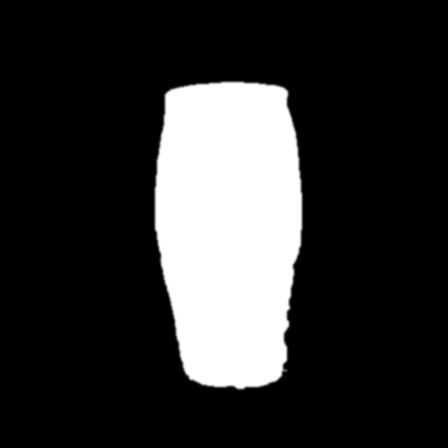}
    \includegraphics[width=0.15\textwidth]{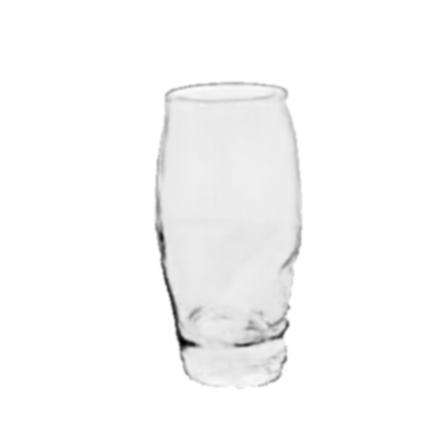}
\\
    \raisebox{6\height}{\makebox[0.02\textwidth]{\footnotesize (c)}}
    \includegraphics[width=0.15\textwidth]{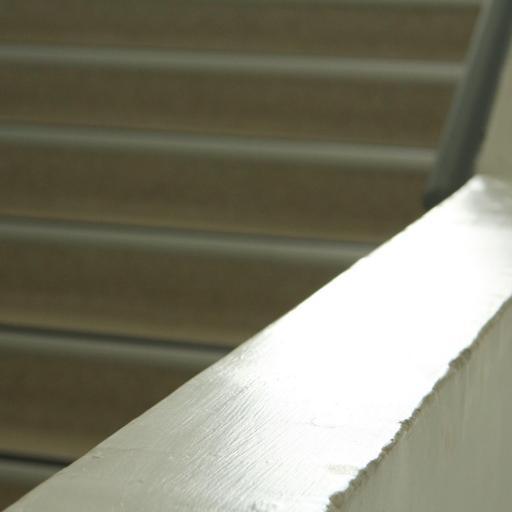}
    \includegraphics[width=0.15\textwidth]{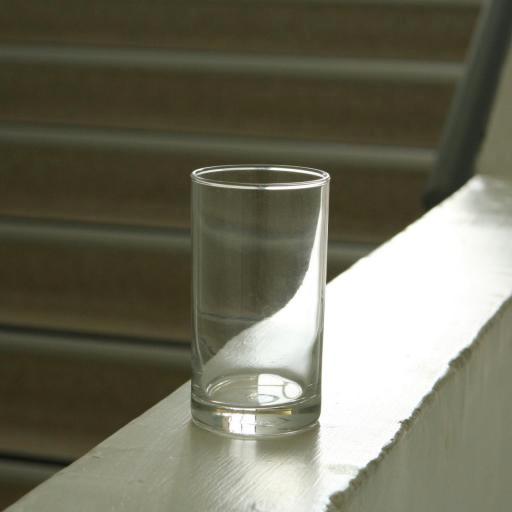}
    \includegraphics[width=0.15\textwidth]{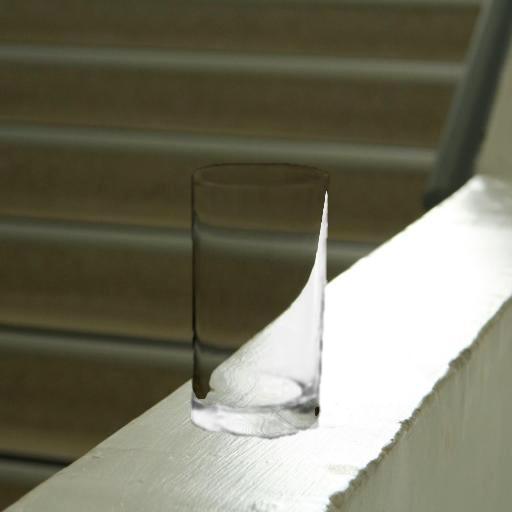}
    \includegraphics[width=0.15\textwidth]{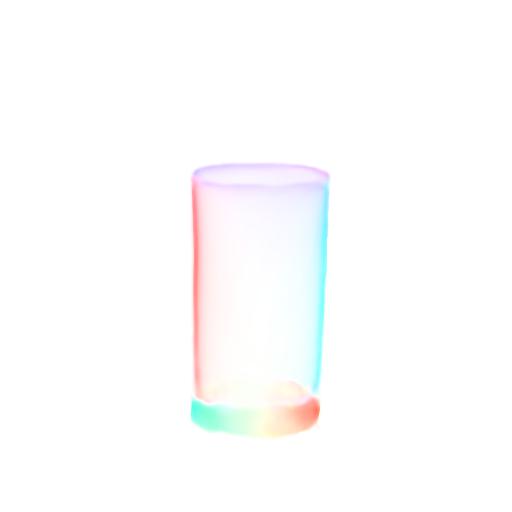}
    \includegraphics[width=0.15\textwidth]{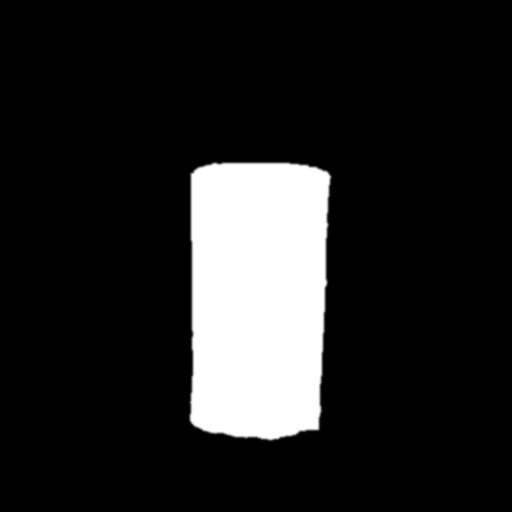}
    \includegraphics[width=0.15\textwidth]{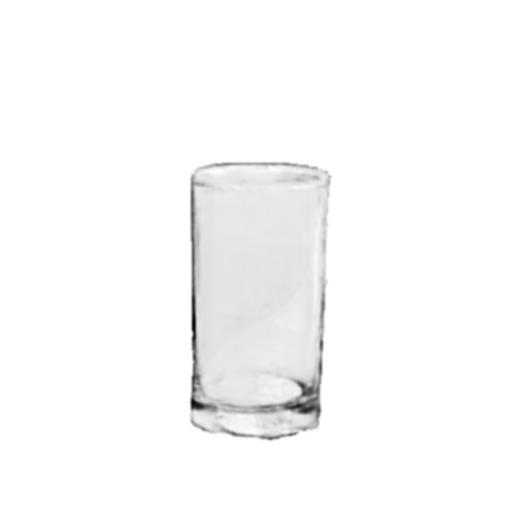}
\\
    \raisebox{6\height}{\makebox[0.02\textwidth]{\footnotesize (d)}}
    \includegraphics[width=0.15\textwidth]{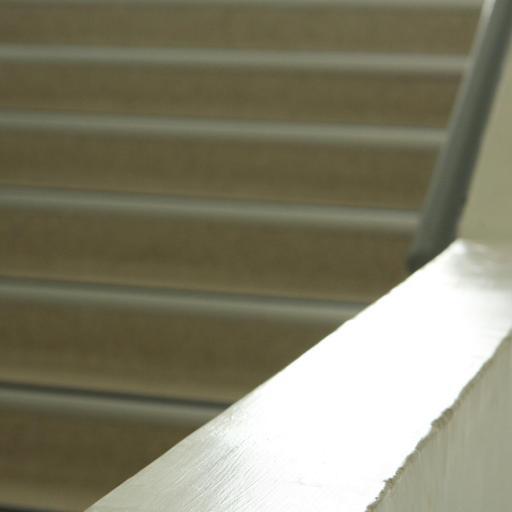}
    \includegraphics[width=0.15\textwidth]{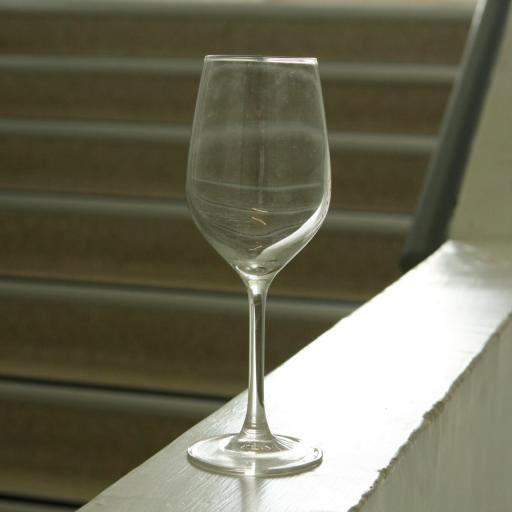}
    \includegraphics[width=0.15\textwidth]{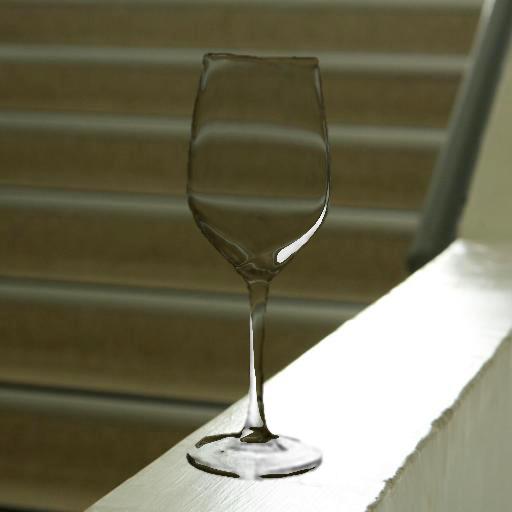}
    \includegraphics[width=0.15\textwidth]{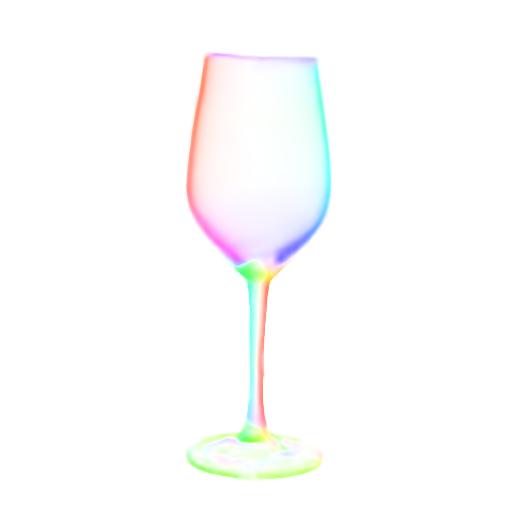}
    \includegraphics[width=0.15\textwidth]{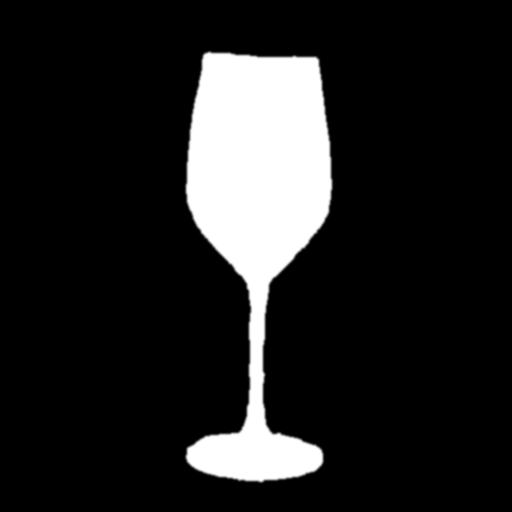}
    \includegraphics[width=0.15\textwidth]{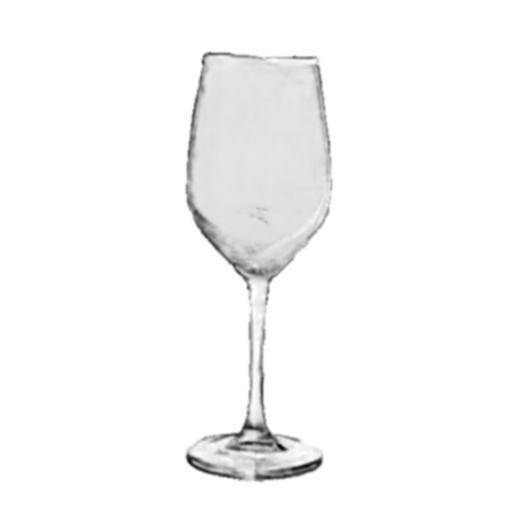}
\\
    \raisebox{6\height}{\makebox[0.02\textwidth]{\footnotesize (e)}}
    \includegraphics[width=0.15\textwidth]{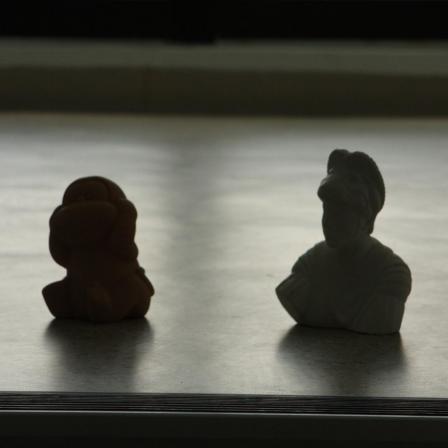}
    \includegraphics[width=0.15\textwidth]{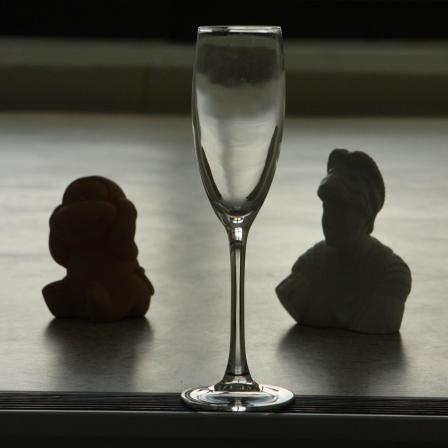}
    \includegraphics[width=0.15\textwidth]{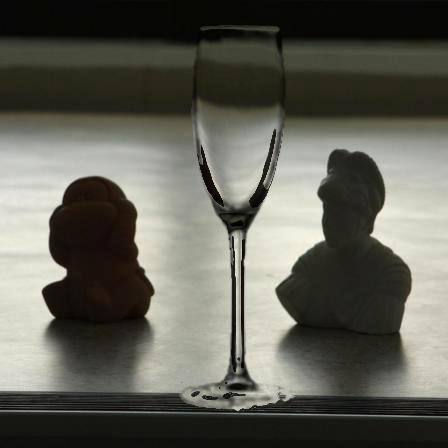}
    \includegraphics[width=0.15\textwidth]{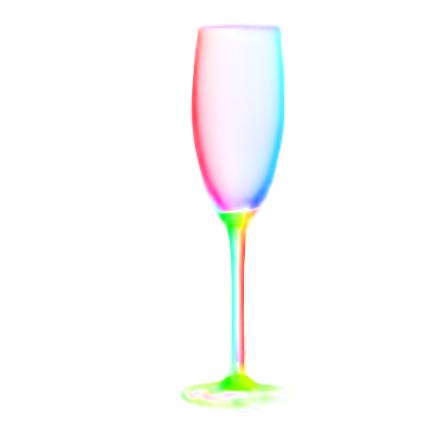}
    \includegraphics[width=0.15\textwidth]{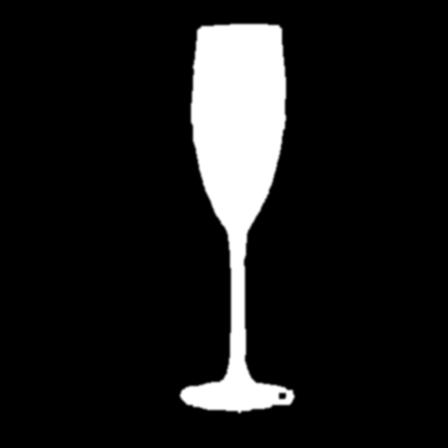}
    \includegraphics[width=0.15\textwidth]{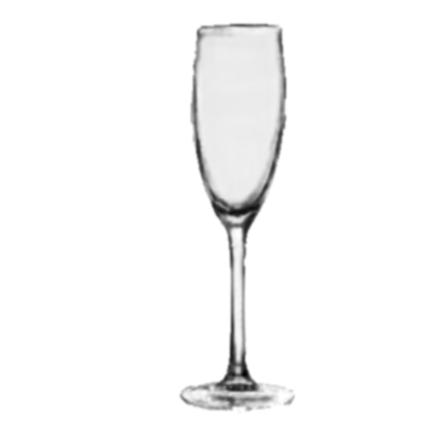}
\\